\documentclass[letterpaper, 10 pt, journal, twoside]{ieeetran}   
\IEEEoverridecommandlockouts                              
\usepackage{times}
\usepackage[pdftex]{graphicx}
\graphicspath{{imgs/pdf/}{imgs/png/}{imgs/jpg/}{imgs/eps/}}
\DeclareGraphicsExtensions{.pdf,.png,.jpg,.eps}
\usepackage[caption=false,font=footnotesize]{subfig}
\usepackage{amsmath} 
\usepackage{amssymb}  
\usepackage{hyperref} %
\usepackage{siunitx}
\usepackage{color}
\usepackage[table]{xcolor}
\usepackage{mathptmx} 
\usepackage{times} 
\usepackage{latexsym}
\usepackage{psfrag}
\usepackage[caption = false, font = footnotesize]{subfig}
\usepackage{cite}
\usepackage{soul}
\usepackage{enumerate}
\usepackage{epsfig}
\usepackage{array} %
\usepackage{wrapfig}
\usepackage{comment}
\usepackage{multirow}
\usepackage{gensymb}
\usepackage[symbols,nogroupskip,sort=none]{glossaries-extra}
\usepackage[normalem]{ulem}
\useunder{\uline}{\ul}{}
\usepackage{algorithm2e}

\begin{document}
\title{Anthropomorphic Twisted String-Actuated Soft Robotic Gripper with Tendon-Based Stiffening}

\markboth{THIS PAPER IS CURRENTLY UNDER REVISION IN IEEE TRANSACTIONS ON ROBOTICS}
{Bombara \MakeLowercase{\textit{et al.}}: Anthropomorphic Twisted String-Actuated Gripper}  

\author{David Bombara$^{* \dagger}$, Revanth Konda$^{*}$, Steven Swanbeck$^{*}$, and Jun Zhang
\thanks{$^*$ D.B., R.K., and S.S. contributed equally to this work.

$^{\dagger}$ Corresponding author.

The authors are with the Department of Mechanical Engineering, University of Nevada, Reno, 1664 N. Virginia St., Reno, NV 89557, USA
{\tt\small \{dbombara, rkonda, stevenswanbeck\}@nevada.unr.edu},
{\tt\small jun@unr.edu}

This work was supported in part by a NASA Space Technology Graduate Research Opportunities award.}}

\maketitle

\begin{abstract}
\textcolor{red}{Realizing high-performance soft robotic grippers is challenging because of the inherent limitations of the soft actuators and artificial muscles that drive them, including low force output, small actuation range, and poor compactness. Despite advances in this area, realizing compact soft grippers with high dexterity and force output is still challenging.} This paper explores twisted string actuators (TSAs) to drive a soft robotic gripper. TSAs have been used in numerous robotic applications, but their inclusion in soft robots has been limited. The proposed design of the gripper was inspired by the human hand. Tunable stiffness was implemented in the fingers with antagonistic TSAs. The fingers’ bending angles, actuation speed, blocked force output, and stiffness tuning were experimentally characterized. The gripper achieved a score of 6 on the Kapandji test and recreated 31 of the 33 grasps of the Feix GRASP taxonomy. It exhibited a maximum grasping force of 72 N, which was almost 13 times its own weight. A comparison study revealed that the proposed gripper exhibited equivalent or superior performance compared to other similar soft grippers.
\end{abstract}
\section{Introduction}
\textcolor{red}{Traditional robotic grippers, which utilize rigid components and actuators, offer high grasp strengths, high dexterity, and high accuracy and precision in executing complex manipulation tasks}. However, this rigidity is not ideal for grasping delicate, deformable, or soft objects because rigid grippers lack the compliance to conform to different shapes \cite{Shih2017,Yafeng2021}. Furthermore, grippers which exhibit high dexterity are usually difficult to control and design with rigid parts \cite{MATTAR2013517,Deimal2016}. In addition, to grasp delicate objects, traditional robotic grippers require complex force control algorithms \cite{Shih2017,SoftGripperReview}. In contrast, soft robotic grippers have been demonstrated to be better suited for dexterous manipulation, manipulation of wide range of objects with different shapes and sizes, and even human-robot interaction without requiring highly complicated force control strategies \cite{Shih2017,Yafeng2021,Deimal2016,SMA_Hand2015,RBO3}. 
\begin{figure}
    \centering
    %
    %
    \subfloat[]{
    \includegraphics[height=5.5cm]{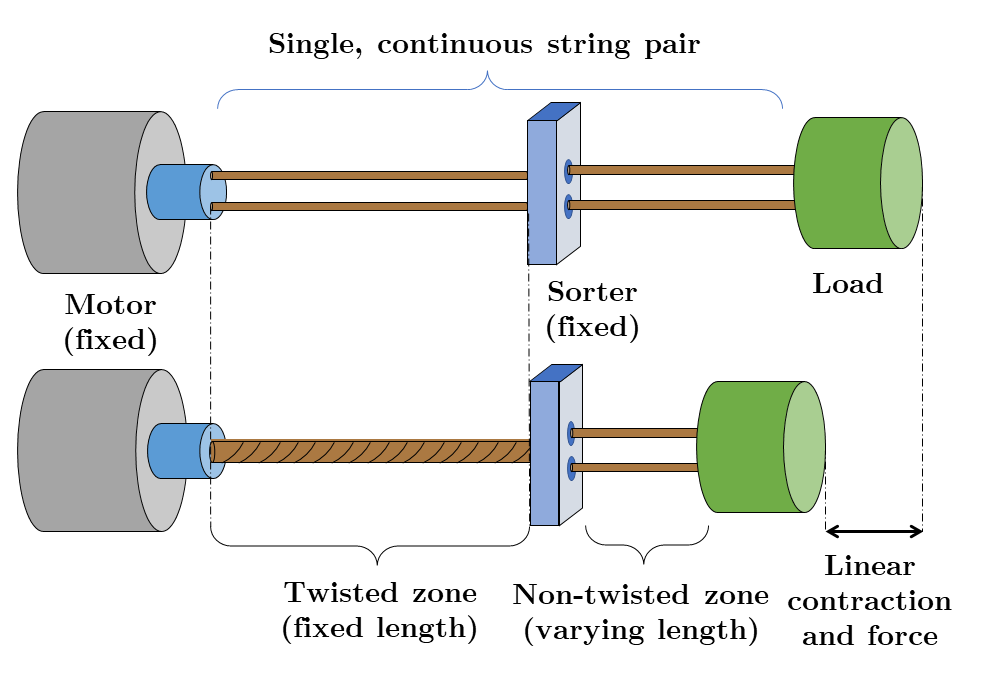}}
    \hfill
    \subfloat[]{
    \includegraphics[height=5.5cm]{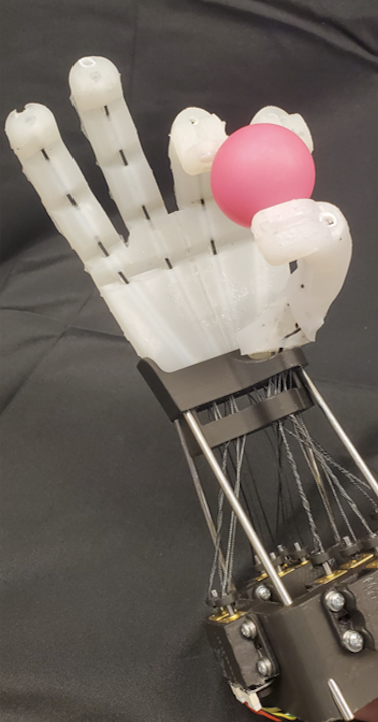}}
    \hfill
    \subfloat[]{
    \includegraphics[height=5.5cm]{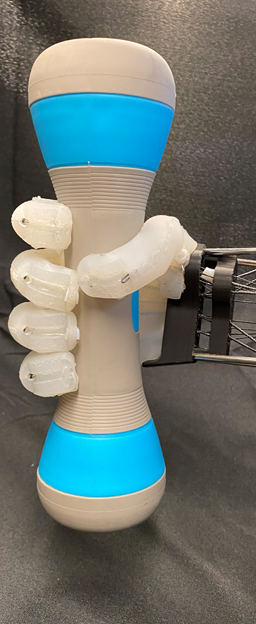}}
    \hfill
    \subfloat[]{
    \includegraphics[height=5.5cm]{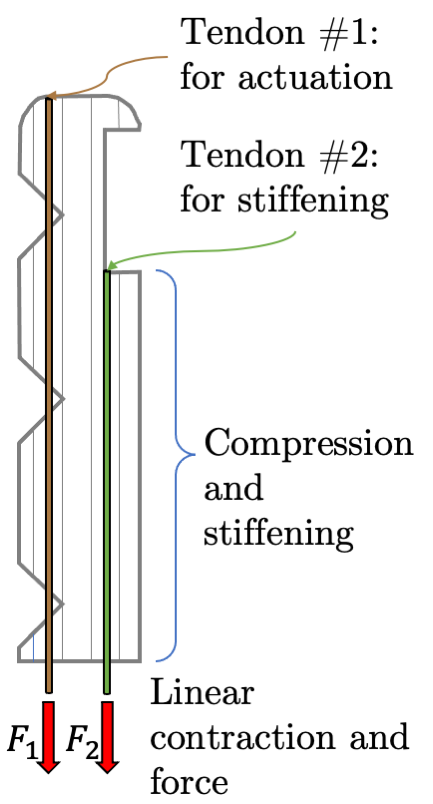}}
    \hfill
    \caption{{(a) Schematic of the twisted string actuator (TSA) with the fixed twisting zone and sorter mechanism. (b)-(d): The anthropomorphic soft robotic gripper driven by TSAs. (b) An example of the gripper's precision grasping and (c) power grasping. (d) The gripper is also capable of tendon-based stiffening via an antagonistic TSA. By actuating both TSAs in the finger, the silicone will compress and stiffen. \textcolor{red}{The bending angle of the finger will be determined by the net relative displacement of tendons  \#1 and \#2}.}}
    \label{fig:1}
\end{figure}
Although soft robotic grippers are promising for many applications, it is challenging to realize high-performance soft robotic grippers that are compliant, compact, low-cost, and generate sufficient force. This is mainly because most soft actuators and artificial muscles that drive existing soft robotic grippers exhibit one or more limitations \cite{tro_zhang}, such as (a) fabrication difficulty \cite{Shih2017,Sun2021, Hasel_2021}, (b) high power requirement \cite{Hasel_2021, dea_control_17}, (c) slow actuation \cite{SMA_Hand2015, SMA_Octopus2012, michael2017}, or (d) insufficient force generation {\cite{Tang_2019, lauDEA2017}}. A twisted string actuator (TSA) is a mechanism that consists of at least two strings connected to an electric motor at one end and a load at the other end of the strings \cite{Model_Tmech14}. As shown in Fig. \ref{fig:1}(a), actuation is realized by twisting the strings with a motor to shorten the strings' length and linearly displace the attached load \cite{Model_Tmech14}. TSAs typically generate strains of 30--40\% of their untwisted length, exhibit high energy efficiency of 72-80\%, and possess a power density of 0.5\,W/g \cite{Model_Tmech14, tro_zhang}. TSAs are advantageous over spooled motor tendon actuators (SMTAs) as TSAs convert rotational motion to translational motion without the use of any external mechanisms such as gears \cite{auxilio_17}. 
TSAs can also output higher force with less input torque than motors and spools \cite{impedance_iccas14, Shin_TRO_2012}. For these reasons, TSAs are more efficient than the motor and spool configuration. A brief study comparing the torque outputs of TSAs and SMTAs was conducted in {\cite{thulanijournal2021}, which demonstrated that TSAs produce significantly higher force with less input torque than SMTAs with the same motors. This allows for less-powerful and lighter motors to be used in TSAs to generate similar forces.} \cite{tro_zhang} presented a detailed quantitative comparison of TSAs with popular artificial muscles. {As evidenced by these comparisons, TSAs offer an advantageous combination of high force output and high energy efficiency while minimizing mechanical complexity and cost by allowing the use of compact, lightweight, and inexpensive motors.}

TSAs have been widely employed in multiple robotic applications, such as tensegrity robots, robotic fingers, robotic hands, and exoskeletons \cite{auxilio_17,OT4,MULLER2015207,Shisheie2015}, not only as driving mechanisms but also to achieve variable stiffness \cite{Popov2014}. However, they have only been used in one other soft robot so far \cite{DB_RK_2022}. This is likely because TSAs require motors, which are difficult to incorporate into soft structures. However, the successful applications of TSAs in exoskeletons and assistive devices \cite{6697204,thulanijournal2021,TSA_Suit_2020} demonstrate their strong promise in areas where safe interaction with humans is necessary, supporting their use in soft robots. Additionally, although the strings used in traditional TSAs are not largely stretchable, their flexibility could be very useful in soft robots. These advantages of employing TSAs to drive soft robotic structures have been explored in our previous work \cite{DB_RK_2021,DB_RK_2022}, in which the design and performance of a TSA-driven soft robotic manipulator were presented. Actuating soft robotic grippers with TSAs could present unique advantages due to TSAs' inherent properties, and could result in the realization of large DOF-robotic structures while maintaining system compactness. 

In this paper, we present a soft robotic gripper actuated by TSAs (Fig. \ref{fig:1}(b)--(d)). Firstly, the design of the proposed robotic gripper is presented. The design is inspired by the human hand, consisting of four fingers and a thumb. The TSAs enable six independently controllable motions --- one for each of the four fingers and two for the thumb. Each finger bends due to the contraction of its corresponding TSA, whereas the thumb can both bend and roll using two TSAs. It is widely reported that the thumb is responsible for more than 50\% of a human hand's gripping capabilities \cite{Zhou2019,Deimal2016}. Therefore, the 2-DOF thumb is highly desirable to allow the gripper to replicate more anthropomorphic grasps and perform in-hand manipulation. Each finger, including the thumb, is also equipped with an antagonistic TSA which can adjust the stiffness of the fingers and the thumb. Secondly, the angular positions, angular velocities, blocked force output, and the tunable stiffness of the fingers were experimentally characterized. Lastly, the powerful and dexterous grasping and some in-hand manipulation capabilities of the gripper were demonstrated using various objects.

This is the \textit{first} study on TSAs in soft robotic grippers. The main contributions of this paper are:
\begin{itemize}
    \item Development of a multi-DOF human hand-inspired soft robotic gripper driven by TSAs. The employment of TSAs resulted in a high-performance, multiple-DOF, and compact soft robotic gripper.
    \item Development of a monolithic multi-DOF soft robotic thumb driven by TSAs. The thumb enabled the proposed gripper to efficiently realize different types of grasps.
    \item Realization of TSA-based tunable stiffness in the soft fingers of the proposed robotic gripper. 
\end{itemize}

The remainder of the paper is organized as follows. Firstly, related works are discussed in Section II. Secondly, the design and fabrication procedures are discussed in Section III. Thirdly, the experimental characterization of the robotic gripper is shown in Section IV. Next, dexterity, grasping, and in-hand manipulation experiments, as well as a comparison of our gripper to other similar grippers, are presented in Section V. Limitations of the current design with potential solutions for future iterations are discussed in Section VI. Lastly, concluding remarks and discussions on future work are presented in Section VII.
\section{Related Work}
\textcolor{red}{TSAs have been used in anthropomorphic robotic grippers in past literature \cite{DEXMART,UCHand,TSAHandRAM2013}. Among these notable works is the gripper presented in \cite{TSAHand2010} which utilized under-actuated fingers driven by TSAs. This was one of the first works to utilize TSAs in anthropomorphic robotic grippers. Similarly, the DEXMART robotic hand, with sixteen controllable DOFs, achieved complex grasps using different hand configurations \cite{DEXMART}. The robotic hand presented in \cite{TSAHandRAL2017}, while not possessing many DOFs, still achieved the desired biomimetic actuation. However, none of the previously developed robotic grippers were fabricated using completely soft material. Further, most existing TSA-driven anthropomorphic grippers did not possess the high level of dexterity required to perform complex manipulation tasks such as in-hand manipulation or achieve many of the possible human grasps \cite{TSAHand2010,TSAHand2013,TSAHandRAL2017}. These limitations with previous employments of TSAs to drive anthropomorphic traditional robotic grippers motivated further exploration into their application in soft robotic grasping and manipulation.}

Since research on soft robotic grippers spans over five decades of work, brief discussions on the most relevant studies will be presented in this section. For a detailed review on soft grippers, readers are directed to \cite{SoftGripperReview}. The current research on the design and fabrication of soft robotic grippers has mainly focused on (1) developing grippers with multiple DOF and (2) developing grippers with varying levels of stiffness \cite{SoftGripperReview}. Both of the aforementioned functionalities are highly desirable but are challenging to realize. This is mainly because most soft actuators and artificial muscles that drive existing soft robotic grippers exhibit significant limitations \cite{tro_zhang}.

For example, SMTAs have been used to drive high-performance soft grippers and realize tunable stiffness \cite{Zhou2019,SoftGripperReview,Shiva2016,Stilli2014,Maghooa2015}. However, due to their low force outputs, SMTAs require high-torque motors or high-reduction gearboxes \cite{thulanijournal2021,DB_RK_2022}. High-torque motors and high-reduction gearboxes both add extra mass to the system.

Similarly, dielectric elastomer actuators (DEAs) and hydraulically amplified self-healing electrostatic (HASEL) actuators generate sufficient actuation but require complicated and expensive fabrication procedures \cite{Hasel_2021,dea_control_17}. DEAs and HASEL actuators also require high voltage power supplies, which may be difficult to include in a compact form factor. Lastly, thermally-actuated SMAs and SCPs have driven soft robotic grippers and been used for tunable stiffness \cite{SMA_Hand2015, SMA_Octopus2012,spie15_variable,icra19_scp}. However, these thermal actuators have low bandwidth and low force output. Their high temperatures may also be dangerous to the soft robotic devices in which they are embedded. 

Pneumatic actuators, by exhibiting appreciable strain and force generation, have been widely adopted to drive soft robotic grippers and enable tunable stiffness \cite{SoftGripperReview, tro_zhang,Yafeng2021}. However, pneumatic actuators require bulky pumps or compressors \cite{Shih2017, Deimal2016}. Further, in pneumatically-actuated soft grippers, additional actuators would increase the degrees of freedom (DOFs) of the gripper but also increase its size and weight \cite{SoftGripperReview, Deimal2016}. This is because of the numerous compressors or pumps which are necessary for pneumatic actuators. Although some studies have demonstrated the use of pneumatic actuators to drive multi-DOF soft grippers \cite{Shih2017,Bhatt2021,Jianshu2018,Jianshu2019,Deimal2016}, the extra equipment outside the grippers quickly becomes large and heavy, and therefore difficult to incorporate in mobile robotic grippers. 

In terms of the design, many previous studies have used three- or four-finger designs \cite{SoftGripperReview,Yafeng2021}. These designs have fingers that are uniformly spaced around a circular base. Grasping is achieved by either actuating all fingers simultaneously or separately \cite{Shih2017,Shunya2019,Amir2017,Charbel2019}. Although this design has been effective, it is incapable of realizing many grasp types required to hold different types of objects. Another common strategy to realize soft grippers is to adopt a design inspired by a human hand \cite{SoftGripperReview}. The functionalities of anthropomorphic soft grippers are greater in comparison to other designs, especially when they include a multi-DOF thumb \cite{Jianshu2018,Jianshu2019,Zhou2019,Bhatt2021}. %
Existing designs of anthropomorphic soft grippers predominantly use pneumatic actuators and SMTAs. Therefore, realizing an anthropomorphic soft gripper with high DOFs is often challenging because the additional actuators considerably increase the volume and weight of the gripper. However, the full potential of an anthropomorphic soft design could be realized when compact and high-force actuators, such as TSAs, are used. In this paper, a compact soft gripper capable of dexterous manipulation is presented. The proposed gripper, which is driven by TSAs, demonstrates a high degree of dexterity, as it was able to achieve 31 of the 33 grasps from the Feix GRASP taxonomy \cite{Feix2016} and performed basic in-hand manipulation actions.

\section{Design and Fabrication}
\subsection{Design}
\begin{figure}
    \centering
    \includegraphics[height=6.5cm]{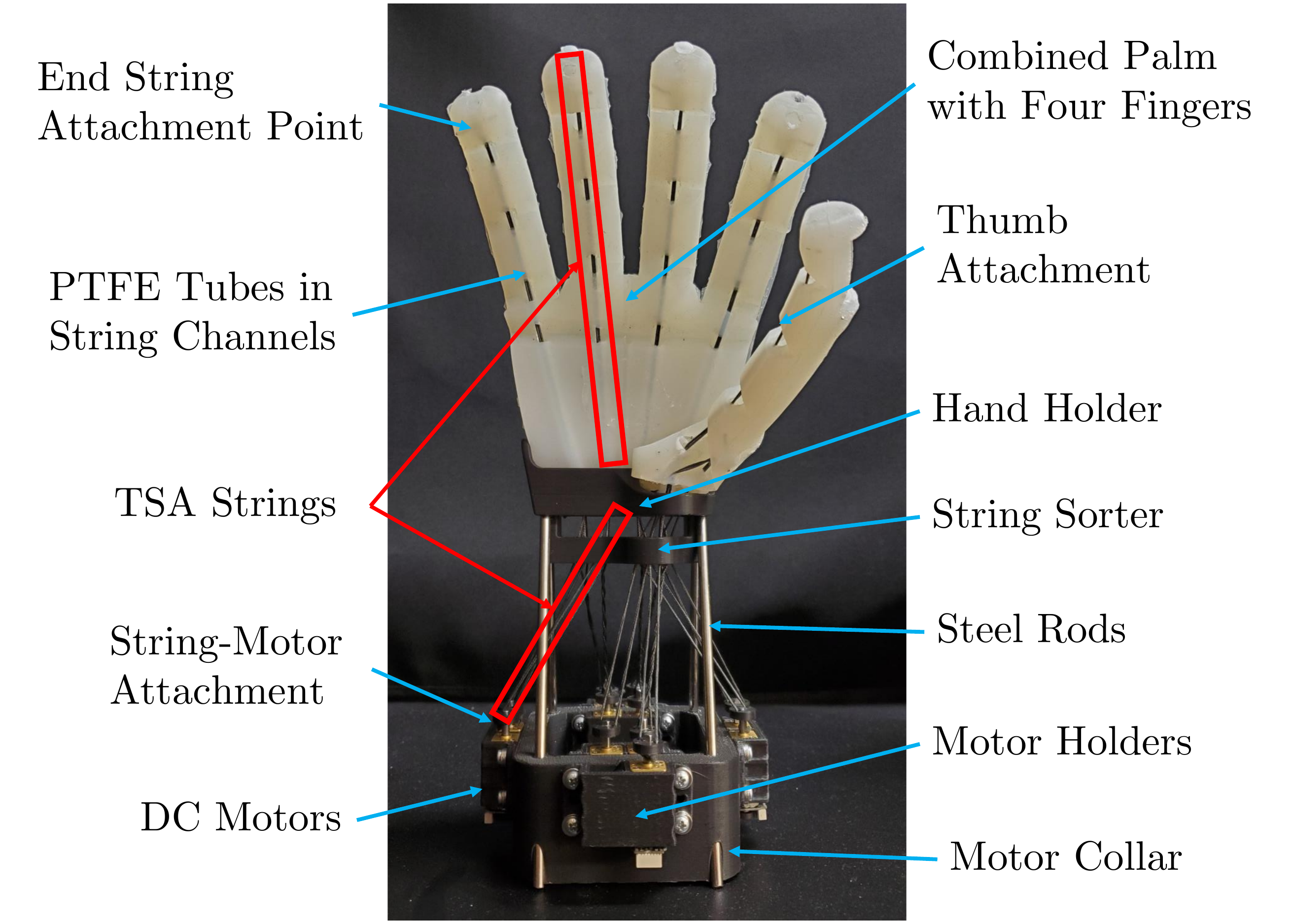}
    \hfill
    \caption{\textcolor{red}{Prototype of the proposed gripper with important parts labeled. }}
    \label{fig:Gripperphoto}
\end{figure}

\textcolor{red}{The gripper consisted of a soft monolithic palm with four fingers, a soft monolithic two-actuator thumb attachment that interfaced with the palm structure, and a rigid base that housed the motors and routed the strings as required to achieve the desired actuation. The joint locations, their range of motion, and the general sizing of the gripper emulated the human hand. The gripper prototype with important parts labeled is shown in Fig. \ref{fig:Gripperphoto}.}
\begin{figure*}[h!]
    \centering
    \includegraphics[height=9cm]{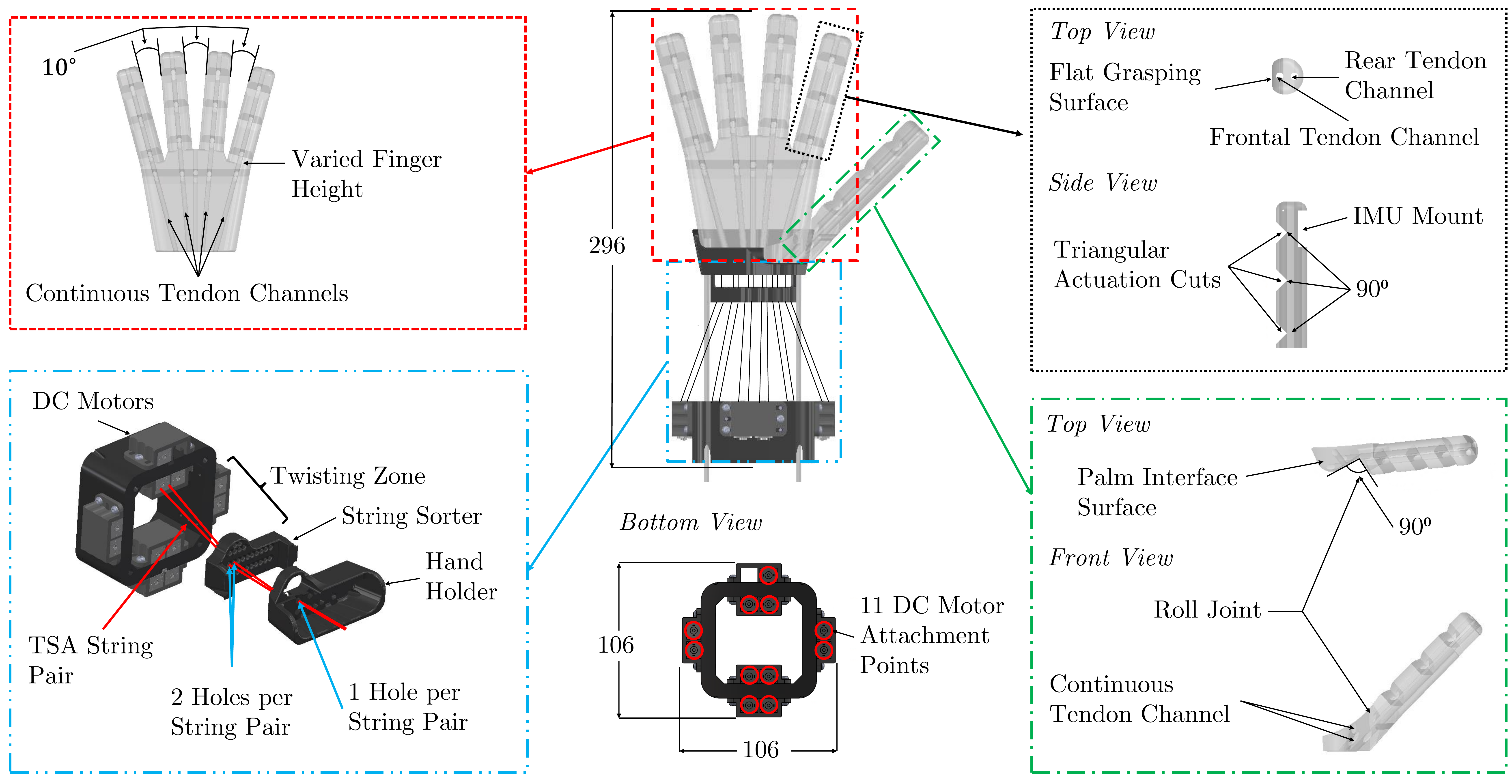}
    \hfill
    \caption{\textcolor{red}{Computer-drafted model of gripper with major components of the hand, base, and actuators shown in detail. The center top figure depicts the full assembly of the gripper. The top left, top right, bottom left, and bottom right figures show the detailed depictions of the palm, finger, base with the sorting mechanism, and the thumb, respectively. The center bottom figure shows the locations (highlighted in red) which housed the eleven motors of the TSAs.}}
    \label{fig:CADdiagram}
\end{figure*}

\textcolor{red}{Each finger of the gripper, including the thumb, was identical beyond its interface with the greater palm structure. This ensured each finger had the same actuation behavior. For the thumb, this meant a design that deviated from the human thumb, featuring an additional joint and a longer overall protrusion. Although anthropomorphism was considered for the design, achieving greater dexterity with minimal controllable DOFs was the higher priority. This thumb was dexterous (as revealed in Section V.A) and also helped the gripper achieve a high number of Feix grasps (as revealed in Section V.B). Therefore, this design was used for this version of the gripper. In future work, modifications can be made to more closely mimic a human hand.}

\textcolor{red}{The finger design is shown on top right figure in Fig. \ref{fig:CADdiagram}. Each finger featured a flat surface along its length planar to the palm, to increase the grasping contact area {\cite{ProstheticHand2020, Mariangela2015}}. {Each finger also featured three 90$\degree$ triangular cuts parallel to the primary axis of bending designed to localize and concentrate the bending at pseudo-joints {\cite{ProstheticHand2020, Mariangela2015}}. This allowed the fingers to mimic the human finger motion, which features one metacarpophalangeal and two interphalangeal joints \cite{act5010001}. The actuation cuts also reduced the finger's resistance to bending at those locations, therefore decreasing the required actuation force \cite{polym10080846}.} Each finger also featured two parallel tendon channels along its length to internally house the TSA strings (top view in  top right figure of Fig. \ref{fig:CADdiagram}). The frontal tendon was used to primarily bend the finger while the rear tendon was used to counteract the actuation of the frontal tendon and enable adjustable stiffness.}

\textcolor{red}{The palm structure of the soft gripper was designed to provide a large, flat surface to aid in grasping. The finger interfaces were angled 10$\degree$ relative to one another (top left of figure of Fig. \ref{fig:CADdiagram}), allowing the gripper to mimic the resting splayed position of the fingers in the human hand. The finger interfaces of the first and fourth finger positions were also more inset than those of the second and third finger positions (top left of Fig. \ref{fig:CADdiagram}). The palm also had continuous tendon channels in-line with those in each finger. This allowed the tendons to pass through the palm and into the base, where they were connected to the motors (bottom left figure of Fig. \ref{fig:CADdiagram}). The palm with the primary four fingers was a monolithic structure (top left figure of Fig. \ref{fig:CADdiagram}). This avoided complex assembly procedures \cite{RBO3,Deimal2016,Jianshu2019} which could require screws and nuts that increase the gripper's weight \cite{ProstheticHand2020} and decrease its compliance. The thumb was a separate structure and its design (bottom right figure of Fig. \ref{fig:CADdiagram}) was inspired by the designs presented in \cite{Zhou2019,Jianshu2018}. It was mounted to an additional jointed section that allowed the thumb to roll across the palm. This roll joint was actuated using an additional TSA. The joint had a theoretical maximum bending angle of 90$\degree$.}

\textcolor{red}{The fingers, palm, and thumb structures were cast from Dragon Skin\texttrademark 20 silicone rubber. Reusable molds for the combined palm with four fingers and separate thumb structures were designed and 3D-printed using acrylonitrile butadiene styrene (ABS) plastic. Once the parts were cast, the two  components of the hand were glued together using silicone rubber adhesive (SIL-Poxy, Smooth-On).}

\textcolor{red}{The base for the gripper (bottom left figure in Fig. \ref{fig:CADdiagram}) consisted of a holder piece that interfaced with the proximal end of the hand to hold it in place, a sorter mechanism that constrained the twisting of the strings, and a collar to hold the TSA motors in place. The sorter mechanism worked by threading the two strings of each TSA string pair through separate channels in the mechanism, creating a constant twisting zone between the motors and sorter mechanism that prevented the strings from twisting within the hand. The separation between the twisted and untwisted regions is shown in Fig. \ref{fig:1}(a) \cite{Tavakoli2016}. This prevented any twisting of the strings within the silicone hand itself, which could result in friction between the strings and the inner surface of the tendon channels \cite{suthar_conduit_2018,palli_sliding_2016}. Steel rods held the collar away from the sorter, creating a twisting region large enough to provide sufficient actuation. The motor collar featured attachment points for the eleven motors used in the TSAs. Overall, the gripper had a footprint of 106\,mm $\times$ 106\,mm and a total length of 295\,mm. The full assembly of the gripper is presented in the top center figure of Fig. \ref{fig:CADdiagram}}

\textcolor{red}{The components of the base were 3D-printed similarly to the molds used to cast the hand parts. Once the gripper was assembled, to configure the TSA actuators, the string pairs were routed from the motors in the base, through the sorter, hand holder, and the tendon channels in the palm and fingers. The strings were pinned at the fingertips and tied to be slightly taut when unactuated. This helped the gripper maintain a similar neutral position in different base orientations. All the channels in the palm and the fingers were equipped with polytetrafluoroethylene (PTFE) tubes, which both reduced the frictional force experienced between the strings and silicone and reinforced the soft silicone structure. The tendons used in this work were 0.7-mm-diameter ultra-high-molecular-weight polyethylene (UHMWPE) strings \cite{Model_Tmech14,ModelControl_Tmech13,DB_RK_2022}. UHMWPE was selected because its frictional coefficient with PTFE is extremely low, previously reported to be 0.04--0.06 \cite{quaglini2011friction,lee2019effect}.}

\textcolor{red}{These strings in the TSA act as a rotary-to-linear gear that greatly amplify the force output compared to the same motor with a spool attached (the increased force output causes a corresponding decrease in speed). As a result, using TSAs allows the use of smaller motors in the base, decreasing the overall footprint of the base and increasing the compactness of the system. Due to the TSA geometry, the motors are also arranged parallel to the direction of linear actuation, further enhancing the ability of the gripper to be made compact relative to a system using SMTAs, which would require the motors to be arranged perpendicular to the actuation direction. The use of TSAs does necessitate the inclusion of a twisting zone and sorter mechanism, though, potentially increasing the required length of the gripper relative to one driven by SMTAs.}

\subsection{Electronics and Control}
To actuate the TSAs, the gripper used eleven brushed DC motors (Micro Metal Gearmotor HPCB 6V, Pololu; locations shown in the center bottom figure of Fig. \ref{fig:CADdiagram}) with 30:1 reduction gearboxes. Each motor weighed 9.5\,g. A magnetic Hall-effect encoder disc (Magnetic Encoder, 12CPR, Pololu) was attached to each motor to count the motor rotations for control and data acquisition. This gearbox and encoder combination enabled 360 ($30 \times 12$) counts per rotation, resulting in a 1$\degree$ ($2.78\times 10^{-3}$ rotations) sensing resolution in the motor shaft. Each of these motors could output 0.45\,kg$\cdot$cm of torque before stalling. These motors were all mounted on the motor collar of the base.

The remaining electronics used to power and control the gripper were housed outside of the robot and were wired to the motors. Each motor was controlled using a brushed DC motor driver (MAX14870, Pololu). A 32-bit ARM core microcontroller (Due, Arduino) was used for motor control, automated experimental procedures, and data acquisition. Data was automatically logged with a custom-written open-source Python script \footnote{\url{https://github.com/EmDash00/ArduinoLogger}}. Inertial measurement units (IMUs) (9-DOF Absolute Orientation IMU Fusion Breakout - BNO055, Adafruit) were mounted to each fingertip and the secondary joint of the thumb structure for characterization purposes but were removed for later grasping demonstrations. 

The motors were controlled with input voltages that were proportional to the error in the motor shaft angle. Closed-loop control using the bending angle of the finger as measured by an attached IMU was also used for the stiffness experiments, which are described in Section IV. For the grasping demonstrations presented in Section V, push-buttons and a custom-written C++ script were used to manually control the actuation of each TSA in the gripper.
\begin{figure}
    \centering
    \includegraphics[width=0.5\linewidth]{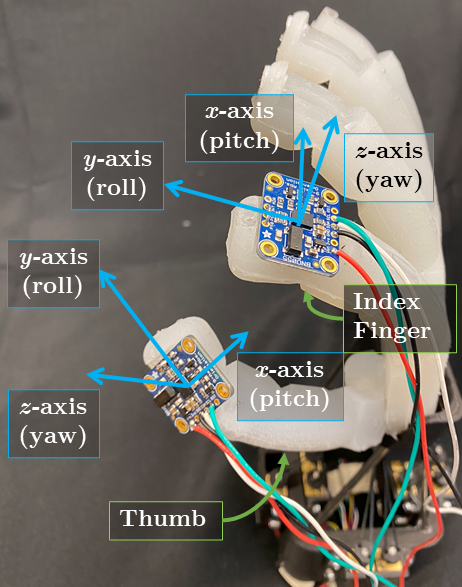}
    \caption{The locations of the inertial measurement units (IMUs) on the thumb and finger. To characterize the other three fingers, the IMUs were placed on the same corresponding locations.}
    \label{fig:imu_locations}
\end{figure}
\begin{figure}
    \centering
    \includegraphics[width=0.5\linewidth]{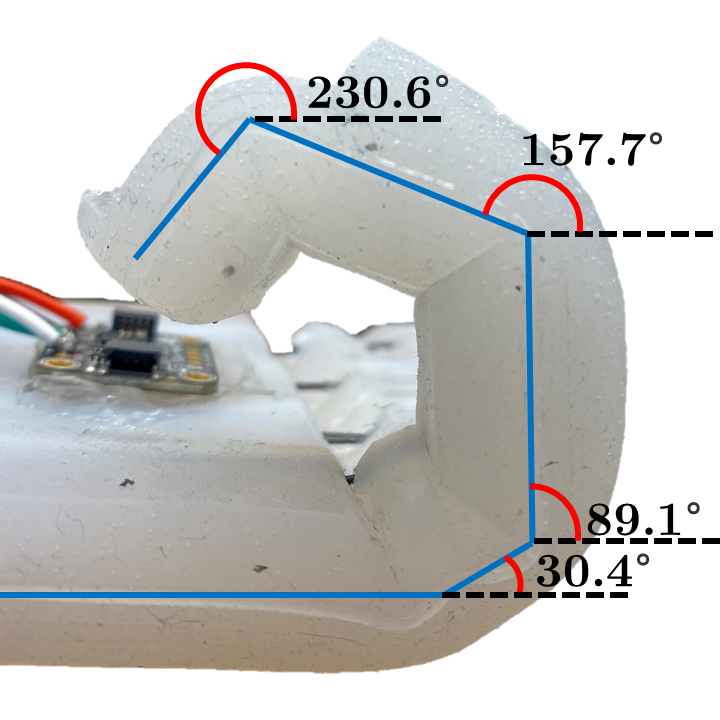}
    \caption{The maximum bending obtained with a motor having a stall torque of 0.45\,kg$\cdot$cm.}
    \label{fig:max_bending}
\end{figure}
\begin{figure*}
    \centering
    \subfloat[]{
    \includegraphics[width=0.32\linewidth]{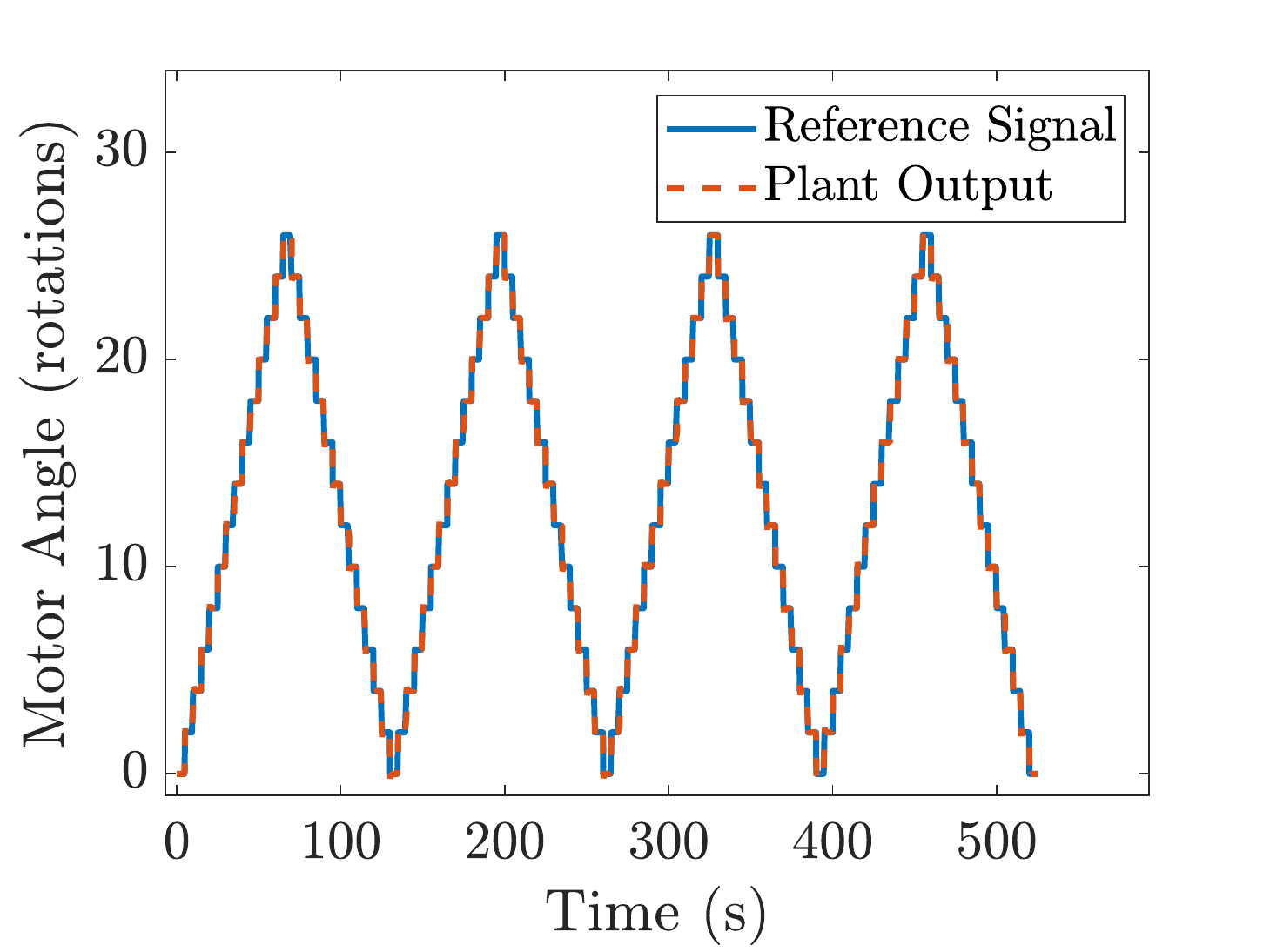}}
    \hfill
    \subfloat[]{
    \includegraphics[width=0.32\linewidth]{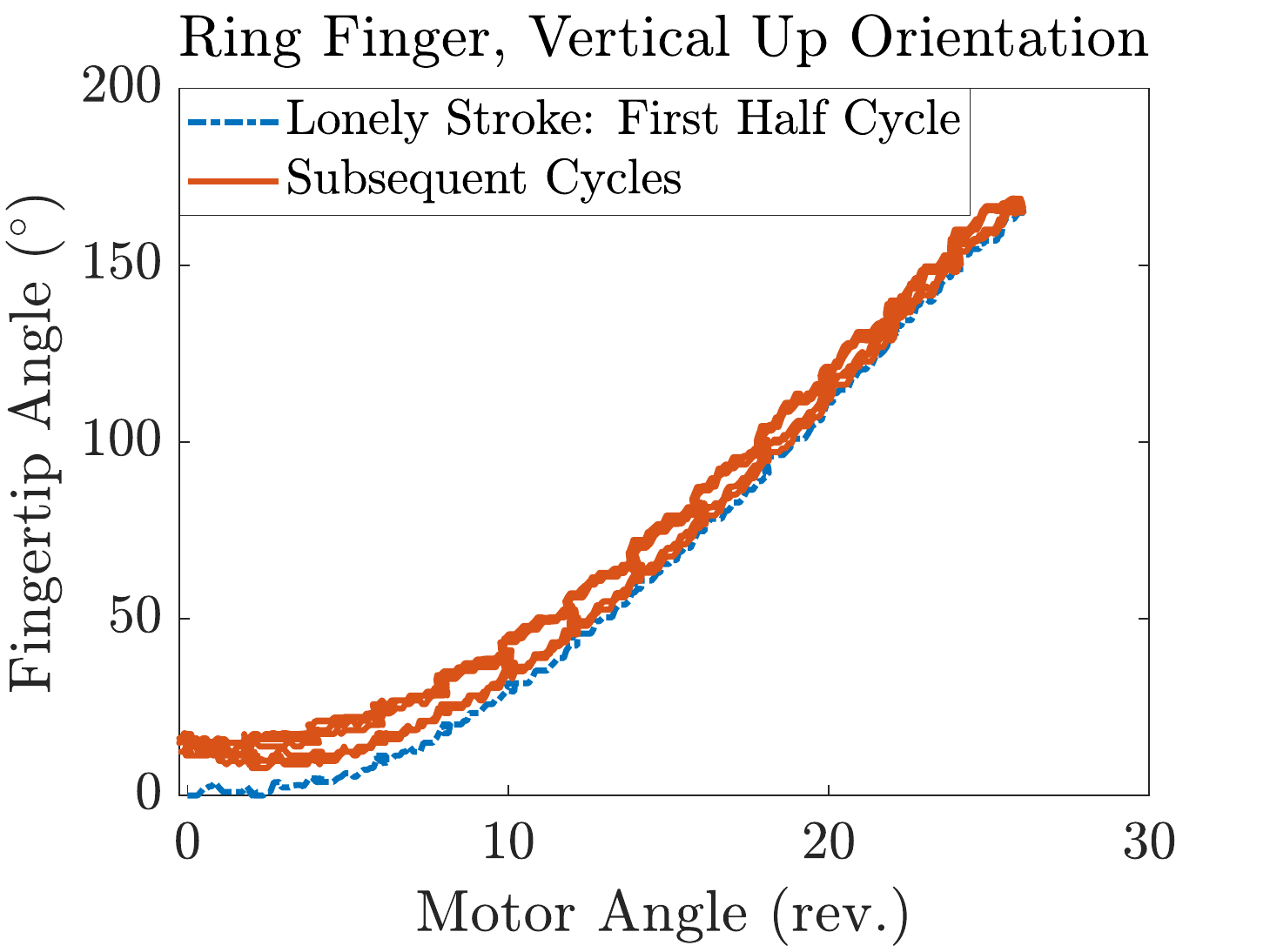}}
    \hfill
    \subfloat[]{
    \includegraphics[width=0.32\linewidth]{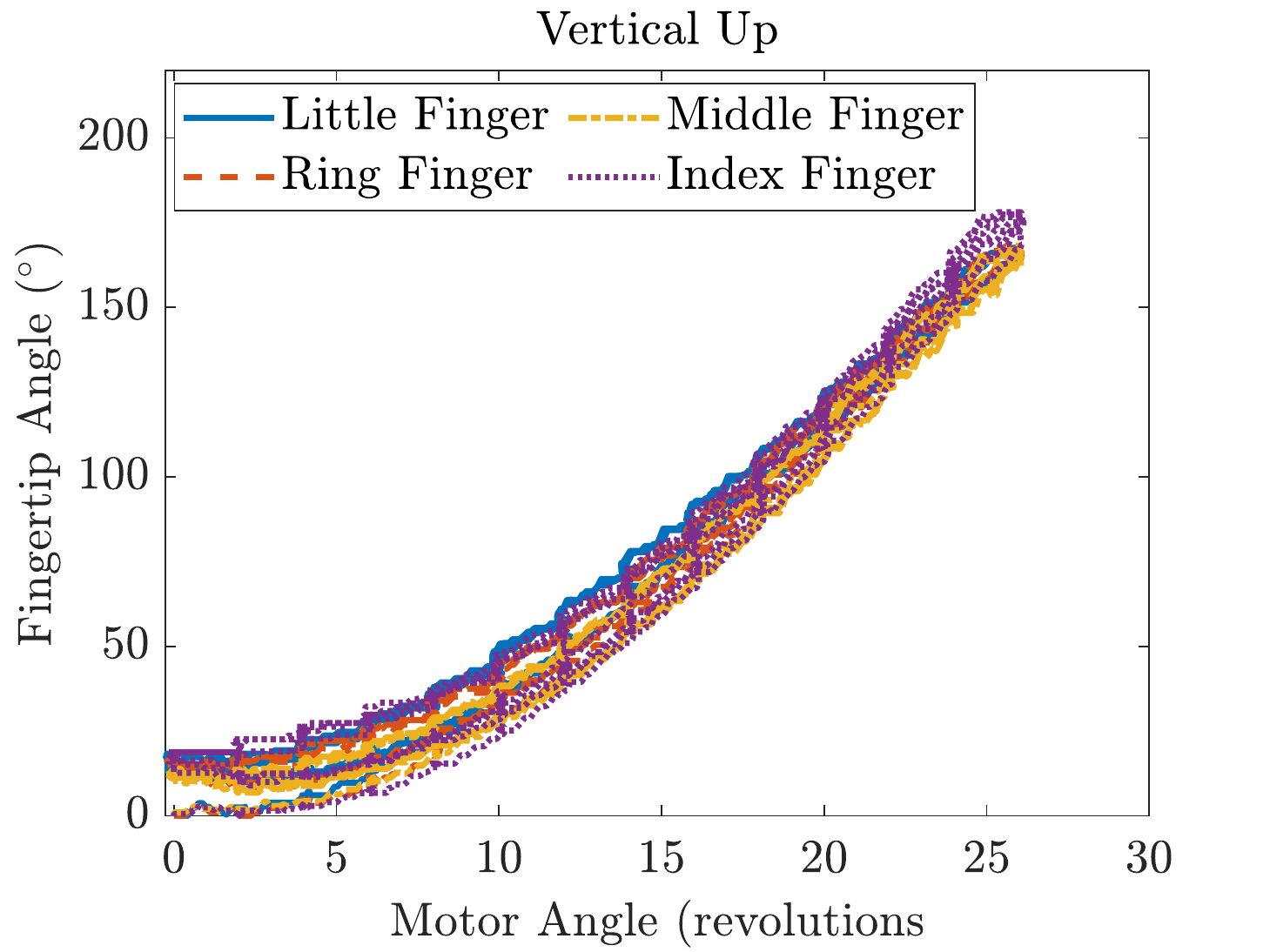}}
    \hfill
    \subfloat[]{
    \includegraphics[width=0.32\linewidth]{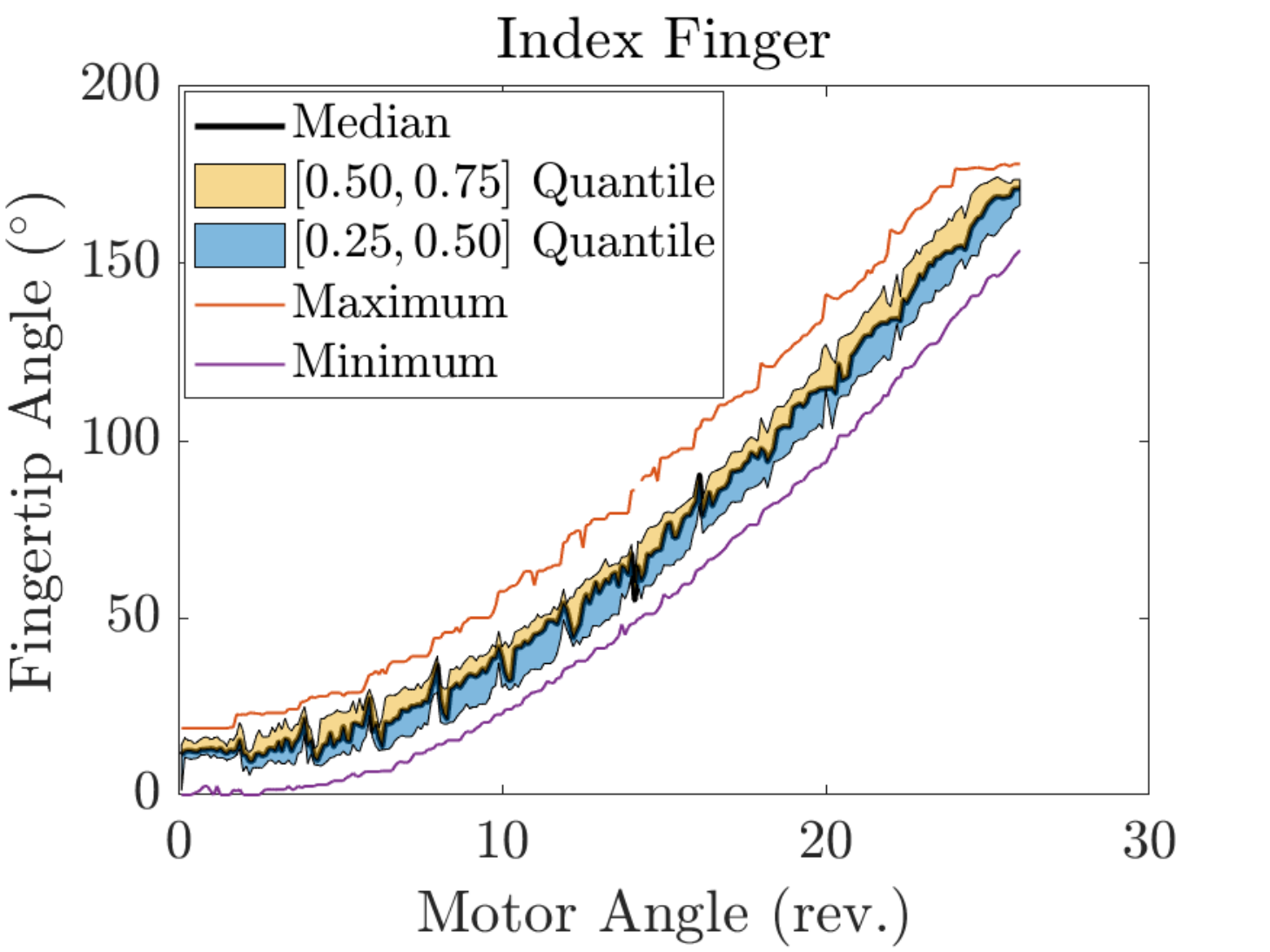}}
    \hfill
    \subfloat[]{
    \includegraphics[width=0.32\linewidth]{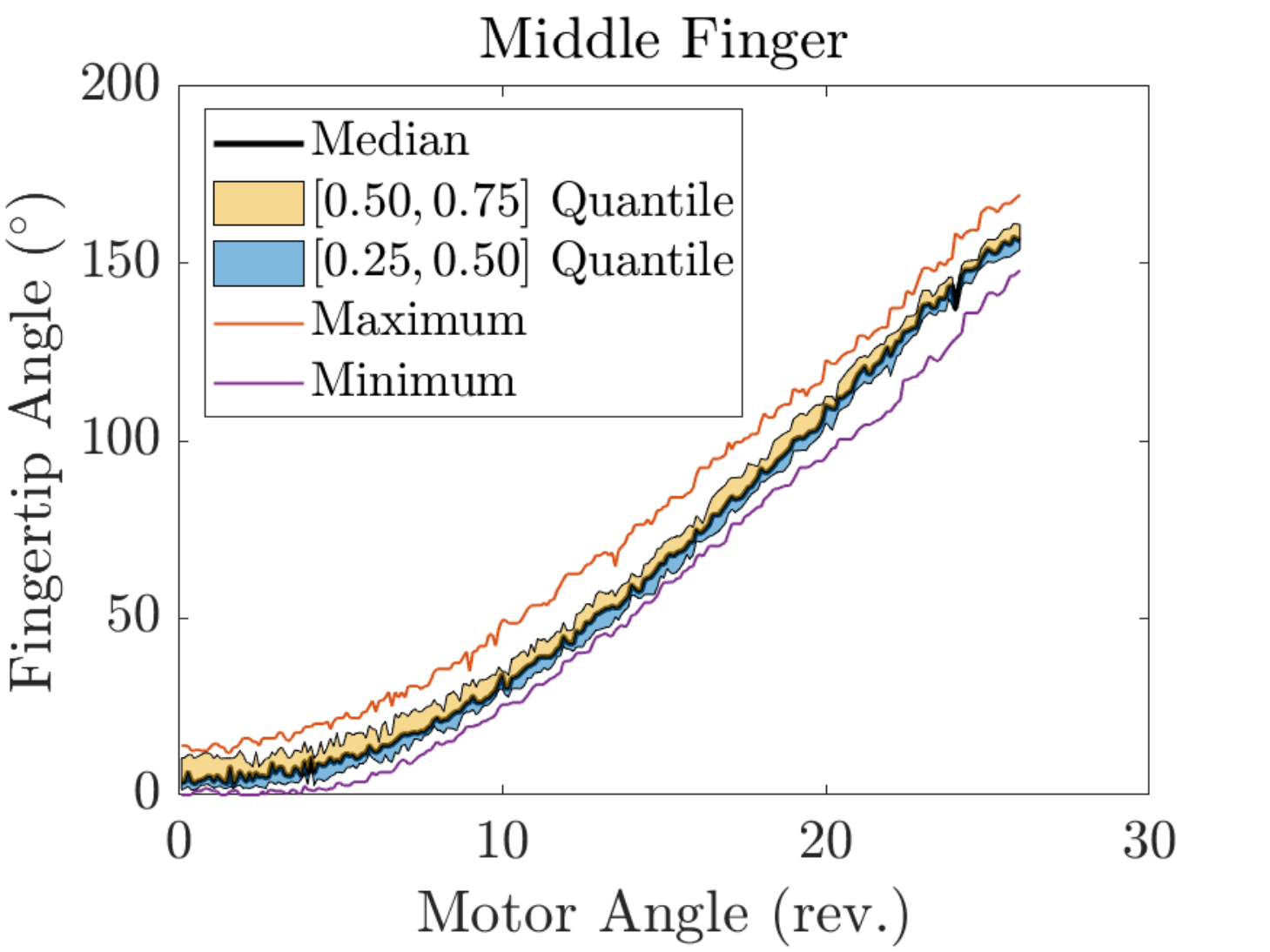}}
    \hfill
    \subfloat[]{
    \includegraphics[width=0.32\linewidth]{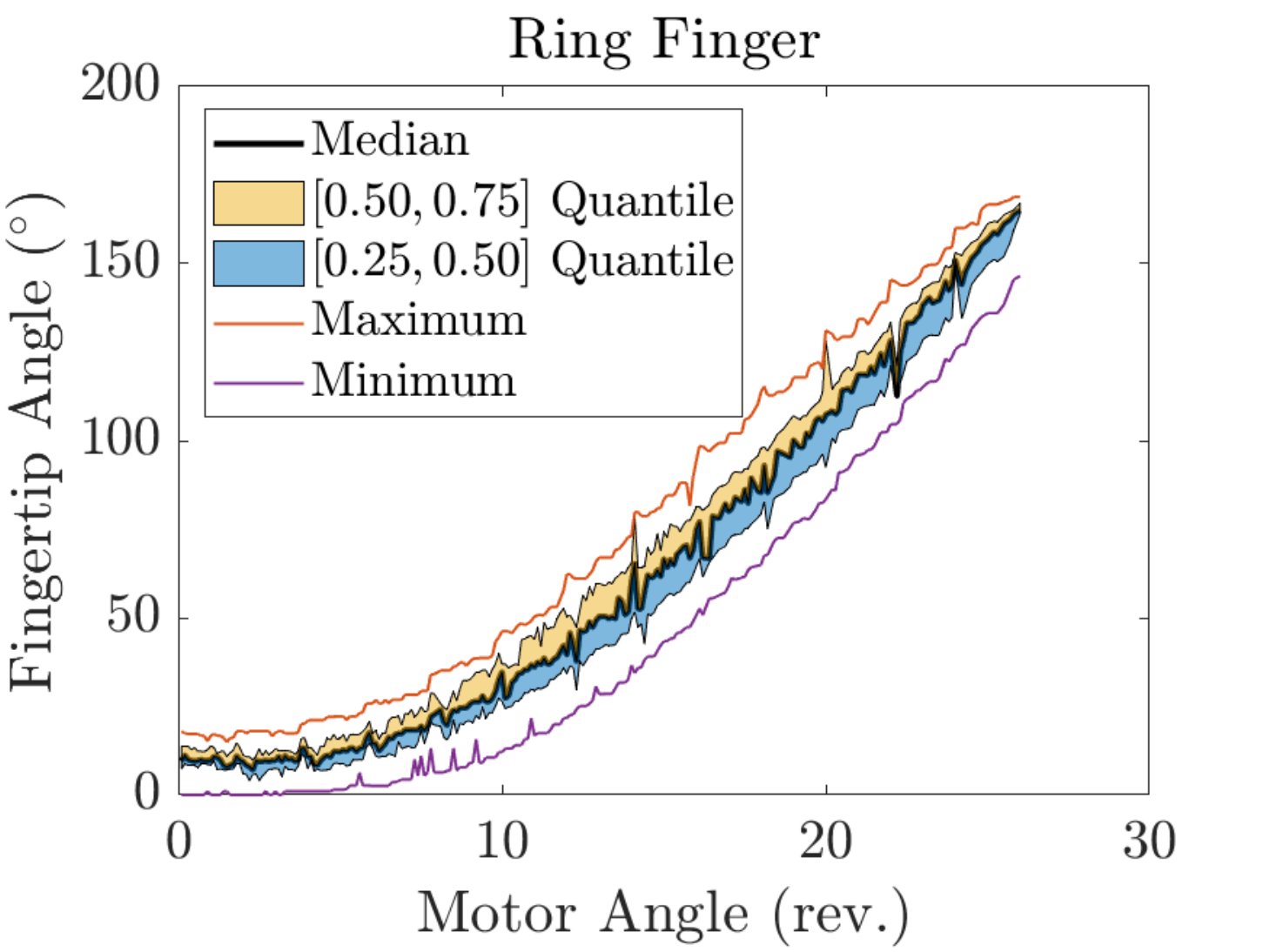}}
    \hfill
    \subfloat[]{
    \includegraphics[width=0.32\linewidth]{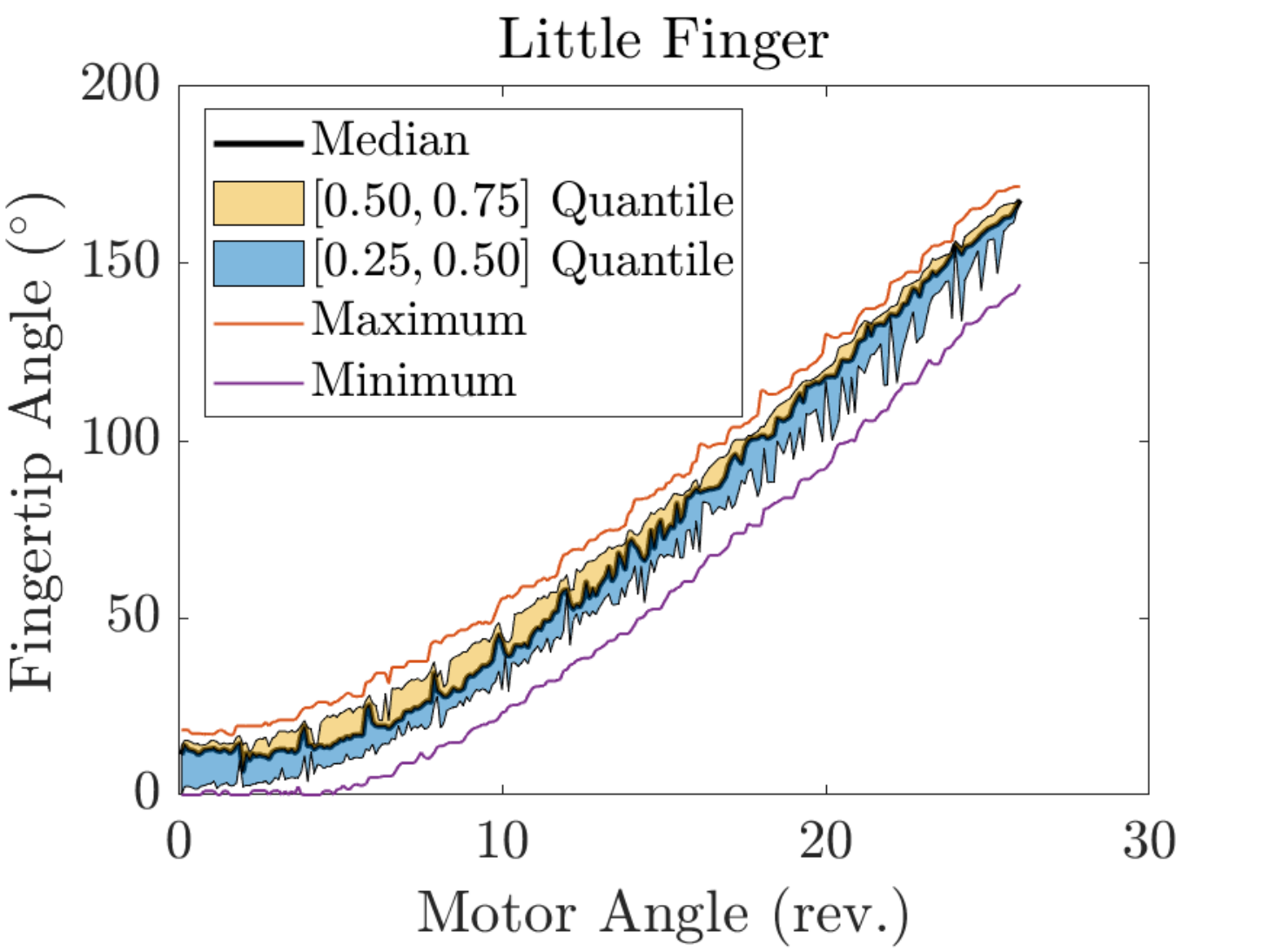}}
    \hfill
    \subfloat[]{
    \includegraphics[width=0.32\linewidth]{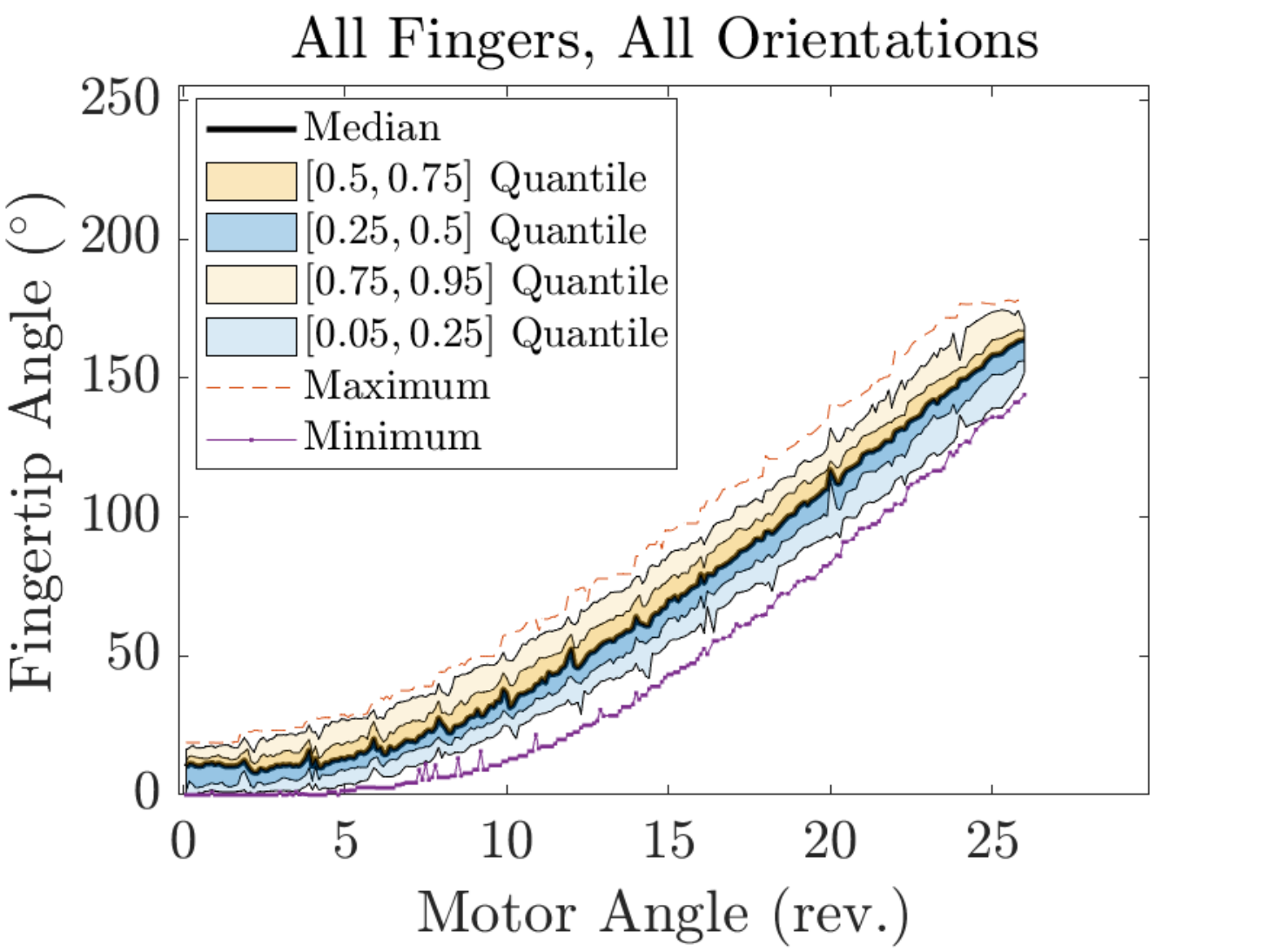}}
    \hfill
    \subfloat[]{
    \includegraphics[width=0.32\linewidth]{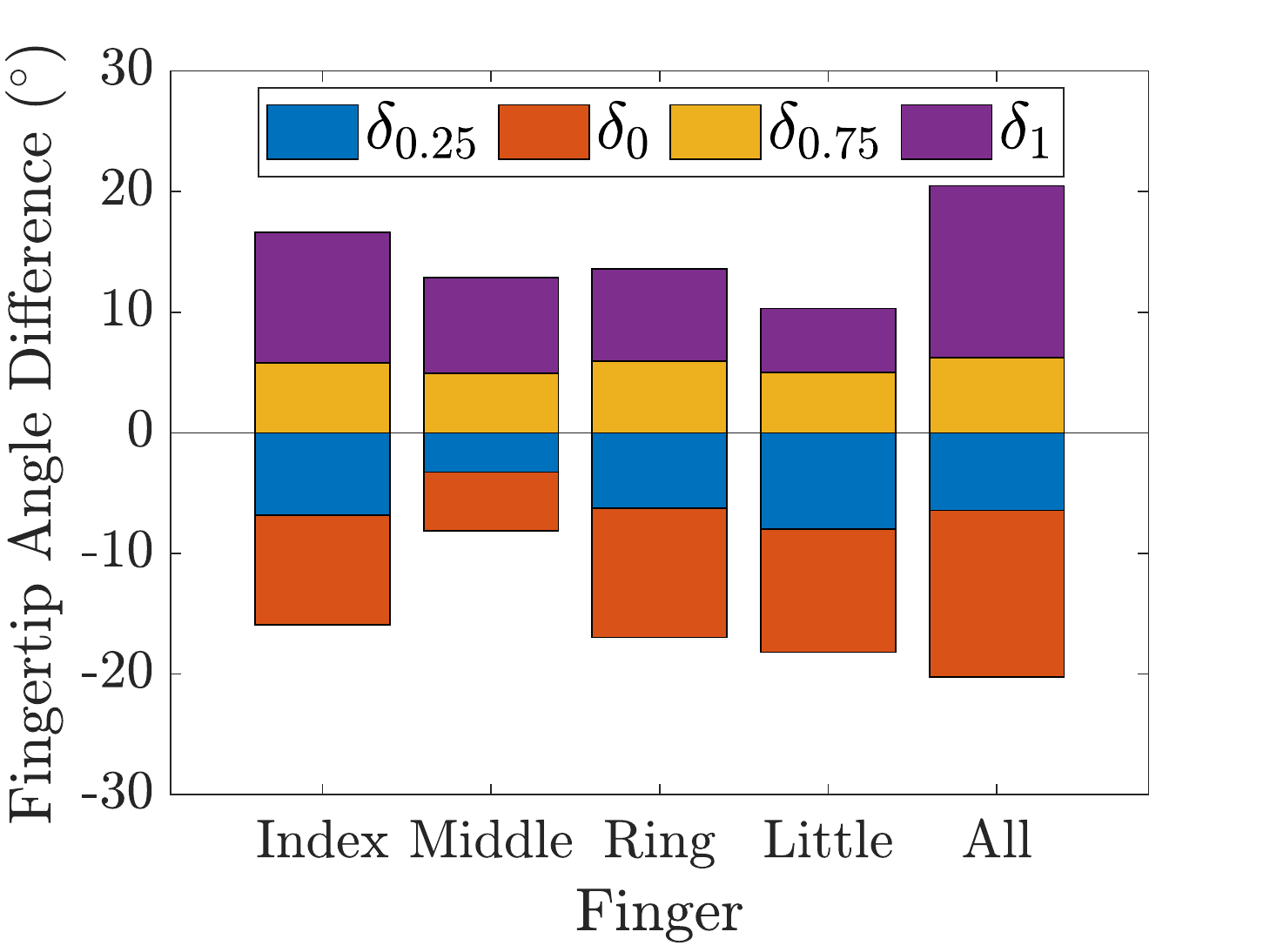}}
    \hfill
    \subfloat[]{
    \includegraphics[width=0.32\linewidth]{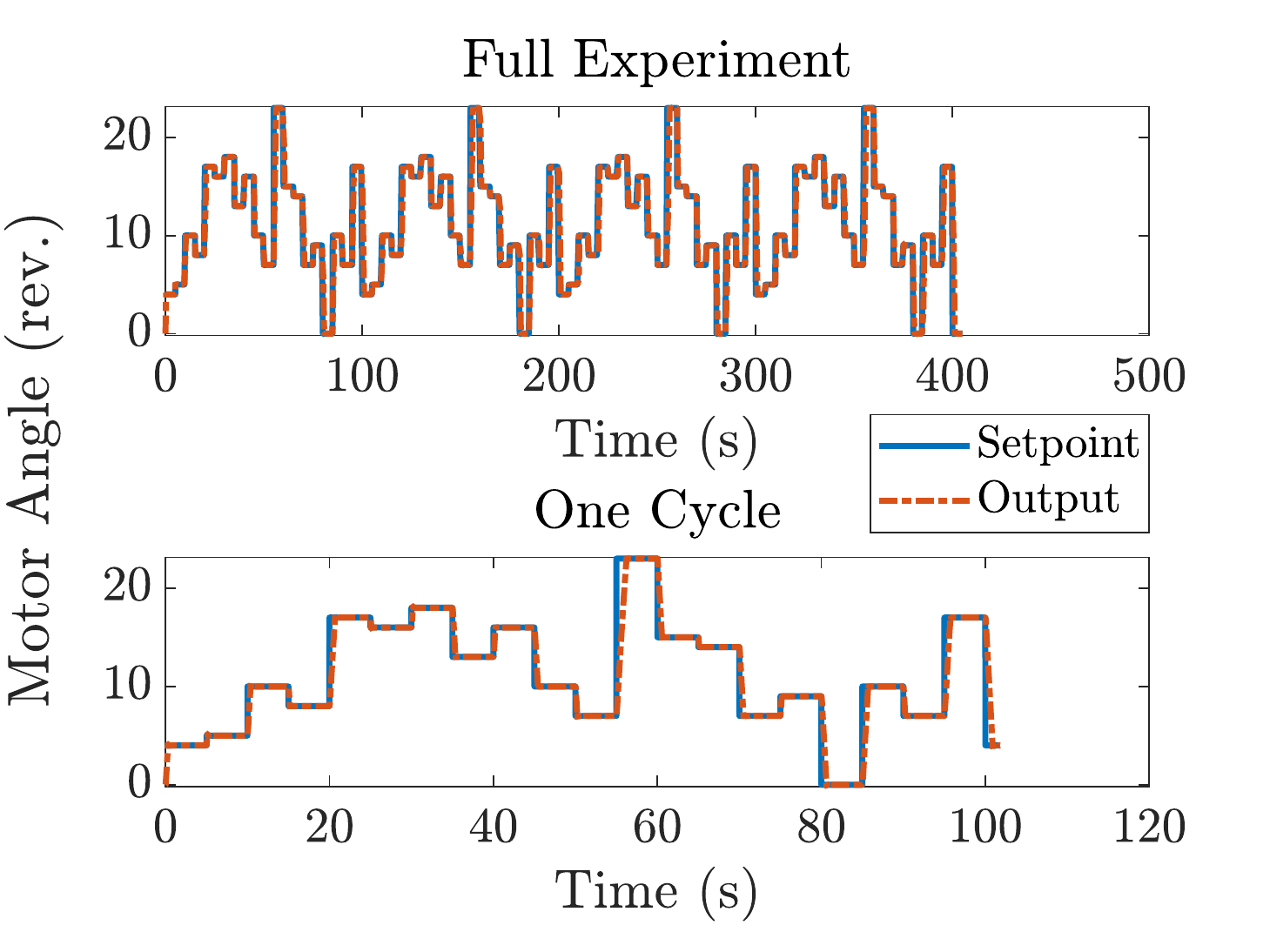}}
    \hfill
    \subfloat[]{
    \includegraphics[width=0.32\linewidth]{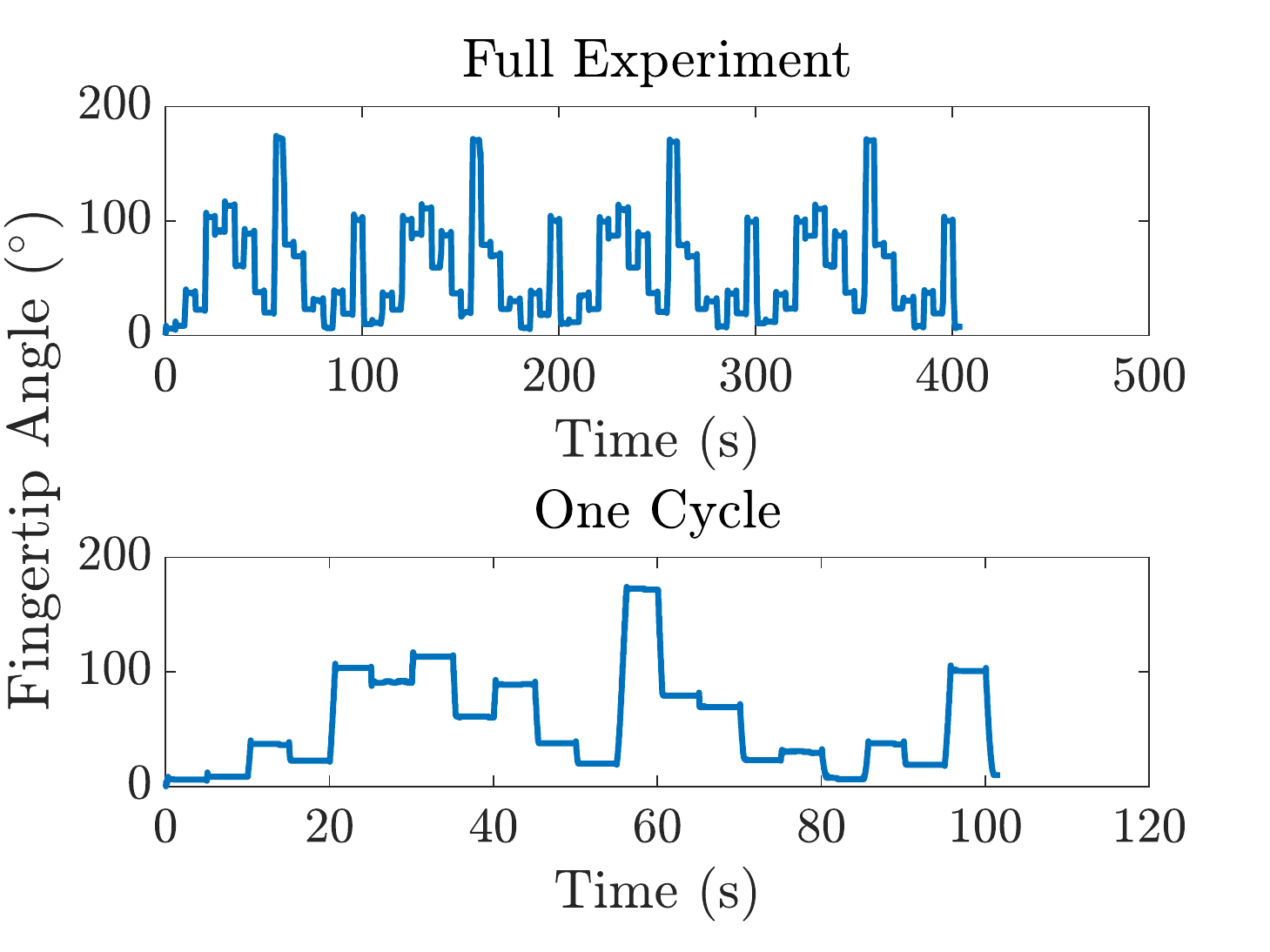}}
    \hfill
    \subfloat[]{
    \includegraphics[width=0.32\linewidth]{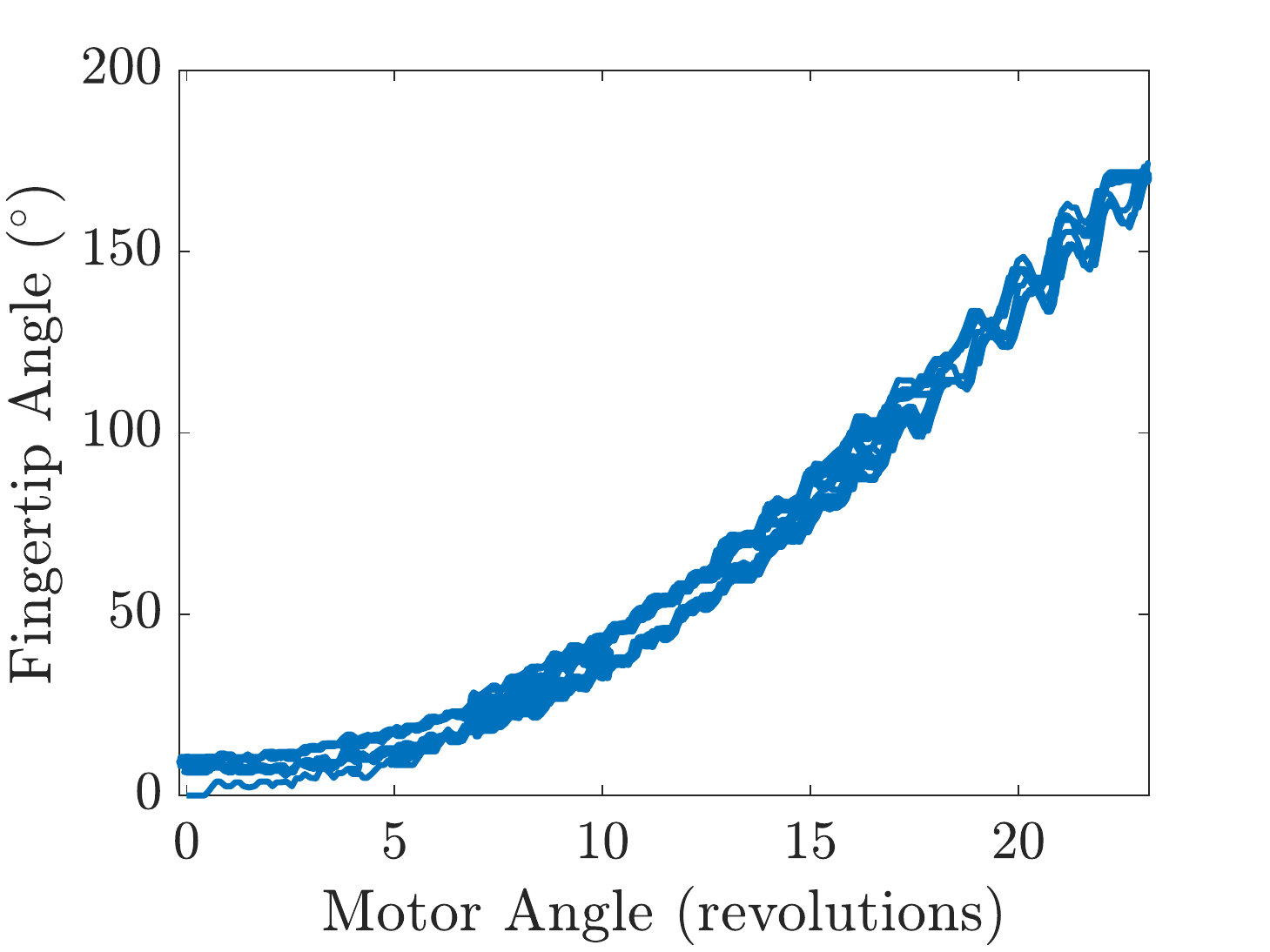}}
    \hfill
    \caption{(a) The motor angle input to the fingers that was used during experimental characterization. (b) A typical correlation between motor angle versus bending angle, with the lonely stroke highlighted. (c) The consistency of the correlations for different fingers in the ``vertically up'' position. \textcolor{red}{(d)--(h) ``Moving box-and-whisker plots'': the time-varying medians, 0.25-quantiles, 0.75-quantiles, minimums, and maximums for the fingertip angle of each finger. Univariate statistics are taken from four orientations: vertically up, vertically down, horizontally up, and horizontally down. Results are shown for the (d) index finger, (e) middle finger, (f) ring finger, (g) little finger, and (h) all fingers.} (i) Summary statistics that average the minimums, maximums, 0.25-quantiles, and 0.75-quantiles for each finger over each experiment. (j) Four cycles of 20 random step motor angle set points. (k) The fingertip angle corresponding to the motor angle input sequence presented in (j). (l) The correlation between motor angle and fingertip angle when the motor angle setpoint varied randomly.}
    \label{fig:characterization}
\end{figure*}
\section{Experimental Characterization}
\subsection{Position}
\textcolor{red}{The purpose of the characterization of the actuation of the fingers and thumb was two-fold. Firstly, the results demonstrated the range of actuation of the fingers and their ability to achieve repeated actuation in different gripper orientations (under varying effects of gravity). Secondly, the results revealed the nonlinear mapping between the strings' twists and the bending of the fingers. These results lay the foundation for further research on the modeling and control of TSA-driven soft fingers. }
\subsubsection{Fingers}
The position of the robotic hand was characterized by measuring the fingertip angles and corresponding motor rotation angles of the TSA. The IMU was mounted on the back side of the fingertip using the mount as shown in {Fig. \ref{fig:imu_locations}}. The maximum possible fingertip angle (with the particular motors in the gripper) was 230.6$\degree$, as shown in Fig. \ref{fig:max_bending}. Fig. \ref{fig:max_bending} also shows the orientations of the finger's other links. Fig. \ref{fig:characterization}(a) shows the motor rotation input sequence used to characterize the fingertips' angular positions. Four cycles of monotonically increasing and decreasing reference motor angles were set. Each setpoint was held constant for five seconds to obtain the steady-state value. The motor setpoints were increased/decreased in steps of two rotations. As shown in Fig. \ref{fig:characterization}(a), the control system for the motor was sufficiently responsive as to drive the motor's shaft to the desired setpoint in a short time relative to the time for which each setpoint was held constant (five seconds).
\begin{figure*}
    \centering
    \subfloat[]{
    \includegraphics[width=0.32\linewidth]{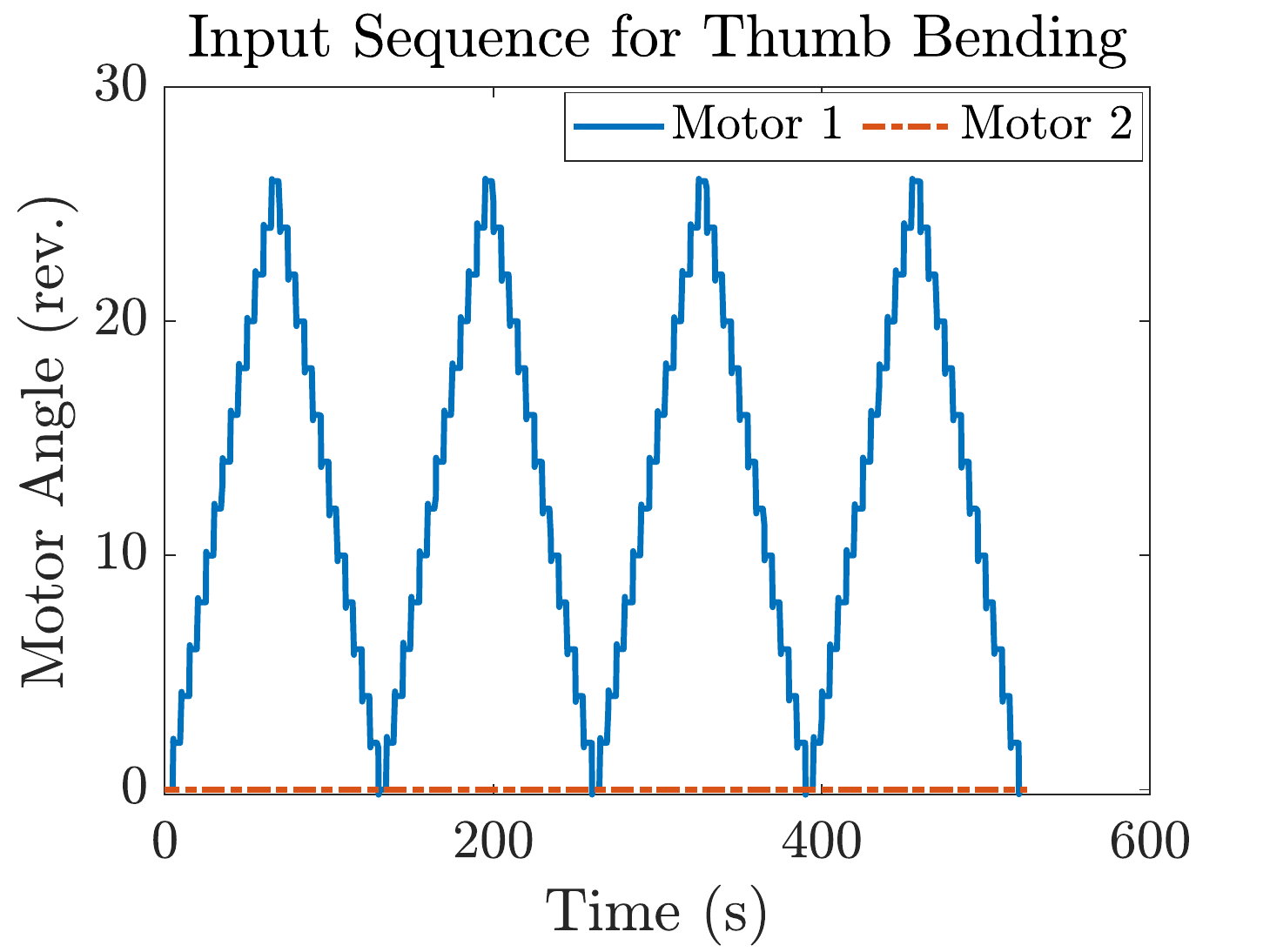}}
    \hfill
    \subfloat[]{
    \includegraphics[width=0.32\linewidth]{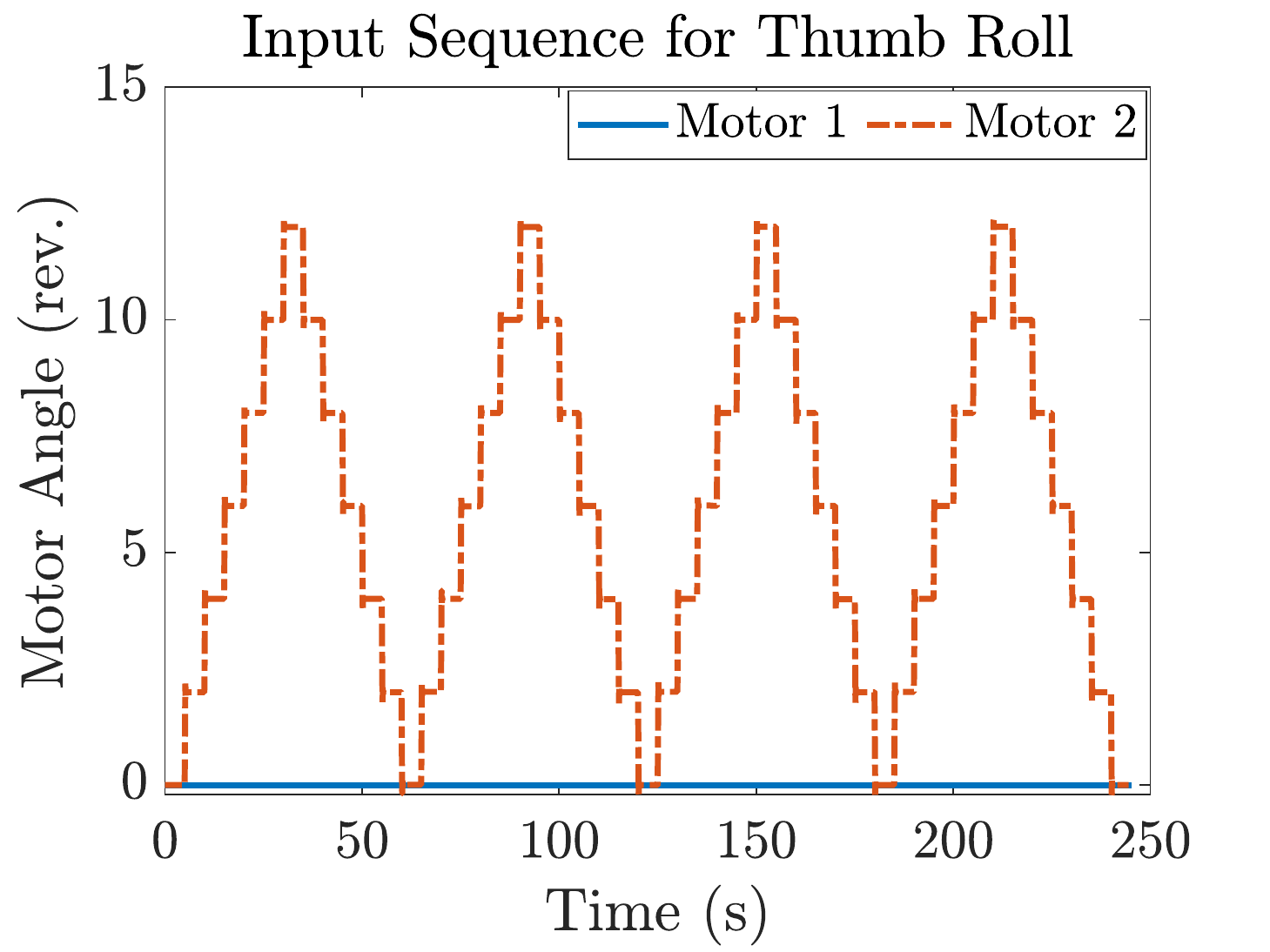}}
    \hfill
    \subfloat[]{
    \includegraphics[width=0.32\linewidth]{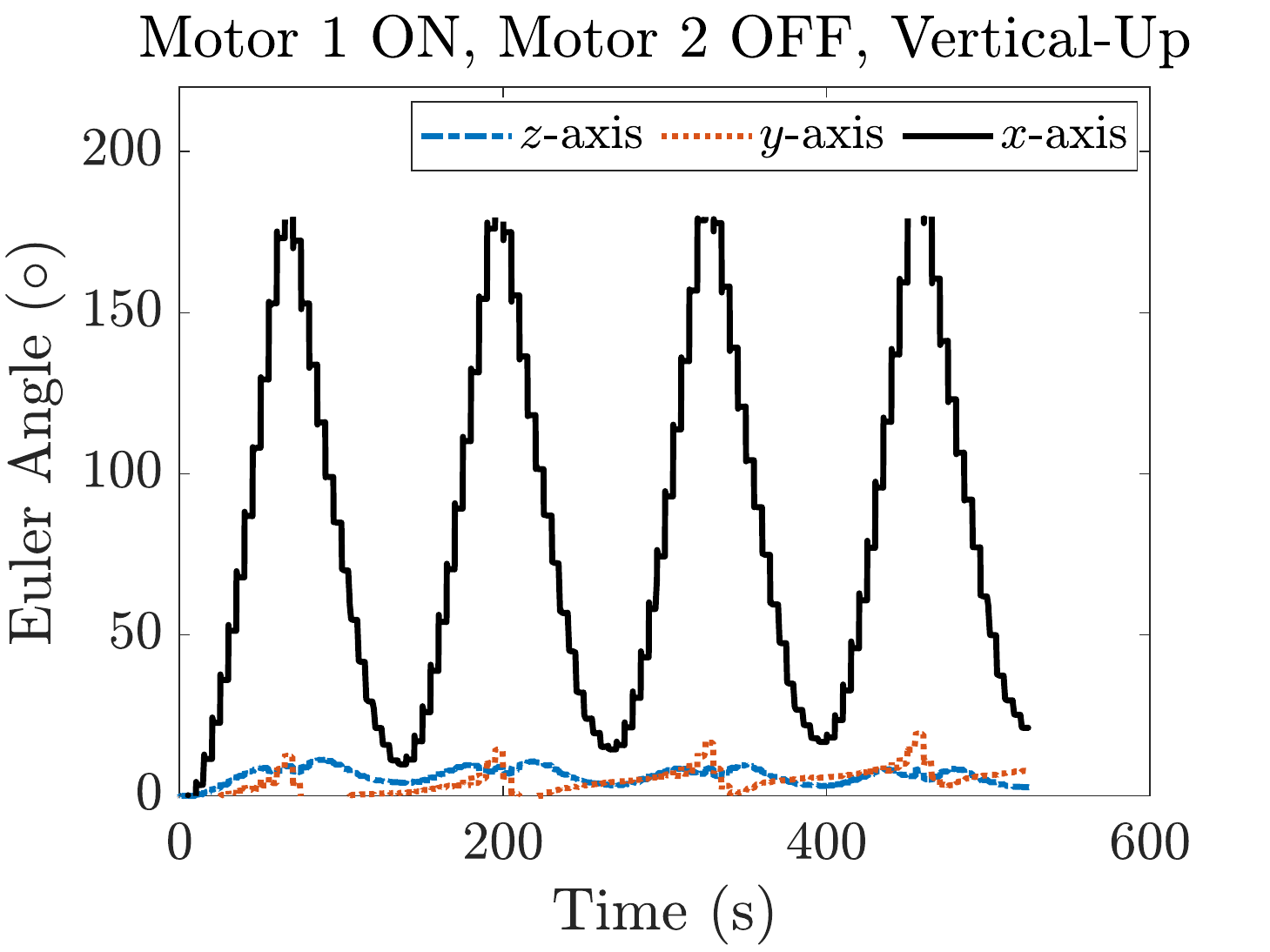}}
    \hfill
    \subfloat[]{
    \includegraphics[width=0.32\linewidth]{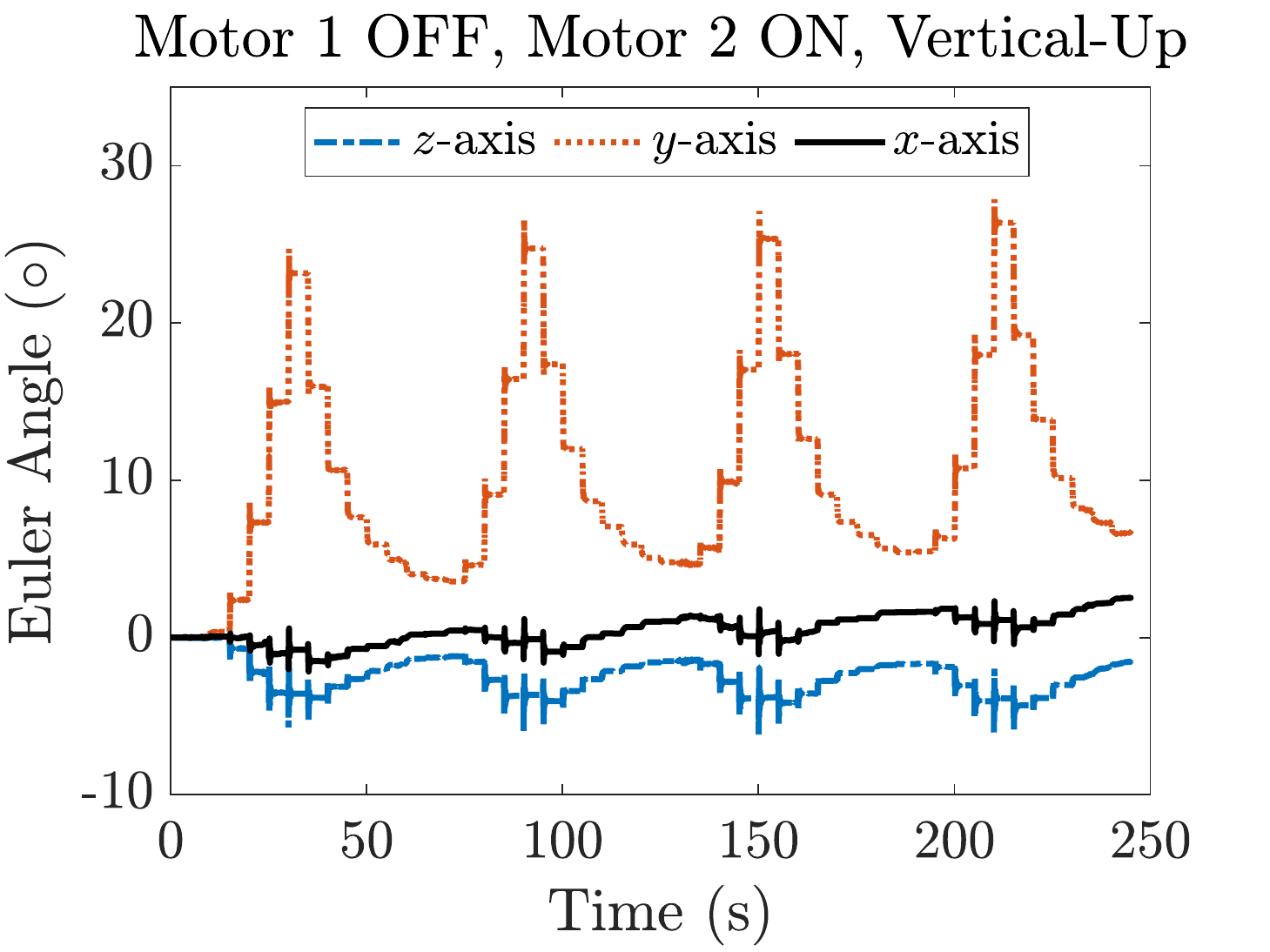}}
    \hfill
    \subfloat[]{
    \includegraphics[width=0.32\linewidth]{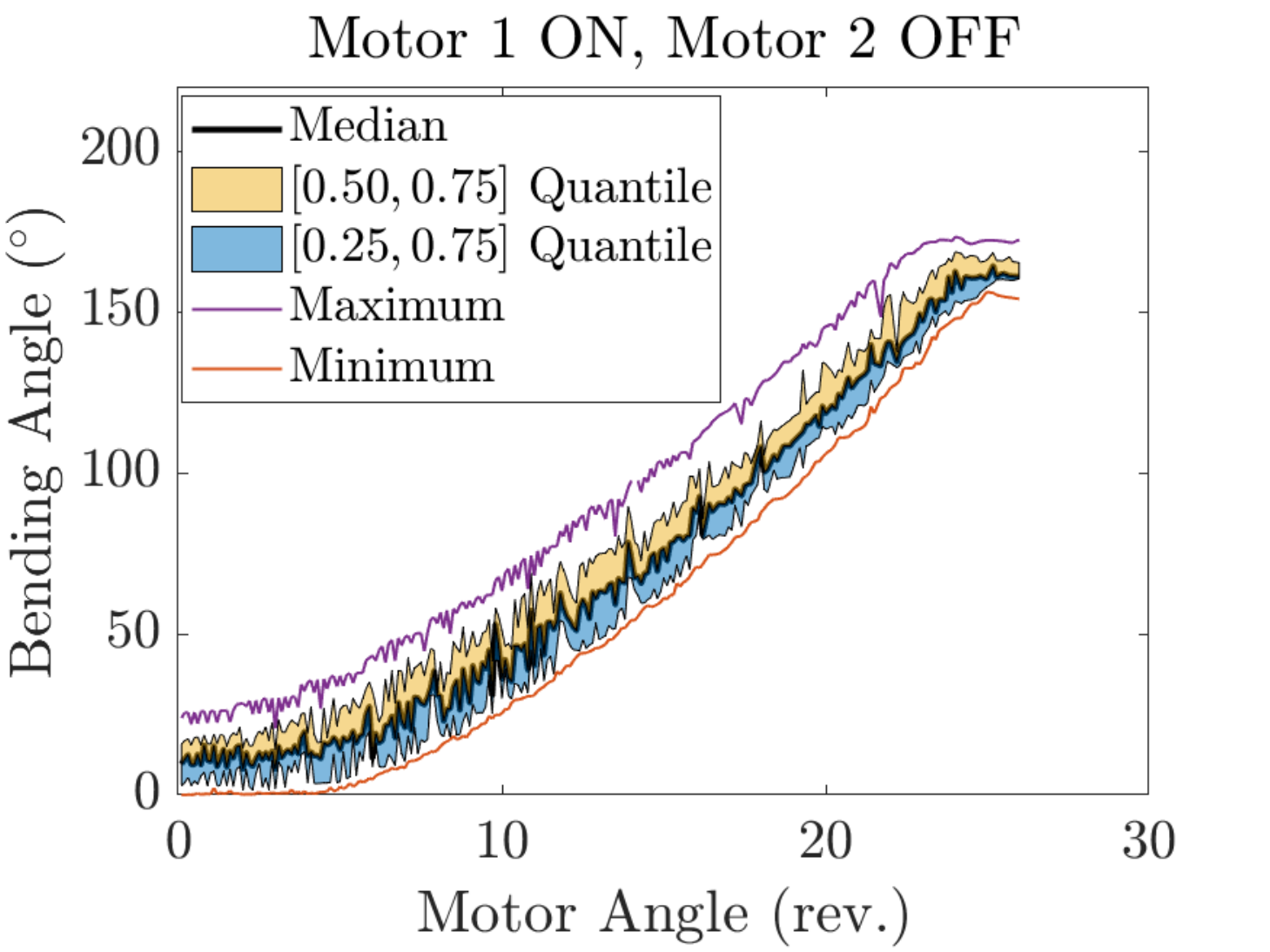}}
    \hfill
    \subfloat[]{
    \includegraphics[width=0.32\linewidth]{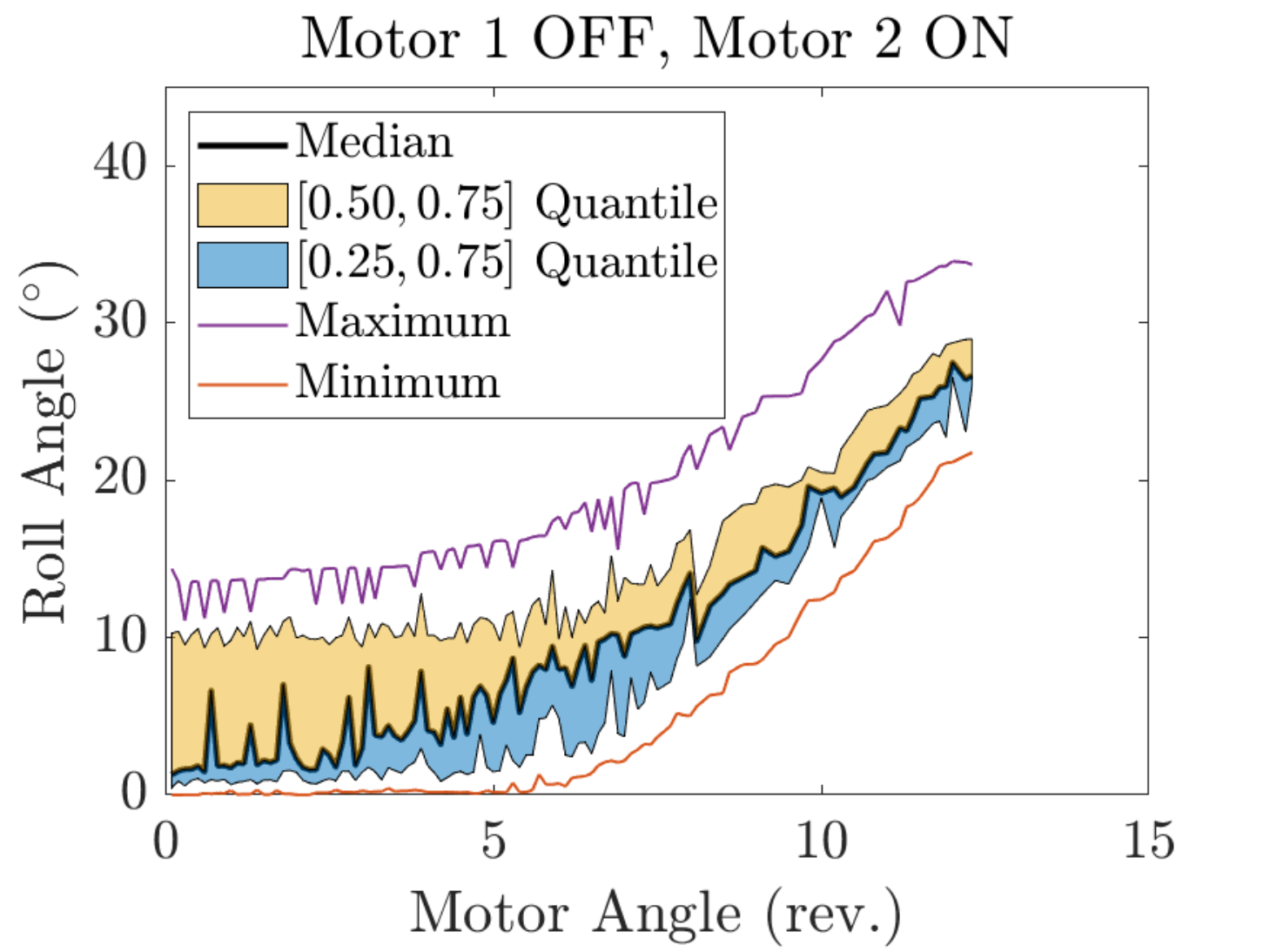}}
    \hfill
    \caption{\textcolor{red}{Results for the experimental characterization of the thumb. The motor angle input sequence to induce (a) the thumb bending and (b) the thumb roll in the ``vertically up" position. The Euler angles for (c) thumb bending and (d) thumb roll. Statistical analysis for (e) thumb bending and (f) thumb roll.}}
    \label{fig:thumb}
\end{figure*}

Each finger was experimentally characterized in four different orientations: 
\begin{enumerate}
    \item ``Vertically Up'': The palm is parallel to the gravitational acceleration direction and the fingers are above the palm.
    \item ``Vertically Down'': The palm is parallel to the gravitational acceleration direction and the fingers are below the palm.
    \item ``Horizontally Up'': The palm is perpendicular to the gravitational acceleration direction, and the palm is pointed upwards.
    \item ``Horizontally Down'': The palm is perpendicular to the gravitational acceleration direction, and the palm is pointed downwards.
\end{enumerate}
In total, sixteen experiments were conducted to characterize the fingertip positions of the four fingers. For all experiments with the four fingers, the fingertip angle was computed relative to its initial orientation. This meant the initial fingertip angle in all experiments was 0$\degree$. A typical experimental result is shown in Fig. \ref{fig:characterization}(b). As indicated in the aforementioned figure, the fingertip angle during the first half cycle followed a different trajectory than during the following cycles. This behavior is known as the ``lonely stroke'' and has commonly been observed in soft robots and artificial muscle actuators \cite{Konda2022}. A study on modeling the lonely stroke was conducted in \cite{Konda2022}. As shown in Fig. \ref{fig:characterization}(b), the correlation between motor angle and fingertip angle was also hysteretic. 

As it can be seen in Fig. \ref{fig:characterization}(c), the motor angle--fingertip angle correlation for different fingers showed an acceptable level of consistency. However, due to the compliance and softness of the gripper, inconsistencies in the motor angle--fingertip angle correlation may have arisen when the hand was tested in different orientations. Therefore, a statistical analysis was performed on the experimental data. The motor angle measurements were discretized using a resolution of 0.1 rotations. The encoder recorded measurements with a resolution of $2.78 \times 10^{-3}$ rotations during experiments and the sampling time was approximately 10\,ms. This meant that during a single experiment, many motor angle measurements could be grouped into a single ``bin'' that was 0.1-rotations wide. After the motor angle measurements were grouped into their respective bins, the fingertip angle measurements were grouped into corresponding bins. Then, the summary statistics of the fingertip angle measurements were computed at each discrete motor angle. At each discrete index, the distances between the quantiles and median could be computed for each finger. These values were computed using the following equation:
\begin{equation}
    \delta_{q} = \mathrm{mean}(|\alpha_{q} - \mathrm{median}(\alpha)|),
\end{equation}

\noindent where $\delta_{q}$ quantifies the average distance between the median fingertip angle and the $q$-quantile, and $q \in [0,1]$. The fingertip angle is expressed as $\alpha$. $\alpha_{q}$ is a vector containing $n$ elements that indicate the $q$-quantile of the fingertip measurements at each discrete index. For each finger, $\delta_0$, $\delta_{0.25}$, $\delta_{0.75}$, and $\delta_{1}$ were computed.

The summaries of the statistical analysis for each finger are shown in Fig. \ref{fig:characterization}(d)--(h). The data are presented as moving box-and-whisker plots. At each discrete motor angle, the median, 0.25-quantile, 0.5-quantile, minimum (0-quantile), and maximum (1-quantile) value is shown. The summary plot for all four fingers is provided in Fig. \ref{fig:characterization}(h). As expected, there is a much greater spread in the distribution of fingertip angles at each discrete motor angle. The 0.05-quantile and 0.95-quantile are also provided in Fig. \ref{fig:characterization}(h). The shaded regions altogether cover 90\% of the experimentally-obtained data distribution. Fig. \ref{fig:characterization}(i) shows the values of $\delta$ for all of the four fingers. \textcolor{red}{The measurement error from the IMU and the encoder may have affected the computed variances in actuation. Based on previous studies, the accuracy of a low-cost incremental encoder is often evaluated based on the resolution \cite{encoder1, encoder2}. Estimating the accuracy in terms of measurement error was beyond the scope of this study. The resolution of the encoder used in our study was $1^\circ$. The measurement error uncertainty of the IMU was $\pm2.5^\circ$.}

The data in Fig. \ref{fig:characterization}(a)--(i) was obtained using the same monotonic motor angle input sequence. This systematic procedure enabled fair comparisons between different fingers' behaviors. However, it was necessary to also characterize the gripper when the motor input sequence was not monotonic. Therefore, four cycles of randomly-generated motor reference signals were applied to the gripper's index finger. The motor angle and fingertip angles are plotted versus time in Fig. \ref{fig:characterization}(j) and \ref{fig:characterization}(k), respectively. For Fig. \ref{fig:characterization}(j) and (k), the top plot shows the entire sequence, whereas the bottom plot shows the sequence for one cycle. Fig. \ref{fig:characterization}(l) shows the correlation between the fingertip angle and motor angle. Despite minor differences, the correlation under random motor angle setpoints resembled that under monotonically increasing/decreasing setpoints. These results suggested that the correlation between motor angle and fingertip angle was mostly independent of the particular input sequence.

\textcolor{red}{The purpose of the random characterization was to qualitatively compare the finger performance against systematic characterizations. Conducting a quantitative comparison between the monotonic and random characterization results was beyond the scope of this study.}  
\subsubsection{Thumb}

\textcolor{red}{Similar to the four fingers, the thumb motion was characterized in four different gripper orientations. Fig. \ref{fig:thumb}(a) shows the motor input sequence to characterize the bending of the thumb, whereas Fig. \ref{fig:thumb}(b) shows the motor input sequence to characterize the roll of the thumb. ``Bending'' indicates that Motor \#1 is ON and Motor \#2 is OFF. ``Roll'' indicates that Motor \#1 is OFF and Motor \#2 is ON. The thumb's design meant that it did not rotate about only one axis. Therefore, it was important to study its changes in orientation about the $x$-, $y$- and $z$-axes, {whose locations relative to the thumb are shown in Fig. \ref{fig:imu_locations}}.}

\textcolor{red}{As an example, Fig. \ref{fig:thumb}(c) presents the change in orientation of the thumb when only Motor \#1 was ON when the gripper was in the ``Vertically Up" orientation. It can be seen from Fig. \ref{fig:thumb}(c) that when Motor \#1 is ON and Motor \#2 is OFF, while the orientation about all the axes changes, the orientation change about $x$-axis was most significant, with negligible creep. Similar results were obtained in the other gripper orientations. As an example, Fig. \ref{fig:thumb}(d) presents the orientation of the thumb when only Motor \#2 was ON when the gripper was in the ``Vertically Up" orientation. When Motor \#1 is OFF and Motor \#2 is ON, while the orientation about all the axes changes, the orientation change about $y$-axis was significant. However, while actuating the roll of the thumb, although the orientation change about $y$-axis is larger than others, the orientation changes about $x$ and $z$ axes were non-negligible (Fig. \ref{fig:thumb}(d)). Similar results were obtained in the other gripper orientations. This shows that the motion about the different axes ($x$-$y$-$z$) were coupled with each other while actuating the roll of the thumb. We believe that since the orientation change about the $x$-axis can be individually controlled through the bending of the thumb, the orientation change about $x$-axis can be controlled independently within a range of angles. In contrast, due to the design of the thumb and the placement of the actuators, the orientation changes about $y$ and $z$ axes cannot be controlled independently and are therefore coupled. It is this coupling that enables the thumb to achieve the desired degree of dexterity.}

\textcolor{red}{Fig. \ref{fig:thumb}(e) and Fig. \ref{fig:thumb}(f) show the statistical analyses for the thumb bending and thumb roll, respectively. These values were determined by computing the absolute changes in orientation by utilizing the measured Euler angles for each case. Fig. \ref{fig:thumb}(e)--(f) shows that while variations in the gripper orientations did not affect the bending actuation, they affected the actuation significantly. This can be inferred by the high variance in the roll actuation profiles. This aspect will be further examined in future work and the design of the thumb will be modified to minimize the effect of gravity on the roll actuation.}

\begin{figure*}
    \centering
    \subfloat[]{
    \includegraphics[height=4.5cm]{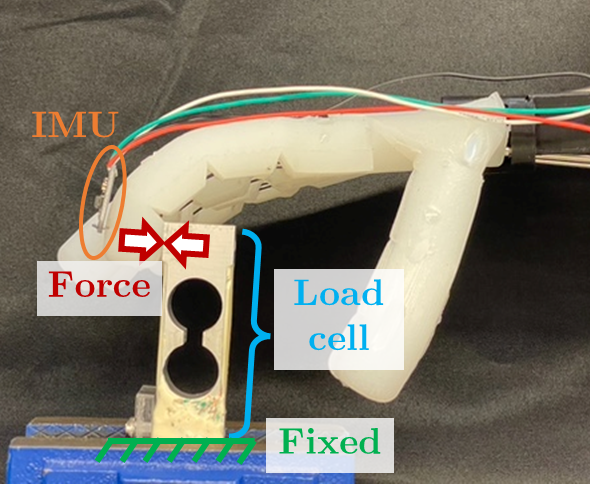}}
    \hfill
    \subfloat[]{
    \includegraphics[height=4.5cm]{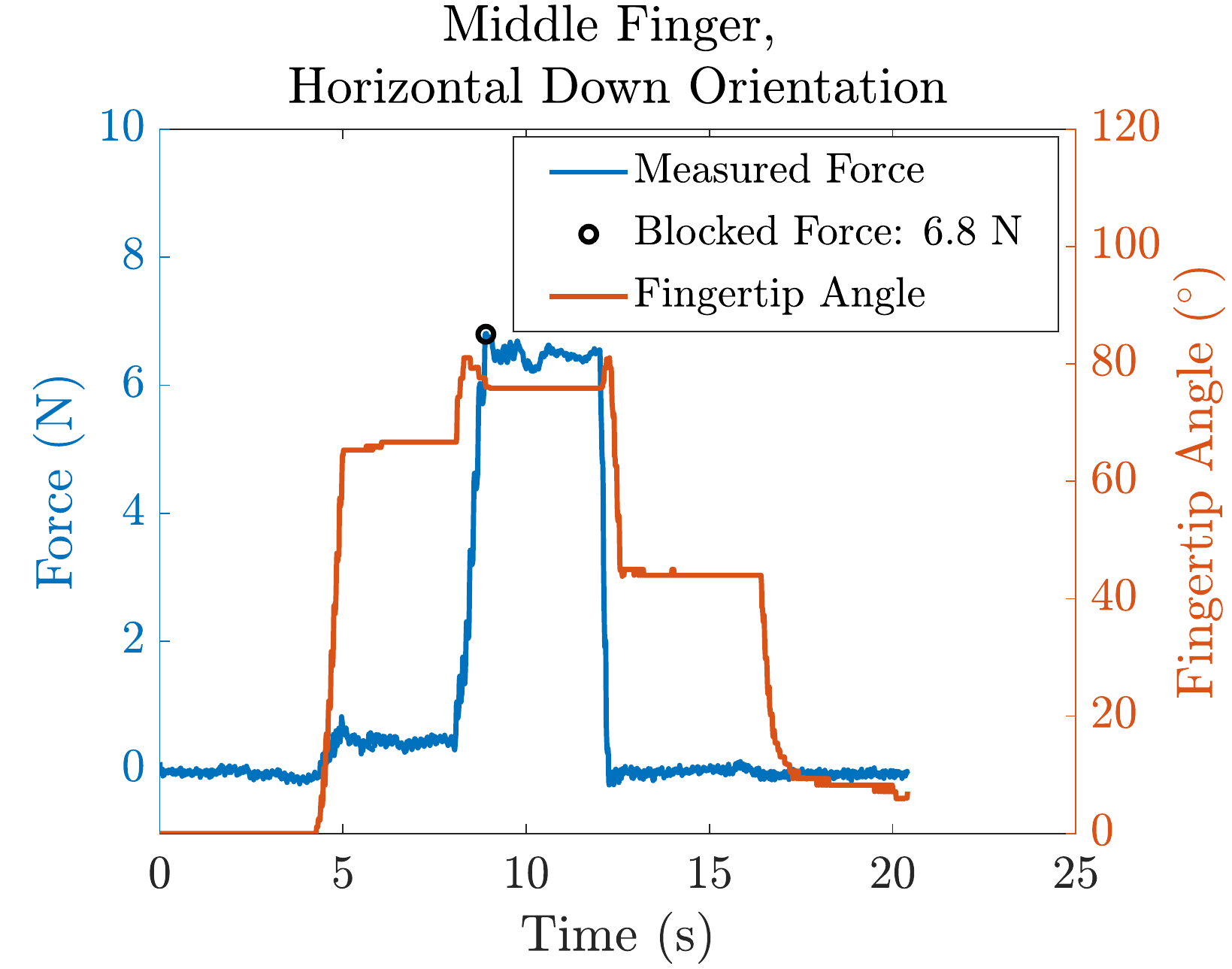}}
    \hfill
    \subfloat[]{
    \includegraphics[height=4.5cm]{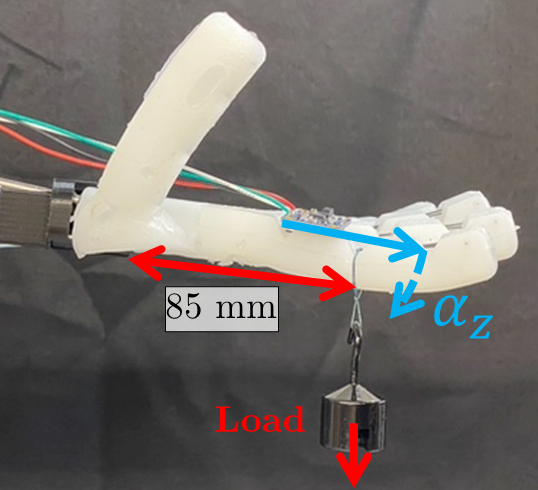}}
    \hfill
    \subfloat[]{
    \includegraphics[width=0.3\linewidth]{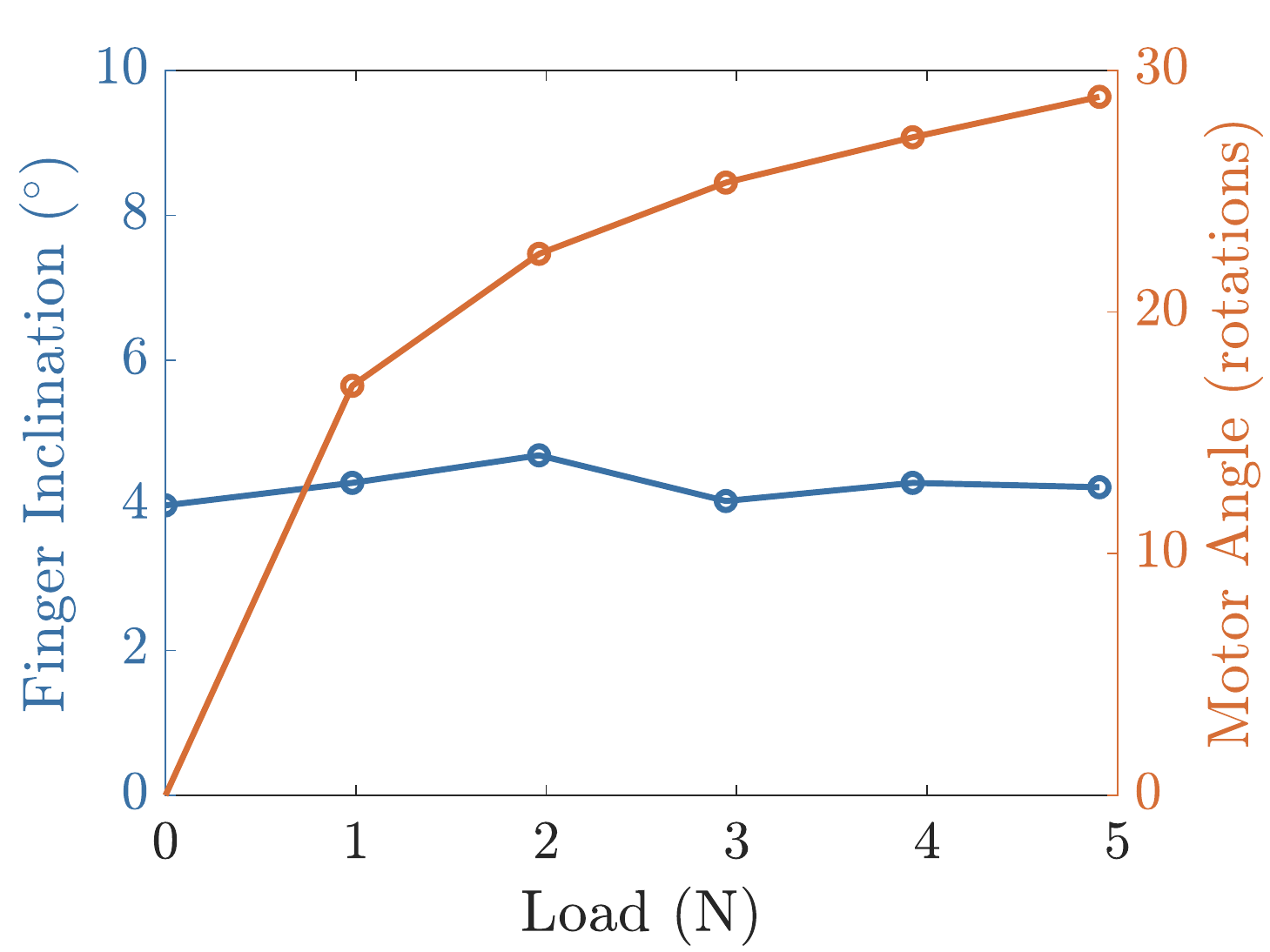}}
    \hfill
    \subfloat[]{
    \includegraphics[width=0.3\linewidth]{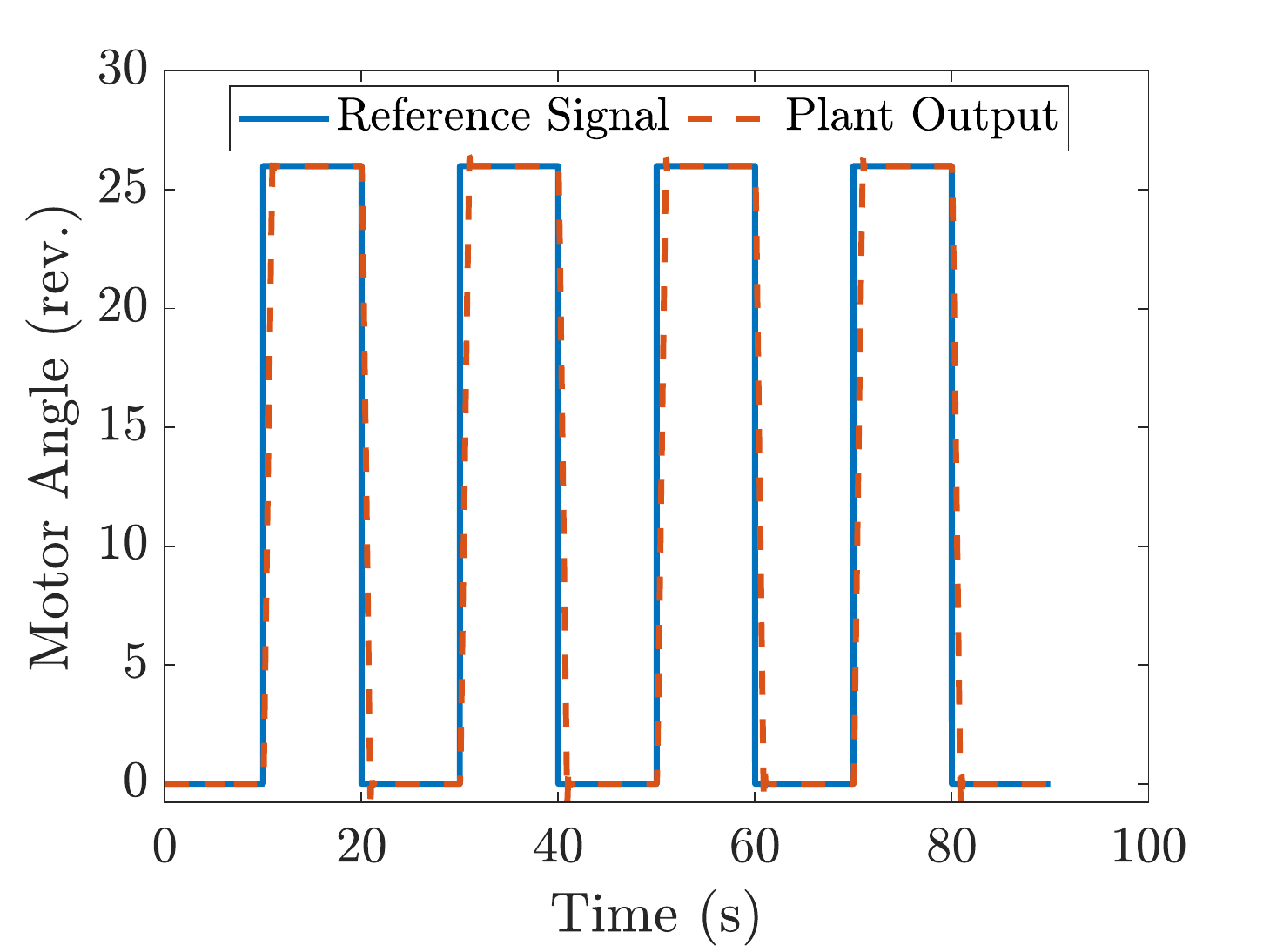}}
    \hfill
    \subfloat[]{
    \includegraphics[width=0.3\linewidth]{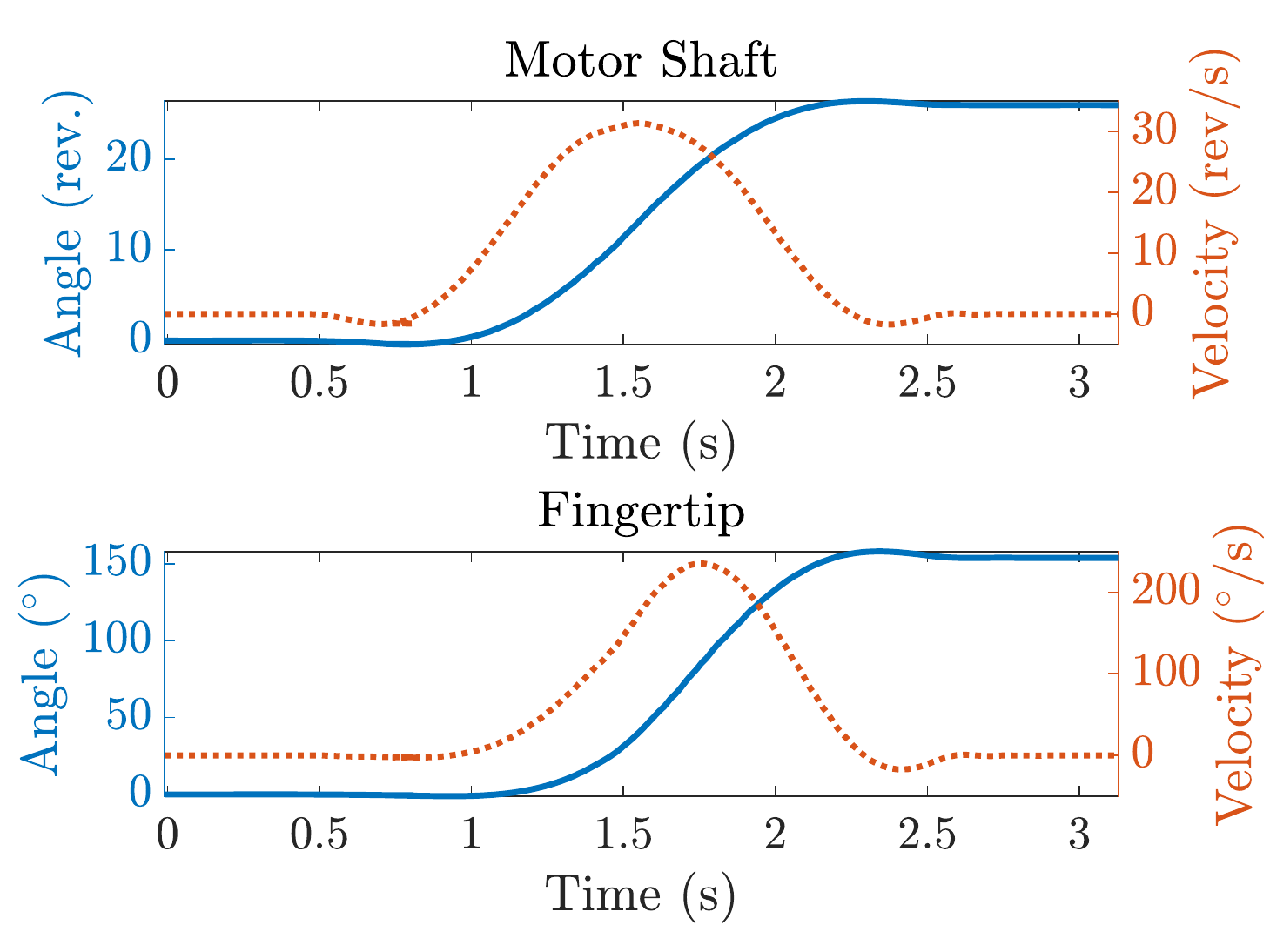}}
    \hfill
    \subfloat[]{
    \includegraphics[width=0.3\linewidth]{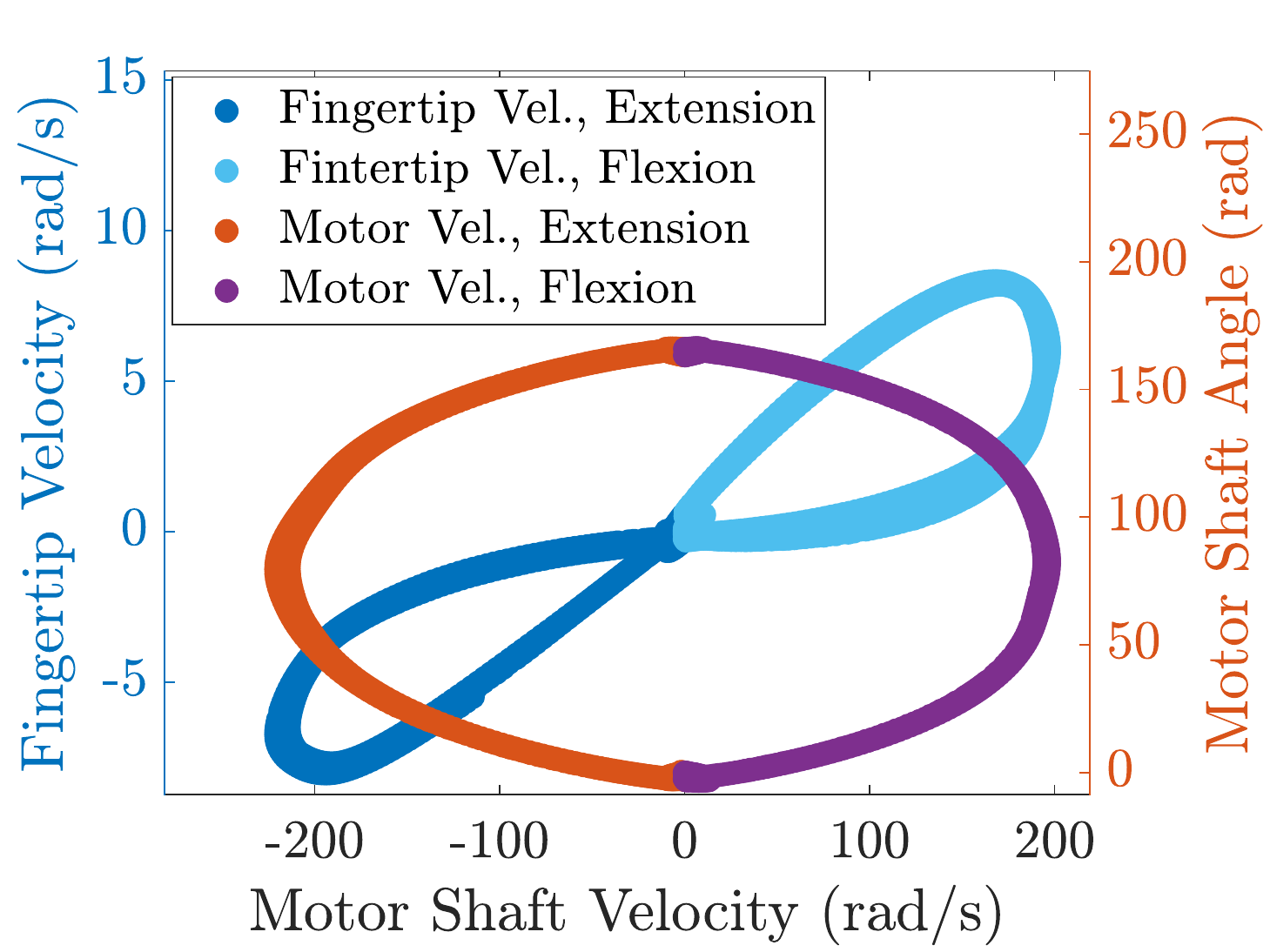}}
    \hfill
    \subfloat[]{
    \includegraphics[width=0.3\linewidth]{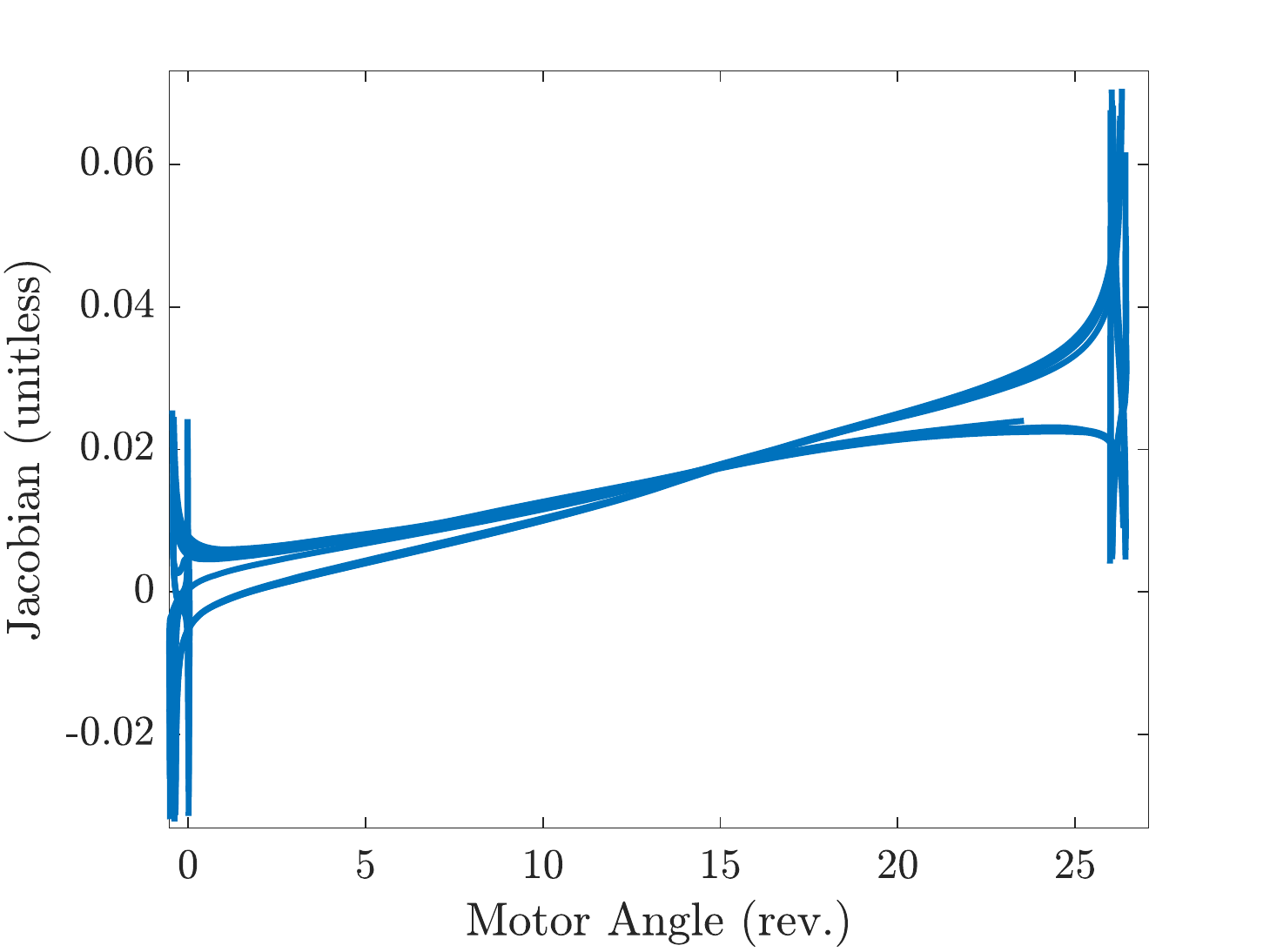}}
    \hfill
    \subfloat[]{
    \includegraphics[width=0.3\linewidth]{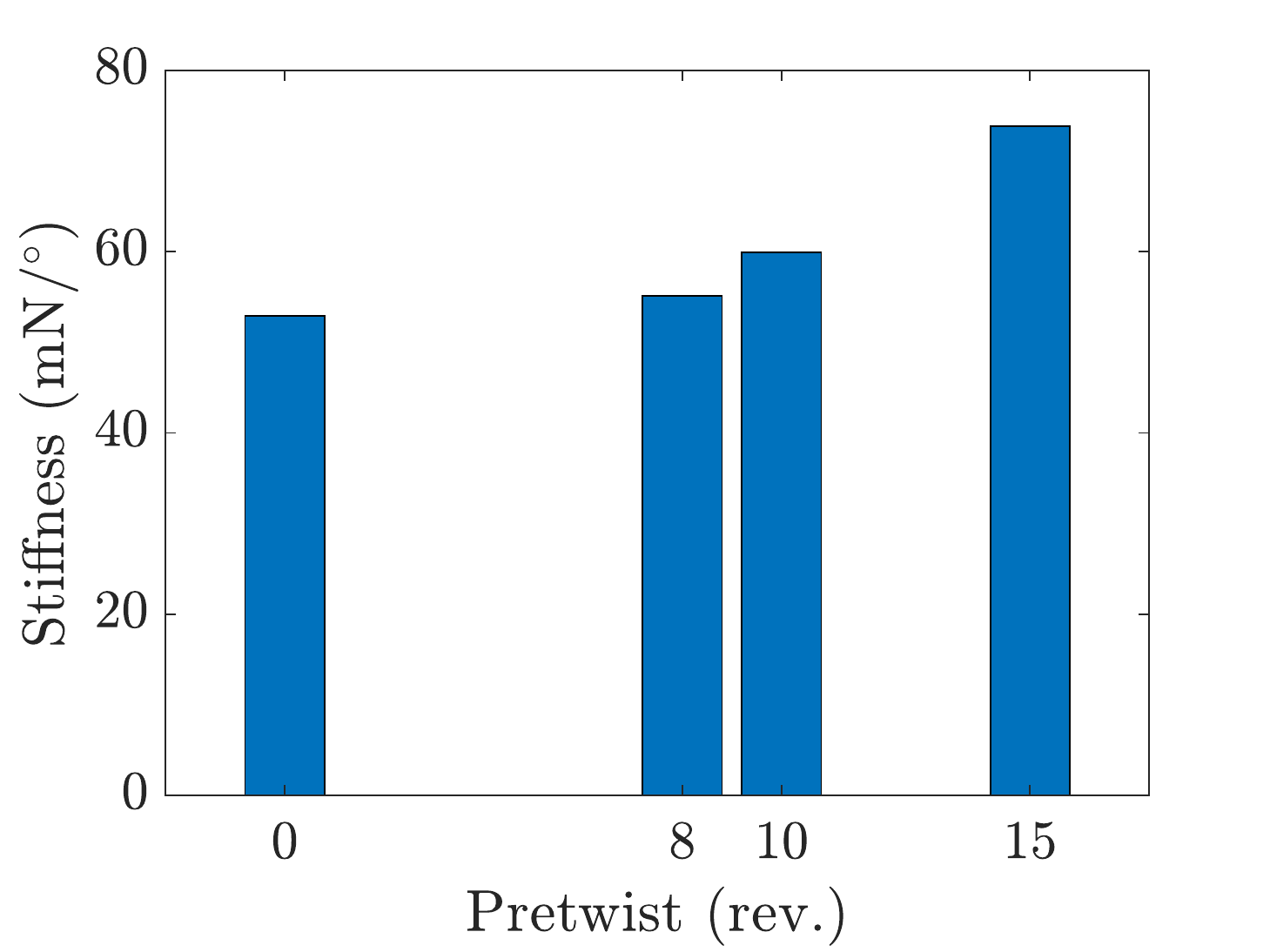}}
    \hfill
     \subfloat[]{
    \includegraphics[width=0.3\linewidth]{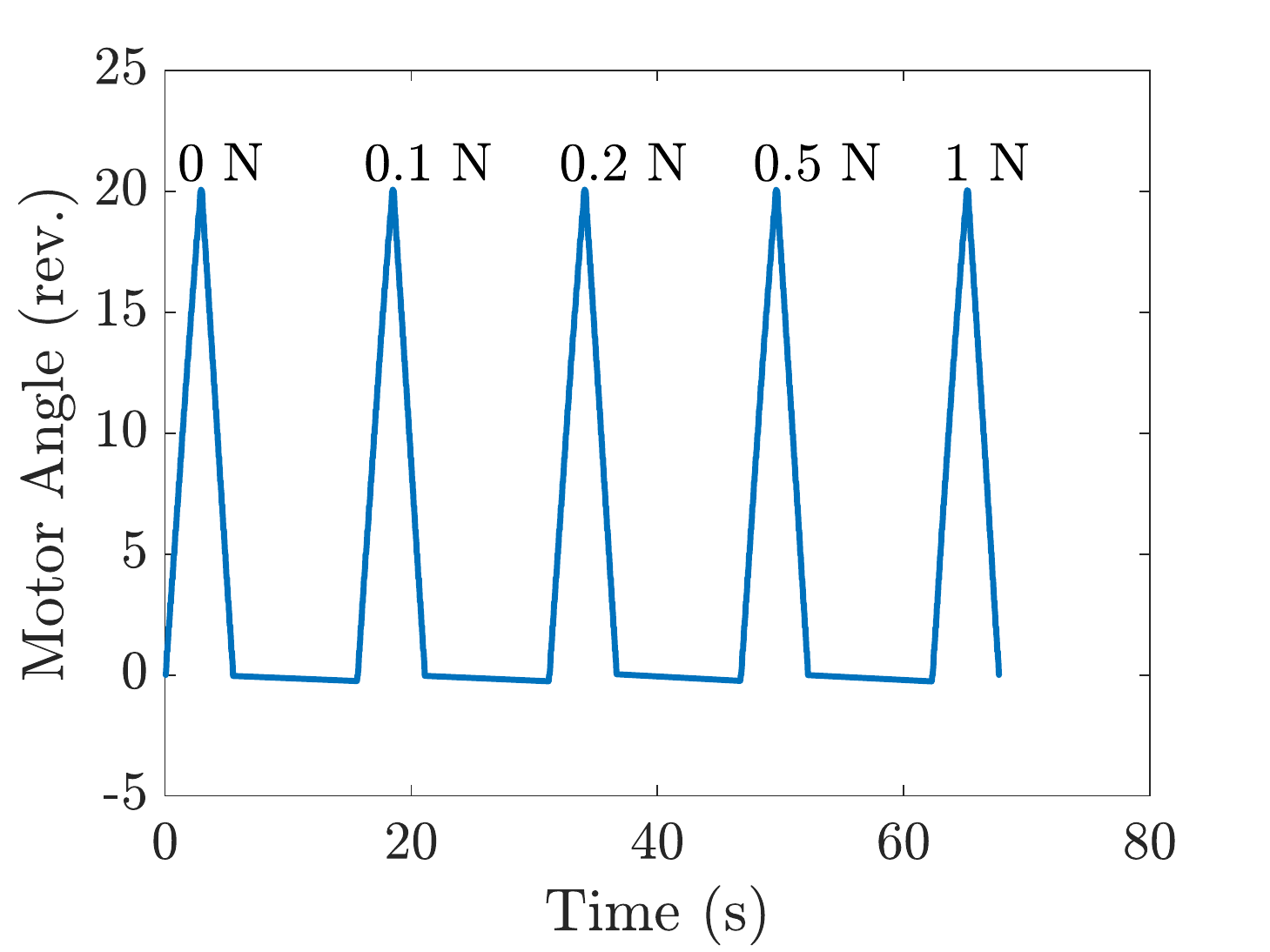}}
    \hfill
    \subfloat[]{
    \includegraphics[width=0.3\linewidth]{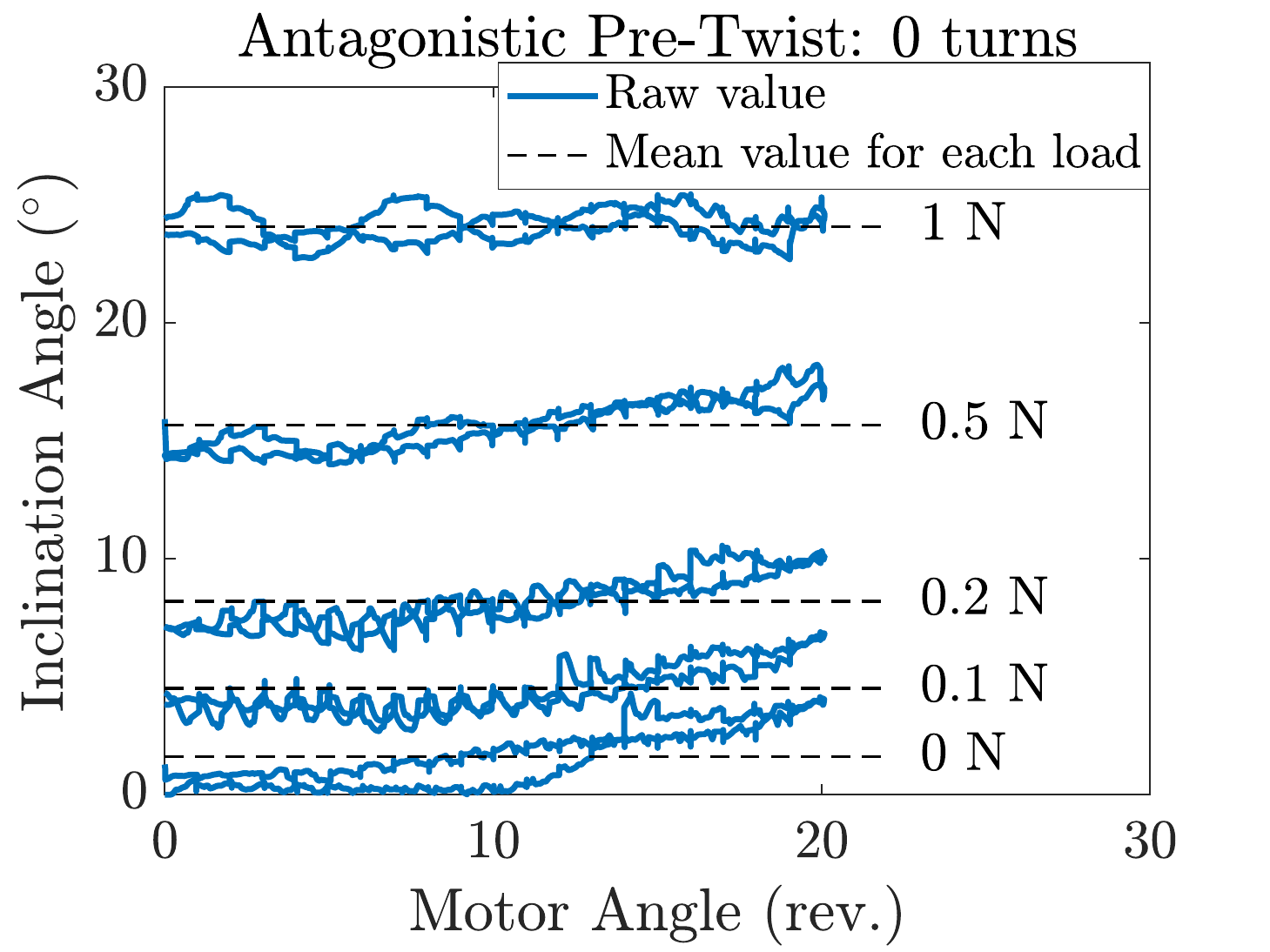}}
    \hfill
    \subfloat[]{
    \includegraphics[width=0.3\linewidth]{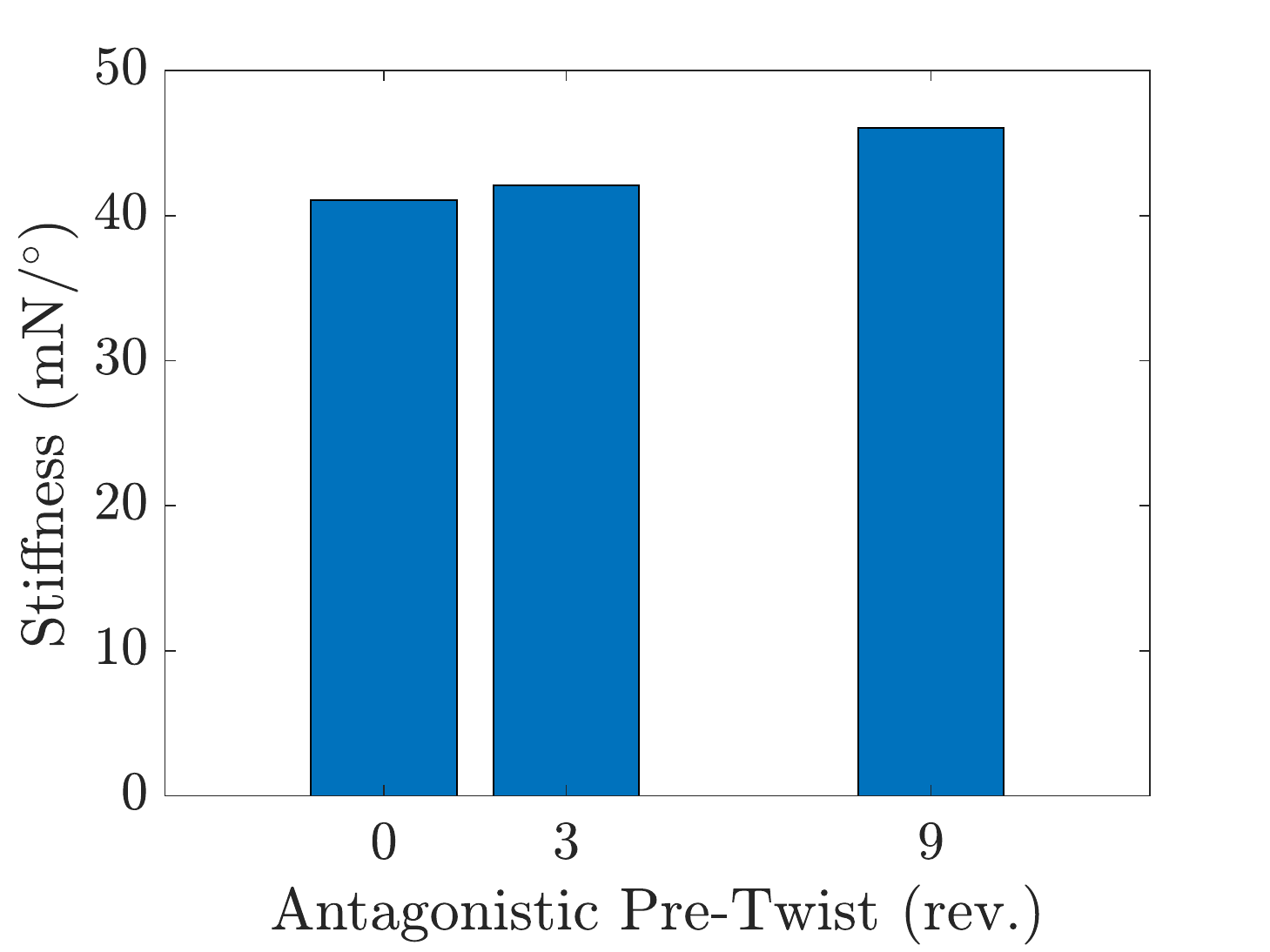}}
    \hfill
    \caption{\textcolor{red}{Experimental characterization of the index finger's velocity and force. (a) The experimental setup used during the first set of force characterization experiments. (b) A plot of the maximum fingertip blocked force. (c) The experimental setup used during the second set of force characterization experiments and tunable stiffness experiments, with the location of the inertial measurement unit (IMU) and hanging load shown.} (d) The increasing motor angles required to maintain a constant fingertip angle under increasing hanging load. (e) The input motor angle input sequence used to characterize the finger's velocity. (f, top) The motor shaft's angular position and velocity versus time. (f, bottom) The index fingertip's angular position and velocity versus time. (g) The fingertip velocity and motor shaft angle versus the motor shaft velocity. (h) The correlation between the motor angle and the Jacobian, which is the fingertip's angular velocity normalized to the motor shaft's velocity. (i) The variation of the finger stiffness with different amount of pre-twist in the primary TSA. \textcolor{red}{Results to characterize the lateral bending stiffness of the fingers. (j) The motor input sequence versus time. (k) The motor angle versus inclination angle when the fingers had zero antagonistic pre-twist. (l) The variation in lateral bending stiffness due to the antagonistic pre-twist.}}
    \label{fig:velocity_force}
\end{figure*}
\subsection{Force Output}
Two experiments were conducted to evaluate the force outputs of the gripper. Since all the fingers shared the same design, the force output of only one finger was characterized.
\subsubsection{Blocked Force Output}
The blocked force output of the index fingertip was experimentally obtained using the setup in Fig. \ref{fig:velocity_force}(a). A load cell (LSP-2, Transducer Techniques) was firstly fixed to a machine vise and the gripper was placed in the vertical down position. Then, the motor was first rotated to 13 rotations, and then to 26 rotations. The motor stalled just before reaching {its} setpoint of 26 rotations. At {its} stall, the finger exerted its maximum blocked force on the load cell. The results are shown in Fig. \ref{fig:velocity_force}(b). As shown in Fig. \ref{fig:velocity_force}(b), the fingertip exerted a maximum blocked force of 6.8~N.
\subsubsection{Constant Deflection and Varying Motor Angle}
For the second experiment, the motor angle of the primary TSA was adjusted to keep $\alpha_z$, the $z$-axis Euler angle, approximately constant under varying loads. Analogous to the human hand anatomy, the IMU was mounted on the index finger's proximal phalange. The experimental setup used for this experiment is presented in Fig. \ref{fig:velocity_force}(c). The following procedure was used for this experiment:
\begin{enumerate}
    \item Let $\alpha_{z0}$ denote the initial angle. Record $\alpha_{z0}$ when the motor angle $\theta = 0$ and the hanging mass $m = 0$. In this study, $\alpha_{z0} = 4.00 \degree$.
    \item Apply increasing hanging masses $m$ to the finger. In this study, $m = \{ 0, 100, 200, 300, 400, 500\}$\,g.
    \item At each mass, adjust $\theta$ such that $\alpha_z = \alpha_{z0} \pm \epsilon$. In this study, push buttons were used to manually twist the motors. The tolerance $\epsilon = 1\degree$.
\end{enumerate}

The purpose of the above experiment was to show that, despite the gripper's softness, its finger could maintain constant bending angles under increasing loads. Experimental results are provided in Fig. \ref{fig:velocity_force}(d). The index finger supported a hanging mass of 500\,g while maintaining its initial deflection angle (within the tolerance of $\epsilon$). This maximum mass was mostly limited by the stall torque of the motors: stronger motors would have allowed the gripper to support a greater amount of mass. However, they could also reduce the speed or increase the mass of the gripper.
\subsection{Velocity}
To study the peak velocity of the gripper's fingertip, the input sequence in Fig. \ref{fig:velocity_force}(e) was used. The velocity was obtained by numerically differentiating the position measurements. A Savitsky-Golay filter was applied to both the fingertip angle measurements and motor angle measurements to eliminate noise. The filter was applied during data post-processing but not during the online functioning of the gripper. Fig. \ref{fig:velocity_force}(f) shows the angular velocities of the motor shaft and fingertip during one cycle of ascending motor rotations. In Fig. \ref{fig:velocity_force}(f), the peak angular velocities were 31.3712~rev/s and 235.2922~$\degree$/s for the motor angle and fingertip angle, respectively. Fig. \ref{fig:velocity_force}(g) shows the correlation between the motor shaft velocity and the fingertip velocity on the left $y$-axis. The two variables were mostly positively correlated, but there was a large amount of hysteresis. There was also evident ``lonely stroke'' in the right quadrant of Fig. \ref{fig:velocity_force}(g). The correlation between the motor shaft velocity and motor shaft angle is shown on the right $y$-axis. This correlation was a product of (1) the proportional control scheme and (2) the voltage--motor angle transfer function. The Jacobian --- the ratio of the fingertip angular velocity to the motor shaft angular velocity --- is provided in Fig. \ref{fig:velocity_force}(h). The Jacobian is defined as:
\begin{equation}
    \mathcal{J} = \frac{d\alpha/dt}{d\theta/dt},
\end{equation}
\noindent where $\mathcal{J}$ is the Jacobian, $d\alpha/dt$ is the angular velocity of the fingertip, and $d\theta / dt$ is the angular velocity of the motor shaft. The Jacobian (also known as the reduction ratio) typically increased as the motor angle increased. It is well documented that the Jacobian of the TSA (its ratio of linear contraction velocity to motor shaft velocity) also increases as the motor shaft rotations increases \cite{Model_Tmech14}. At the extreme ends of the $x$-axis, $\mathcal{J}$ sharply increased in magnitude. {This was likely because of the jerk caused due to the sudden acceleration when the motor changed direction.}
\subsection{Tendon-Based Stiffening}
In the experiment conducted to evaluate the adjustable stiffness property of the gripper, both the primary TSA and the antagonistic TSA were actuated. The setup used for the second set of force characterization experiments (Fig. \ref{fig:velocity_force}(c)), was used for this experiment as well. In this experiment, the following procedure was used:
\begin{enumerate}
    \item With both TSAs fully untwisted, record the $z$-axis Euler angle, $\alpha_{z}$, from the IMU. Let $\alpha_{z0}$ denote the initial angle. The location of the IMU is shown in Fig. \ref{fig:velocity_force}(c, bottom).
    \item Twist the primary TSA by a given amount, $\theta_p$. In this study, $\theta_p = \{0, 8, 10, 15 \}$ rotations. This pre-twist will cause $\alpha_z > \alpha_{z0}$. 
    \item Twist the antagonistic TSA until $\alpha_z = \alpha_{z0}$. This was realized using closed-loop control of the antagonistic TSA, where the voltage input to the motor was proportional to the error in $\alpha_z$. Let $\theta_a$ denote the amount of twists from the antagonistic TSA that makes $\alpha_z = \alpha_{z0}$. Note that $\theta_p \neq \theta_a$ because the primary and antagonistic TSAs may have slightly different amounts of initial tension and string lengths.
    \item Systemically apply monotonically increasing and decreasing hanging masses to the finger. Hanging masses of $m = \{0, 60, 100, 120, 140, 160, 200 \}$\,g were applied in this study. The masses were hung from the string shown in Fig. \ref{fig:velocity_force}(c). Each mass was held constant for 30\,s. To distinguish between the ``lonely stroke'' and the subsequent cycles' behavior, two cycles of increasing/decreasing loading were applied.
    \item At each loading condition, record $\alpha_z$ and compute $\alpha_{z0}  - \alpha_z$.
    \item Compute a first-degree polynomial that correlates the load (in N) to $\alpha_{z0} - \alpha_z$:
    \begin{equation}
        mg = K_\alpha (\alpha_{z0 - \alpha_z}) + K_0,
    \end{equation}
    \noindent where $m$ is the mass of each hanging load, $g$ is the gravitational acceleration, and $K_{\alpha},\, K_0$ are coefficients to be computed. Because $\alpha_{z0}$ is defined as $\alpha_z$ when $m = 0$, $K_0 = 0$. $K_\alpha$ is then the stiffness of the finger, in terms of its angular deflection.
\end{enumerate}
\begin{figure*}
    \centering
    \subfloat[]{
    \includegraphics[height=4.55cm]{{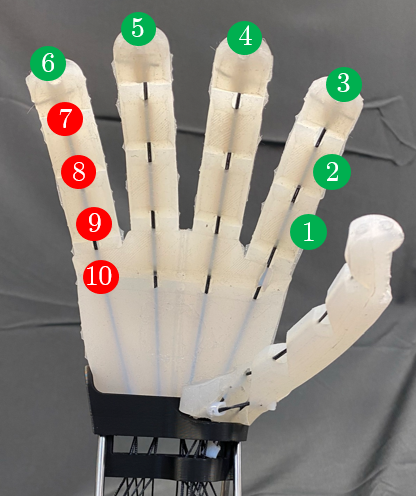}}}
    \hfill
    \subfloat[]{
    \includegraphics[height=4.55cm]{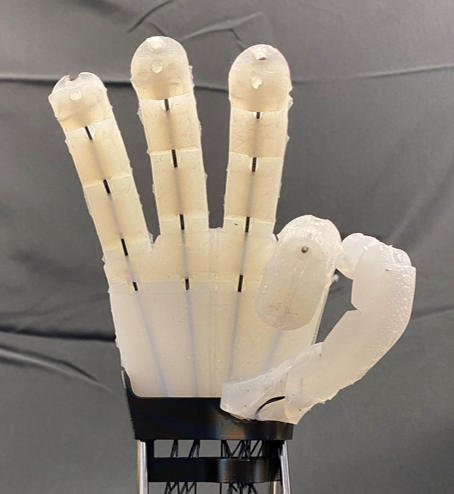}}
    \hfill
    \subfloat[]{
    \includegraphics[height=4.55cm]{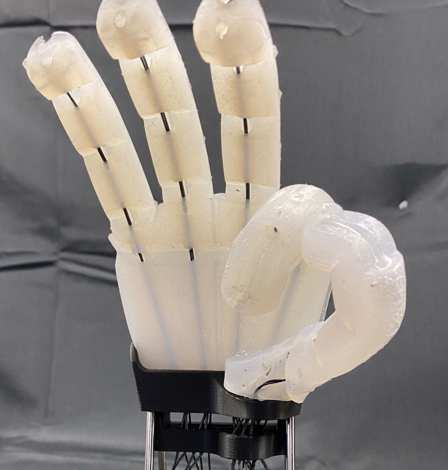}}
    \hfill
    \subfloat[]{
    \includegraphics[height=4.55cm]{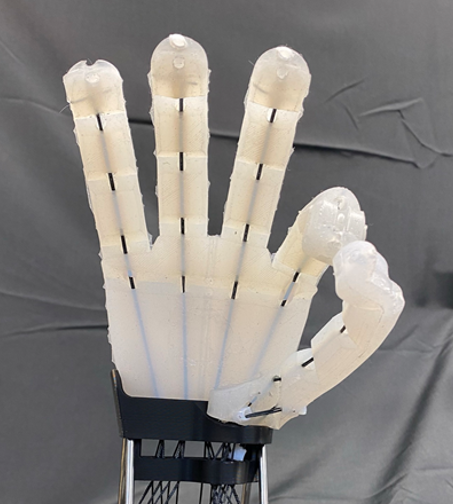}}
    \hfill
    \subfloat[]{
    \includegraphics[height=4.55cm]{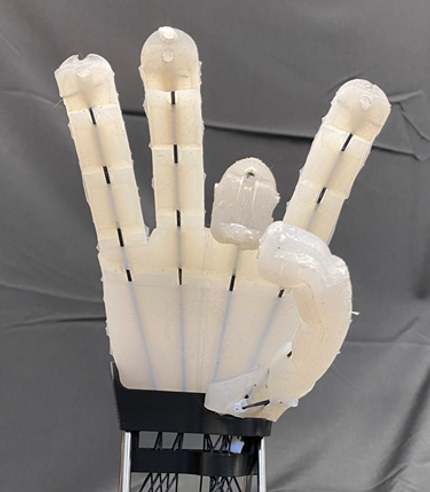}}
    \hfill
    \subfloat[]{
    \includegraphics[height=4.55cm]{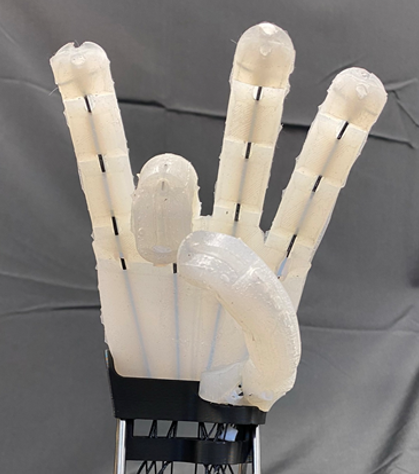}}
    \hfill
    \subfloat[]{
    \includegraphics[height=4.55cm]{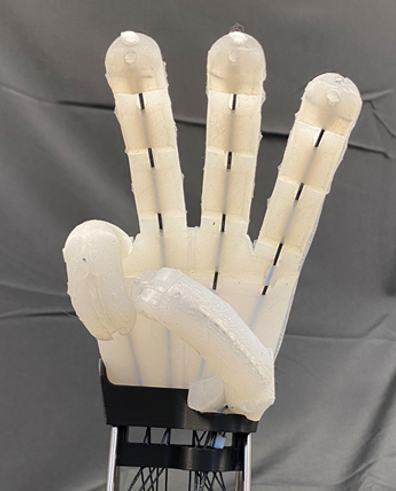}}
    \hfill
    \subfloat[]{
    \includegraphics[height=4.55cm]{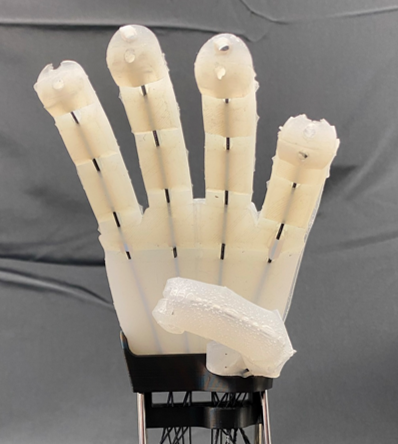}}
    \hfill
    \caption{(a) The locations on the gripper which the thumb has to touch as a part of the Kapandji Test. Attainable contacts are shown in green and failed locations are shown in red. (b) Hand achieving Kapandji Test contact \#1 at proximal phalanx, (c) contact \#2 at middle phalanx, (d) contact \#3 at tip of index finger, (e) contact \#4 at tip of middle finger, (f) contact \#5 at tip of ring finger, (g) contact \#6 at tip of little finger, (h) best effort to achieve contact \#10 at distal palmar crease. Lack of thumb dexterity at maximum roll prevented the achievement of contacts \#7-10.}
    \label{fig:KapandjiResults}
\end{figure*}
\noindent The above procedure was used to obtain the results presented in Fig. \ref{fig:velocity_force}(i). The stiffness increased as the pre-twist increased. However, there was an insignificant change in stiffness $K_\alpha$ between pre-twists of 0 and 8 rotations. This was likely due to two reasons. (1) At low motor angles, the TSA had low linear contraction per twist. (2) The strings may have been slightly loose initially, such that the first twists of the TSAs caused zero linear contraction of the strings. Note that the maximum amount of pre-twist was limited by the torque output of the motor. As the pre-twist increased, the primary motor needed to exert more torque to overcome the opposing force from (1) the silicone material and (2) the antagonistic motor. {These results are similar to those presented in \cite{Popov2014}, in which both empirical and simulated results showed a decrease in the effectiveness of an antagonistic TSA to increase stiffness at the extreme ends of a joint's actuation.}

\textcolor{red}{Although the fingers were designed to bend primarily in one direction, a defining characteristic of soft materials is that they don't rigidly constrain motion in other directions. Thus the lateral/off-axis bending stiffness was experimentally characterized. Two types of experiments were conducted to examine the lateral stiffness of the fingers. Firstly, the effect of bending actuation on lateral bending was investigated. For this purpose, the antagonistic TSA was maintained at 0 pre-twists. The finger was subjected to a load in the lateral direction and the corresponding TSA was actuated to induce bending of the finger. Fig. \ref{fig:velocity_force}(j) shows the motor angle input sequence versus time. Five different hanging loads were applied, with one unique load per cycle. Loads of $\{0, 0.1, 0.2, 0.5, 1.0 \}$\,N were applied. Fig. \ref{fig:velocity_force}(k) shows the inclination versus motor angle. For each load, the average inclination is computed. The results show that the fingers were, due to their soft structures, susceptible to significant deflection as a result of the lateral loading. In these tests, the bending actuation mildly affected the lateral bending of the fingers, with the fingers deflecting slightly more as they were contracted further.}

\textcolor{red}{Secondly, the effect of stiffness tuning on the lateral stiffness of the finger was examined. The three different amounts of pre-twist in the antagonistic TSA were studied: 0, 3, and 9 twists. At each pre-twists, loads of $\{0, 0.1, 0.2, 0.5, 1.0 \}$\,N were applied in a monotonically increasing manner. Fig. \ref{fig:velocity_force}(l) shows the antagonistic pre-twist versus the stiffness. These two variables are positively correlated: adding antagonistic pre-twists increased the lateral stiffness of the fingers. While the two variables were indeed correlated, the amount of stiffness tuning induced was less compared to the stiffness tuning induced in the bending direction.}
\begin{figure*}

    \includegraphics[width=0.13\linewidth]{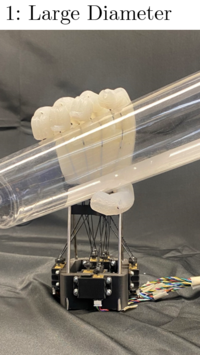}
    \hfill
    \includegraphics[width=0.13\linewidth]{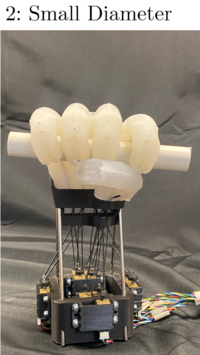}
    \hfill
    \includegraphics[width=0.13\linewidth]{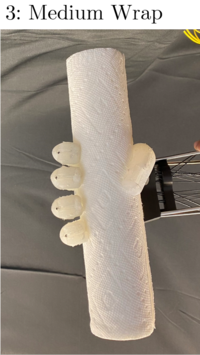}
    \hfill
    \includegraphics[width=0.13\linewidth]{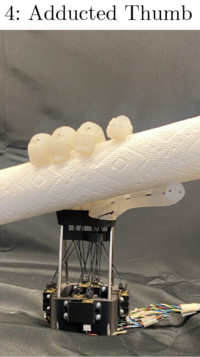}
    \hfill
    \includegraphics[width=0.13\linewidth]{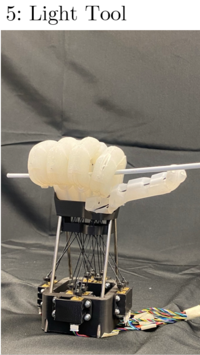}
    \hfill
    \includegraphics[width=0.13\linewidth]{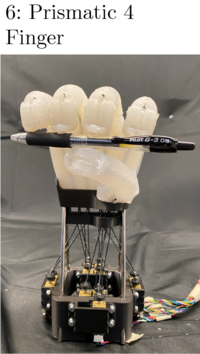}
    \hfill
    \includegraphics[width=0.13\linewidth]{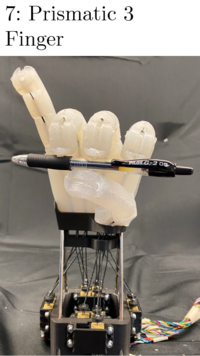}
    \hfill
    \includegraphics[width=0.13\linewidth]{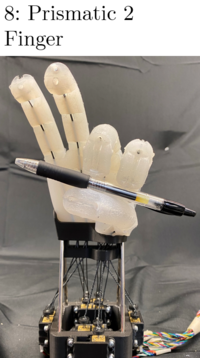}
    \hfill
    \includegraphics[width=0.13\linewidth]{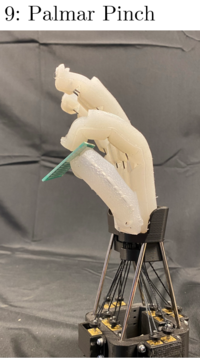}
    \hfill
    \includegraphics[width=0.13\linewidth]{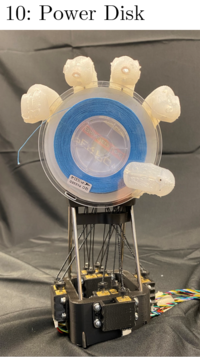}
    \hfill
    \includegraphics[width=0.13\linewidth]{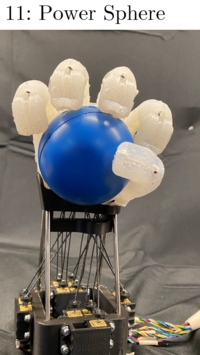}
    \hfill
    \includegraphics[width=0.13\linewidth]{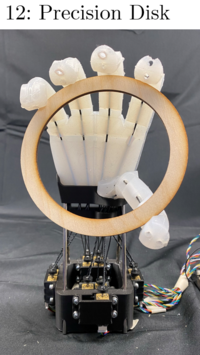}
    \hfill
    \includegraphics[width=0.13\linewidth]{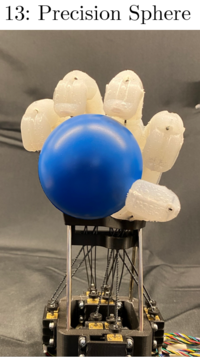}
    \hfill
    \includegraphics[width=0.13\linewidth]{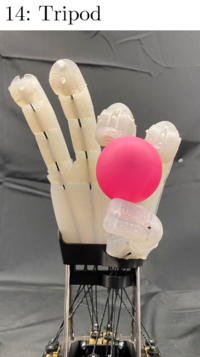}
    \hfill
    \includegraphics[width=0.13\linewidth]{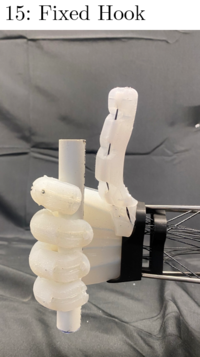}
    \hfill
    \includegraphics[width=0.13\linewidth]{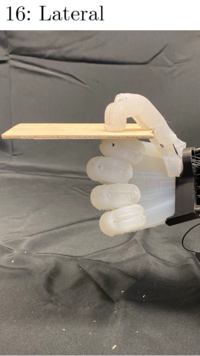}
    \hfill
    \includegraphics[width=0.13\linewidth]{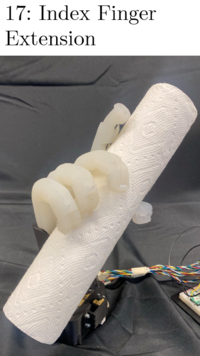}
    \hfill
    \includegraphics[width=0.13\linewidth]{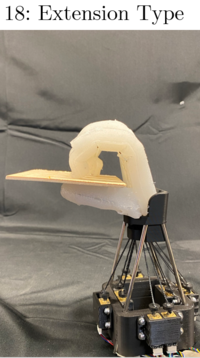}
    \hfill
    \includegraphics[width=0.13\linewidth]{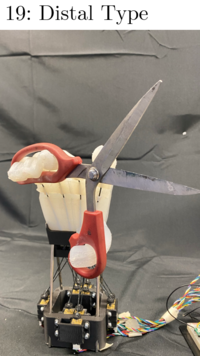}
    \hfill
    \includegraphics[width=0.13\linewidth]{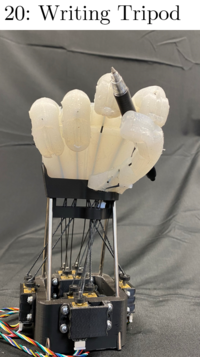}
    \hfill
    \includegraphics[width=0.13\linewidth]{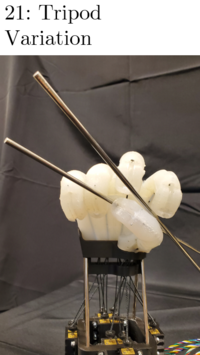}
    \hfill
    \includegraphics[width=0.13\linewidth]{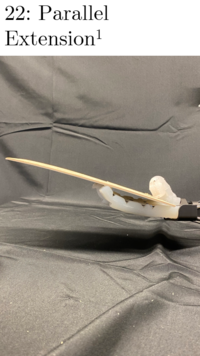}
    \hfill
    \includegraphics[width=0.13\linewidth]{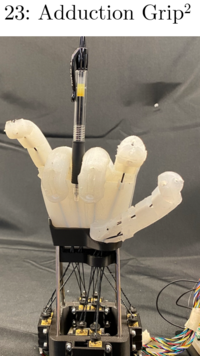}
    \hfill
    \includegraphics[width=0.13\linewidth]{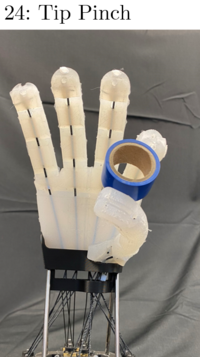}
    \hfill
    \includegraphics[width=0.13\linewidth]{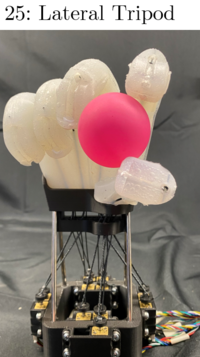}
    \hfill
    \includegraphics[width=0.13\linewidth]{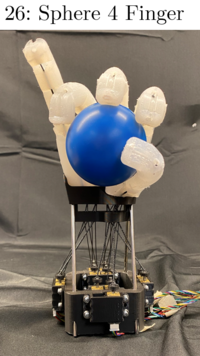}
    \hfill
    \includegraphics[width=0.13\linewidth]{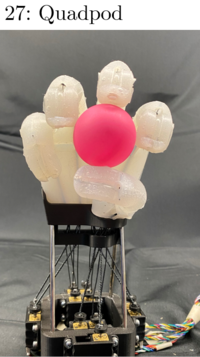}
    \hfill
    \includegraphics[width=0.13\linewidth]{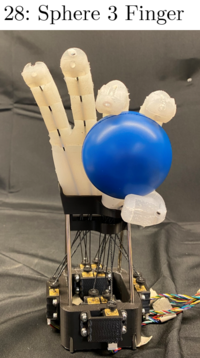}
    \hfill
    \includegraphics[width=0.13\linewidth]{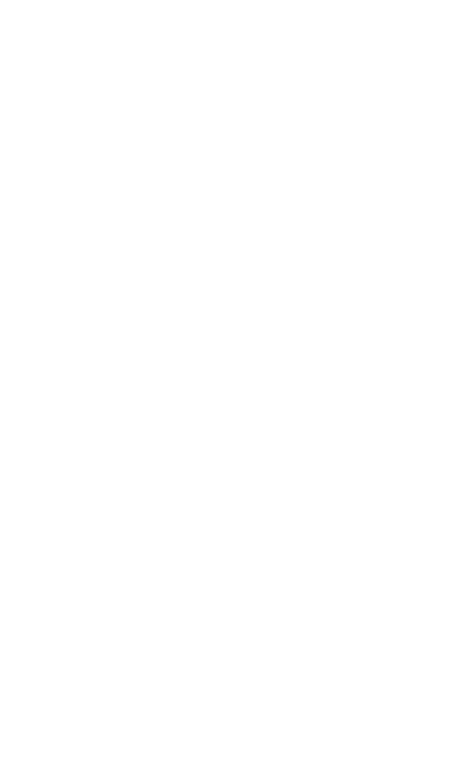}
    \hfill
    \includegraphics[width=0.13\linewidth]{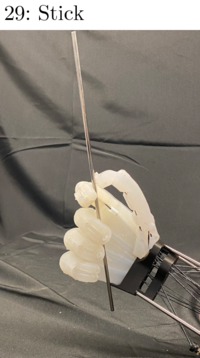}
    \hfill
    \includegraphics[width=0.13\linewidth]{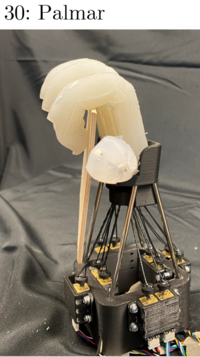}
    \hfill
    \includegraphics[width=0.13\linewidth]{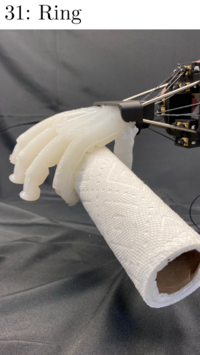}
    \hfill
    \includegraphics[width=0.13\linewidth]{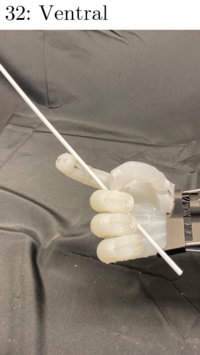}
    \hfill
    \includegraphics[width=0.13\linewidth]{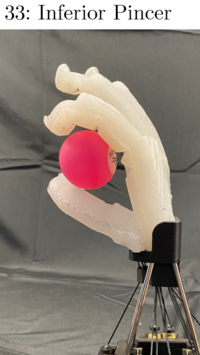}
    \hfill
    \includegraphics[width=0.13\linewidth]{images/GRASPTaxonomy/PLACEHOLDER.png}
    \hfill
    \caption{Achievable grasps from the Feix GRASP Taxonomy; grasps are numbered and labelled as described in \cite{Feix2016}. $^{1}$ Failed grasp. Thumb is not capable of fully realized parallel extension. $^{2}$ Failed grasp. Adduction grip is maintained passively with the actuators not actively applying the force.}
    \label{fig:GRASPTaxonomy}
\end{figure*}
\begin{figure}
    \centering
    \subfloat[]{
    \includegraphics[width=8.5cm]{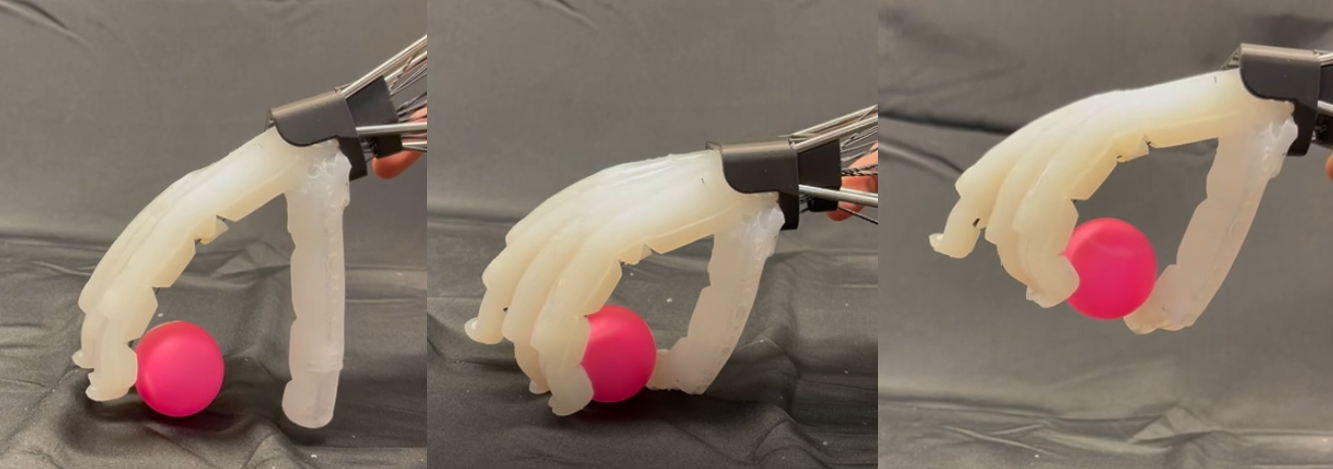}}
    \hfill
    \subfloat[]{
    \includegraphics[width=8.5cm]{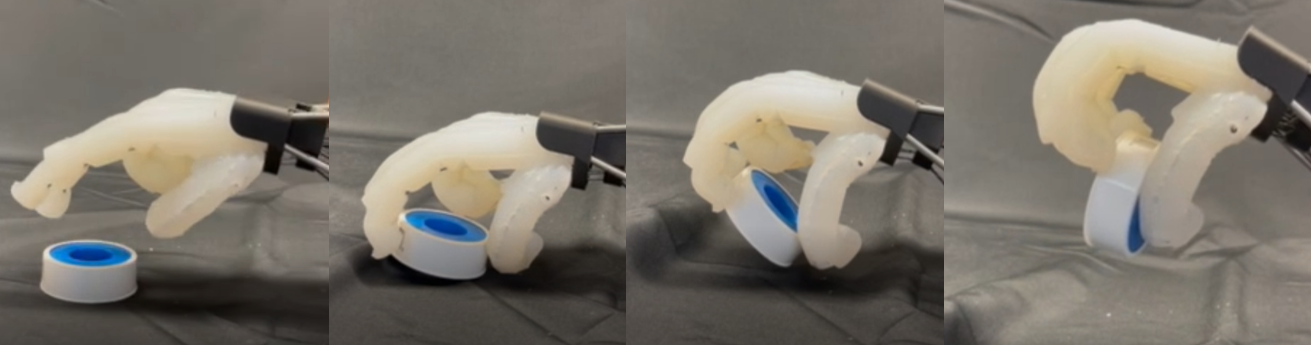}}
    \hfill
    \subfloat[]{
    \includegraphics[width=8.5cm]{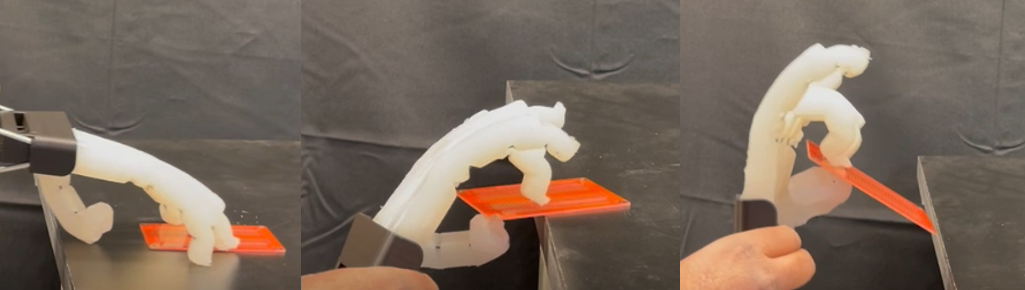}}
    \hfill
    \caption{Hand performing grasp strategies from \cite{Heinemann2015, RBO3}. Shown are (a) top grasp, (b) flip grasp, and (c) edge grasp.}
    \label{fig:GraspTypes}
\end{figure}
\section{Grasping Performance}
\subsection{Thumb Dexterity}
The dexterity of the thumb was quantified using the Kapandji test, which was originally developed to evaluate hand motor functions of patients after stroke or surgery \cite{KAPANDJI198667}. However, with increasing research activity in the field of dexterous manipulation, researchers have employed the Kapandji test to quantify thumb opposability \cite{Jianshu2018,Jianshu2019,Deimal2016}. As shown in Fig. \ref{fig:KapandjiResults}(a), in the Kapandji test, the tip of the thumb has to touch ten different locations on the hand which include the tips of the other fingers. A score is then assigned based on the number of locations the thumb can touch. A score of zero would indicate no thumb opposability, while a score of ten indicates fully anthropomorphic opposability. The soft thumb presented in this study achieved a score of 6. While the thumb was able to touch all the locations on the index finger and the tips of the other fingers, it failed to touch the remaining locations on the little finger and distal end of the palm, as shown in Fig. \ref{fig:KapandjiResults}(b)--(h). This is because the mobility of the thumb decreased at the extreme end of its roll actuation. This also meant that, although achieved, the contact between the thumb and the tip of the little finger was less substantial as compared to the contacts with the other fingers. Increasing the range of motion of the roll joint of the thumb could help improve the Kapandji score of the thumb. The results of the Kapandji test demonstrate the dexterity of the thumb design and its significance in the hand’s ability to achieve different grasps.

\subsection{Grasping Performance}
The performance of the gripper was analyzed by replicating a variety of anthropomorphic grasps according to the Feix GRASP taxonomy presented in \cite{Feix2016}. This taxonomy has been widely used to demonstrate the dexterity of grippers in previous studies \cite{RBO3, Deimal2016}. The results achieved by the proposed gripper are presented in Fig. \ref{fig:GRASPTaxonomy}. The proposed gripper was able to achieve 31 of the 33 grasps presented in \cite{Feix2016}. The gripper failed to achieve the parallel extension grasp (grasp \#22); the best effort is shown in Fig. \ref{fig:GRASPTaxonomy}. To achieve this grasp, the fingers and thumb needed to actuate from their bases to clamp an object while remaining extended. However, due to the pseudo joints in the proposed finger design, the bending was uniformly distributed across the length of the finger. This phenomenon made it nearly impossible for the gripper to achieve the proximal bending concentration required for the parallel grasp. This can be addressed by employing fingers that do not use pseudo joints or adding a second DOF in each finger to allow decoupled proximal bending. Furthermore, for the adduction grip (grasp \#23), the gripper was only able to hold on to the object through passive force and not using a controllable DOF. Enabling this actuation can be addressed in the next iteration of the gripper by adding a controllable DOF in the palm of the gripper.

\begin{figure}
    \centering
    \subfloat[]{
    \includegraphics[width=0.9 \linewidth]{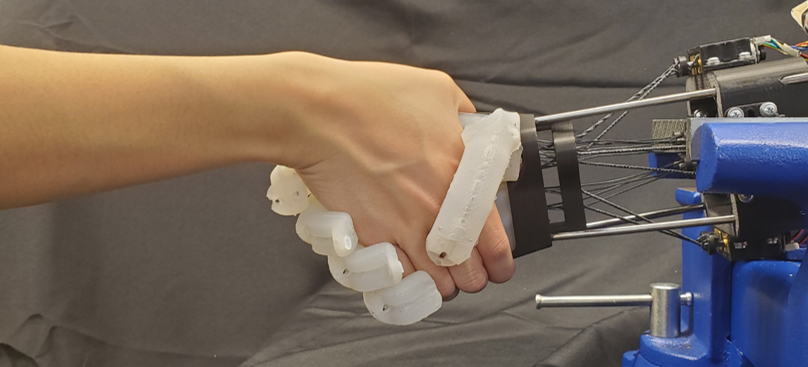}}
    \hfill
    \subfloat[]{
    \includegraphics[width=0.9 \linewidth]{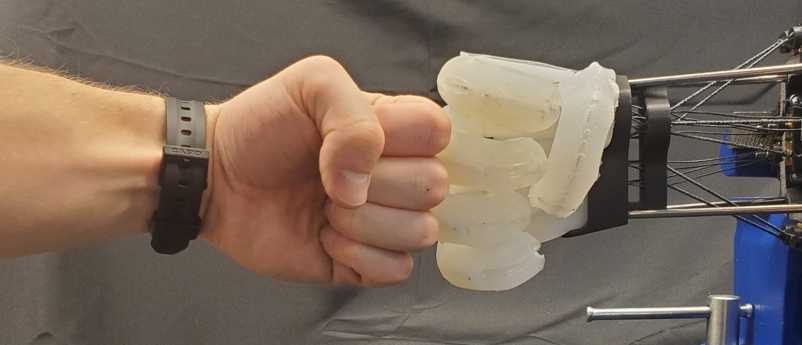}}
    \hfill
    \caption{Images showing safe human-robot interaction with the gripper through (a) a handshake and (b) a fist bump.}
    \label{fig:human-robot interaction}
\end{figure}

In addition, other commonly used grasp strategies by the human hand, such as the top grasp, flip grasp, and the edge grasp strategies \cite{Heinemann2015, RBO3} were also demonstrated. For this purpose, the gripper was controlled in an open-loop fashion to grasp a ping-pong ball, a roll of tape, and a circuit board off a table with a flat surface to demonstrate the top grasp, flip grasp, and edge grasp strategies, respectively. For the top grasp, the tips of the fingers touch the flat surface which then guides their actuation until a precision grasp is achieved. For the flip grasp, the fingers used the opposing force from the thumb to flip the object before achieving a precision grasp to firmly hold the object. During the edge grasp, which is mostly employed to grasp flat objects, the hand makes use of the flat surface of the table to slide the object toward the edge, until the bottom side of the object is accessible. This is followed by the gripper performing a precision grasp to firmly hold the object. As shown in Fig.\ref{fig:GraspTypes}, the proposed gripper was able to achieve each of these grasp strategies. {Additionally, demonstrations to show that the robotic gripper is suitable for human interaction are shown in Fig. \ref{fig:human-robot interaction} through (a) a handshake and (b) a fist bump.}
%
%
\subsection{Grasp Strength}
\begin{figure}
    \centering
    \includegraphics[height=3.5cm]{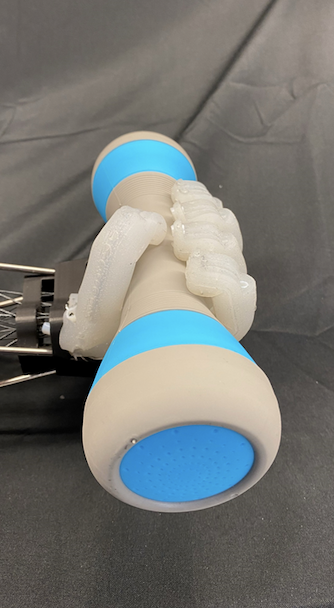}
    \hfill
    \includegraphics[height=3.5cm]{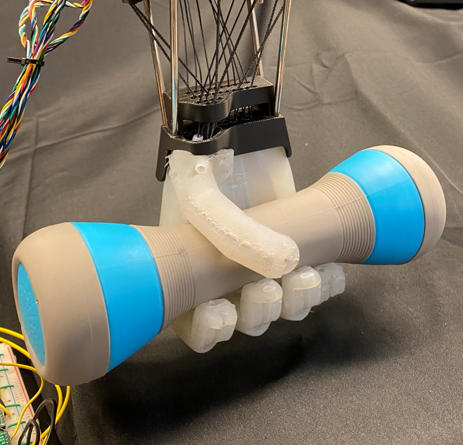}
    \hfill
    \includegraphics[height=3.5cm]{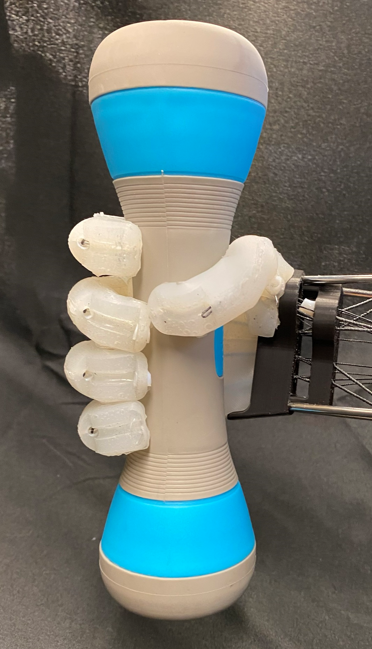}
    \hfill
    \caption{Hand holding a 2\,kg dumbbell in various orientations.}
    \label{fig:dumbbell}
\end{figure}

\begin{figure*}
    \centering
    \subfloat[]{
    \includegraphics[height=8cm]{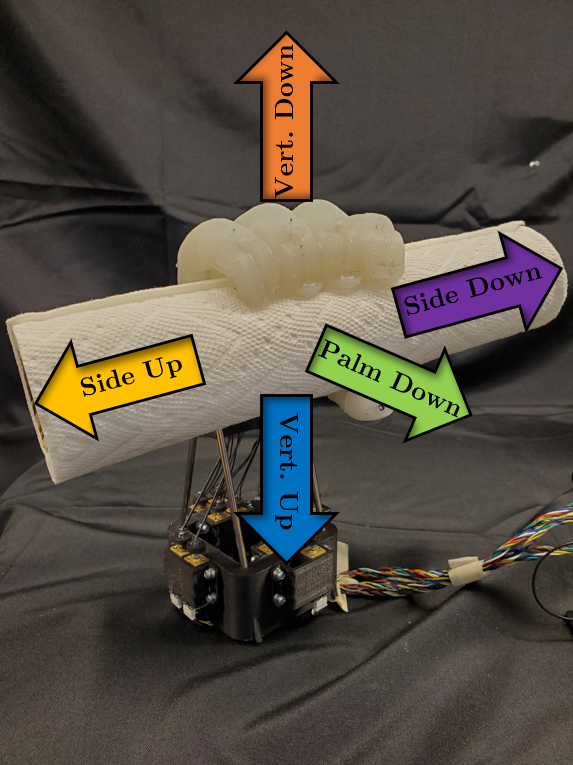}}
    \hfill
    \subfloat[]{
    \includegraphics[height=8cm]{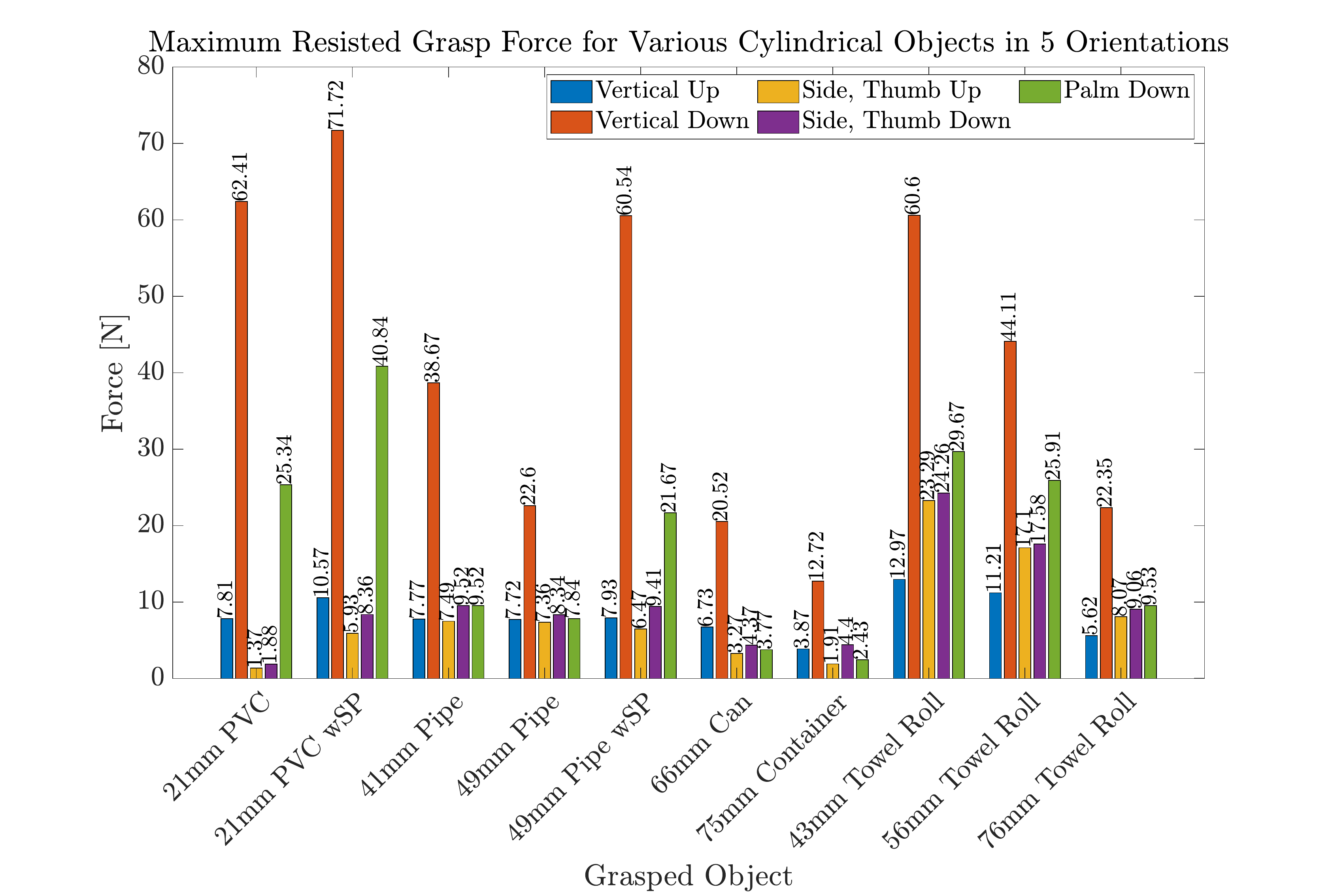}}
    \hfill
    \caption{(a) Various directions in which external force was resisted by the gripper. Load was suspended from a cylinder held in a power grasp for various base orientations. The arrows indicate the direction of this load. \textcolor{red}{(b) The maximum resisted grasp force exhibited by the gripper while holding various cylindrical objects in different loading directions.}}
    \label{fig:GraspingStrength}
\end{figure*}

As a preliminary evaluation of the grasp strength, the gripper was made to lift a 2\,kg dumbbell in different orientations as shown in Fig. \ref{fig:dumbbell}. The total mass of the gripper, including the end-effector, base, and actuators, was 565\,g. This meant it supported nearly six times its own weight in most orientations in a power grasp. The grasp strength of the proposed gripper was further evaluated by investigating its ability to hold on to an object when an external force was applied to pull the object out of the gripper's grasp. It is noted that the gravitational force of the object being held by the gripper was not considered to be an external force. For this purpose, the gripper was made to hold cylindrical objects made with different materials in a power grasp, and forces were applied by suspending known masses from the object. The direction of the load was changed by adjusting the orientation of the base, and the directions in which the forces were applied are shown in Fig. \ref{fig:GraspingStrength}(a).

\textcolor{red}{Fig. \ref{fig:GraspingStrength}(b) shows the maximum force resisted while holding a variety of objects in five different loading directions. The objects grasped were all cylindrical and included a 21-mm-diameter PVC pipe, a 41-mm hard plastic cylinder, a 49-mm metal pipe, a 66-mm hard aluminum can, a 75-mm soft plastic container, and 43-mm, 56-mm, and 76-mm paper towel rolls. The hard pipes, cylinders, and can were rigid objects with smooth surface finishes. The soft plastic container was compliant and had a smooth surface finish. The paper towel rolls were compliant and had high surface friction. Finally, sandpaper was adhered to the 21-mm PVC and 49-mm metal pipe to see how much the added surface friction would affect the grasping performance. These results are presented with the label ``wSP" in Fig. \ref{fig:GraspingStrength}(b). The results show that friction and object compliance significantly increase the grasping potential for the side and vertical up orientations, where the actuation of the TSAs does not directly oppose the loading. Medium-sized objects also performed the best in these orientations, as it was difficult to achieve a tight grasp around smaller objects and wrap the fingers fully around larger objects. In the vertical down and palm down orientations, where the tension in the TSAs could directly oppose the loading, the gripper was found to be much stronger. In these orientations, the compliance of the grasped objects improved the performance. However, size and friction were more deterministic factors, as smaller objects with higher friction provided the highest grasping potential. The 21-mm PVC pipe with sandpaper adhered was the highest tested, resisting almost 72\,N in the vertical down orientation. Additionally, the 21-mm PVC, 49-mm metal pipe with sandpaper, and 43-mm paper towel roll each showed grasping potentials of at least 60\,N in this orientation.}

\subsection{In-hand Manipulation}
\begin{figure}[h!]
    \centering
    \subfloat[]{
    \includegraphics[width=6cm]{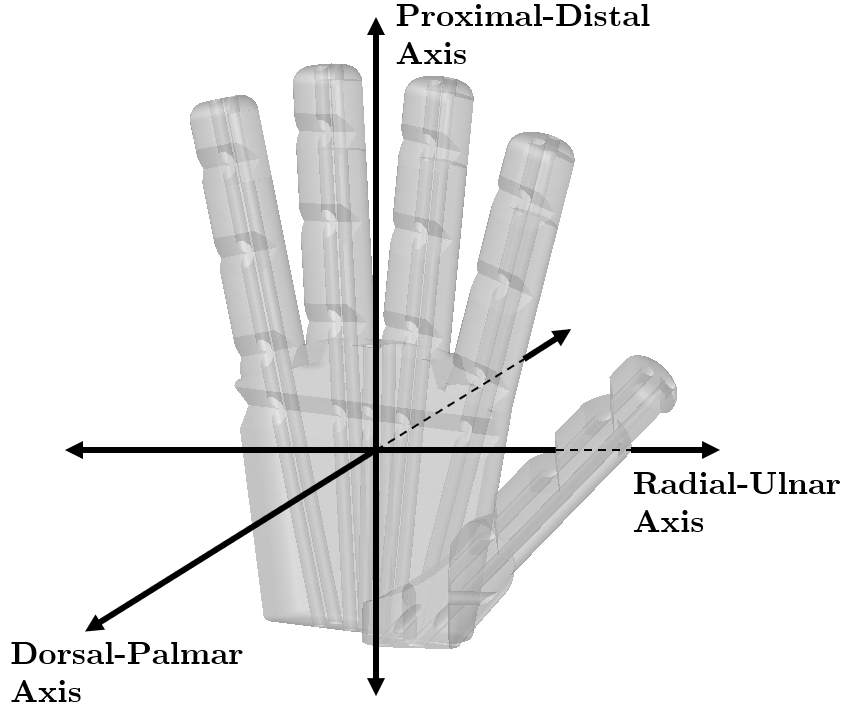}}
    \hfill
    \subfloat[]{
    \includegraphics[width=7.5cm]{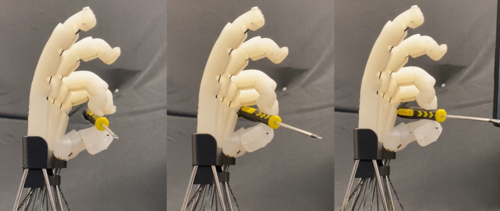}}
    \hfill
    \subfloat[]{
    \includegraphics[width=7.5cm]{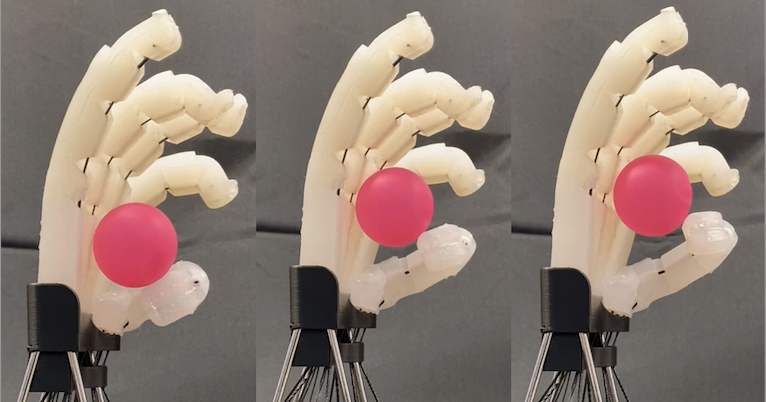}}
    \hfill
    \subfloat[]{
    \includegraphics[width=7.5cm]{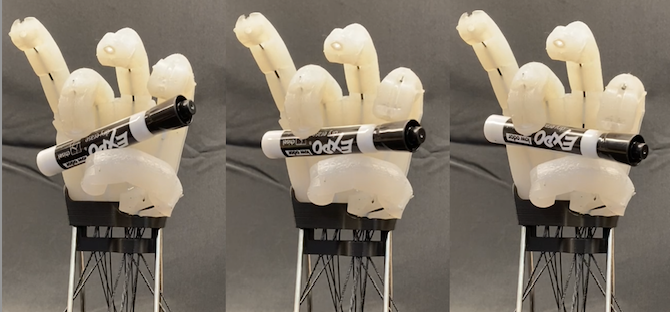}}
    \hfill
    \caption{Hand performing simple in-hand manipulations. (a) Three axes of manipulation of the gripper. The frame is fixed to the palm. (b) Rotation of a screwdriver about the proximal-distal axis. (c) Rotation of a ball about the radial-ulnar axis. (d) Rotation of a marker about the dorsal-palmar axis.}
    \label{fig:InHandManipulation}
\end{figure}

Due to the inclusion of a multi-DOF thumb, the proposed gripper was also capable of preliminary in-hand manipulation. Some simple tests were conducted to assess the gripper's in-hand manipulation ability in its current configuration, though no models or controls for in-hand manipulation were developed. These tests were performed using human-controlled actuation of the various TSAs. Three types of manipulation patterns were demonstrated \cite{RBO3}: (1) rotation of the object about the proximal--distal axis, (2) rotation of the object about radial--ulnar axis, and (3) rotation of the object about the palmar--dorsal axis. These axes are shown in Fig. \ref{fig:InHandManipulation})(a). The objects that demonstrated the aforementioned manipulation patterns were a marker, a screwdriver, and a ping-pong ball, respectively. 

The results of the in-hand manipulation experiments are presented in Fig. \ref{fig:InHandManipulation})(b)--(c). Firstly, rotation about the proximal--distal axis was achieved by using the thumb and the middle finger for maintaining the grip on the object, and the index finger for the manipulation (Fig. \ref{fig:InHandManipulation}(b)). Secondly, rotation about the radial--ulnar axis was achieved by using the thumb for maintaining the grip on the object as well as manipulation (Fig. \ref{fig:InHandManipulation}(c)). Lastly, rotation about the dorsal--palmar axis was achieved by using the thumb for maintaining the grip on the object, and the ring and index fingers for both maintaining the grip on the object as well as manipulation (Fig. \ref{fig:InHandManipulation}(d)).

This manipulation was enabled by combining the high frictional coefficient of the silicone rubber of the gripper and its dexterity. Early results indicated that a future version of a gripper produced with similar materials and design could be capable of more robust in-hand manipulation. {Greater dexterity, which could be achieved by adding more degrees of freedom in the palm, thumb, and base of each finger, would enable more complex manipulation. More work would also be required to model and automate the in-hand manipulation process.}
\subsection{Comparison with Other Anthropomorphic Grippers}
%
{Although the reported metrics of gripper performance vary greatly in existing literature, the metrics that could be collected indicated that our gripper had comparable or superior performance relative to similar grippers that used alternative actuation methods. The comparison of our gripper to others in relevant literature is summarized in Table I.}

{For their combination of controllability and high force output, pneumatic actuators have been used in many grippers aiming to be both anthropomorphic and dexterous \cite{Deimal2016, RBO3, Jianshu2018, Jianshu2019, shorthose2022, wangGripper2021}. Applications of non-pneumatic actuation methods, such as SMAs \cite{kimSMA2016}, SMTAs \cite{ProstheticHand2020}, or combined actuation methods \cite{liDualActuator2021} have been less developed in high-performance grippers, and existing grippers using these actuators have been limited in force output, dexterity, or both. Comparing our gripper to all of these examples, the maximum grasp force was higher than all the grippers, confirming that TSAs can provide high actuation forces in soft robotic grippers. Several of these grippers reported their mass excluding the weight of their full actuation systems \cite{shorthose2022, wangGripper2021}, which made a fair comparison of gripper masses difficult. However, among the papers that do report complete mass, our gripper was the lightest with the exception of \cite{ProstheticHand2020}, which does not claim the high degree of dexterity we achieved. Similarly, the cost was difficult to compare, but the low overall cost of our gripper, which was made possible by using inexpensive motors, indicates that this was also an advantage of using TSAs.}
\begin{table*}[]
\centering

\caption{Comparison with other soft anthropomorphic grippers using different actuation methods in reverse order of publication.}
\label{tab:basic_comparison_table}
\begin{tabular}{|c|c|c|c|c|c|c|c|c|c|c|}

\hline
\begin{tabular}{@{}c@{}}\textbf{Soft} \\ \textbf{Gripper} \end{tabular}     & 
\begin{tabular}{@{}c@{}}\textbf{Actuation} \\ \textbf{Method}\end{tabular}&

\begin{tabular}{@{}c@{}}\textbf{Actuation} \\ \textbf{DOFs}\end{tabular}&
\begin{tabular}{@{}c@{}}\textbf{Force} \\ \textbf{Resisted}\end{tabular}&
\begin{tabular}{@{}c@{}}\textbf{Weight}\end{tabular}&
\begin{tabular}{@{}c@{}}\textbf{Cost}\end{tabular}&
\begin{tabular}{@{}c@{}}\textbf{Dexterous?}\end{tabular}&
\begin{tabular}{@{}c@{}}\textbf{Kapandji} \\ \textbf{Test Score}\end{tabular}&
\begin{tabular}{@{}c@{}c@{}}\textbf{Feix} \\\textbf{Graps} \\ \textbf{(/33)}\end{tabular}&
\begin{tabular}{@{}c@{}}\textbf{Variable} \\ \textbf{Stiffness?}\end{tabular}&
\begin{tabular}{@{}c@{}@{}c}\textbf{In-Hand} \\ \textbf{Manipulation} \\ \textbf{Demo?}\end{tabular}\\ \hline

\begin{tabular}{@{}c@{}}Our \\ Gripper \end{tabular}
& TSAs & 6 & 32 N 
& \begin{tabular}{@{}c@{}@{}c}565\,g \\ (incl. \\ actuators) \end{tabular}
& \begin{tabular}{@{}c@{}@{}c}\$500 \\ (incl. \\ actuators) \end{tabular}
& Y & 6/10 & 31 & Y & Y \\ \hline
\begin{tabular}{@{}c@{}}RBO Hand \\ 3 \cite{RBO3}\end{tabular} 
& \begin{tabular}{@{}c@{}}Pneumatic \\ Actuators\end{tabular} 
& 16 & 39\,N & -- & -- & Y & 10/10 & 33 & N & Y \\ \hline
\begin{tabular}{@{}c@{}@{}c}3D-Printed \\ Pneumatic \\ Hand \cite{shorthose2022}\end{tabular} 
& \begin{tabular}{@{}c@{}}Pneumatic \\ Actuators \end{tabular} 
& -- & 6.42\,N 
& \begin{tabular}{@{}c@{}@{}c}135\,g \\ (excl. \\ actuators) \end{tabular}
& -- & Y & 11/11 & 32 & N & N\\ \hline
\begin{tabular}{@{}c@{}@{}c}Soft Palm \\ Pneumatic \\ Hand \cite{wangGripper2021}\end{tabular} 
& \begin{tabular}{@{}c@{}}Pneumatic \\ Actuators\end{tabular} 
& 20 & 6.15\,N 
& \begin{tabular}{@{}c@{}@{}c}300\,g \\ (excl. \\ actuators) \end{tabular}
& \begin{tabular}{@{}c@{}@{}c}€100-150 \\ (excl.\\ actuators) \end{tabular}
& Y & -- & 32 & N & N \\ \hline
\begin{tabular}{@{}c@{}@{}c}Thermally- \\ Driven \\ Hand \cite{farhan2021}\end{tabular} 
& \begin{tabular}{@{}c@{}}Thermally- \\ Actuated \\ Tendons \end{tabular} 
& 5 & 0.2\,N & -- & -- & N & -- & -- & N & N\\ \hline
\begin{tabular}{@{}c@{}c@{}}Dual-Mode \\ Soft Hand \\ \cite{liDualActuator2021}\end{tabular} 
& \begin{tabular}{@{}c@{}}FEA \& \\ TTA \end{tabular} 
& 11 
& \begin{tabular}{@{}c@{}@{}c}4.3\,N \\ (single \\ finger) \end{tabular}
& \begin{tabular}{@{}c@{}@{}c}1.8\,kg \\ (incl.\\ actuators) \end{tabular}
& -- & Y & -- & -- & N & N\\ \hline
\begin{tabular}{@{}c@{}@{}c}3D-Printed \\ RFiSFA \\ Hand \cite{zhang3dprintedsoftfinger2020}\end{tabular} 
& \begin{tabular}{@{}c@{}}Pneumatic \\ (RFiSFA)\end{tabular} 
& 11 
& \begin{tabular}{@{}c@{}@{}c}0.27\,N \\ (single \\ finger) \end{tabular}
& -- & -- & Y & -- & -- & N & N\\ \hline
\begin{tabular}{@{}c@{}@{}c}Biotensegrity \\ McKibben \\ Hand \cite{liBiotensegrity2020}\end{tabular} 
& \begin{tabular}{@{}c@{}}Pneumatic \\ (McKibben)\end{tabular} 
& 16 
& \begin{tabular}{@{}c@{}}2.26\,N \\ (230\,g) \end{tabular}
& -- & -- & Y & -- & -- & N & N\\ \hline
\begin{tabular}{@{}c@{}c@{}c@{}}SMTA- \\Driven \\ Prosthetic \\ Hand \cite{ProstheticHand2020}\end{tabular} 
& SMTAs
& 5 & 21.5 \,N 
& \begin{tabular}{@{}c@{}@{}c}253\,g \\ (incl. \\ actuators) \end{tabular}
& \begin{tabular}{@{}c@{}@{}c}\$200 \\ (incl. \\ actuators) \end{tabular}
& N & -- & -- & N & N\\ \hline
\begin{tabular}{@{}c@{}}BCL--26 \\ \cite{Jianshu2019}\end{tabular} 
& \begin{tabular}{@{}c@{}}Pneumatic \\ Actuators\end{tabular} 
& 26 & 21.9\,N & -- & -- & Y & 11/11 & 33 & N & Y \\ \hline
\begin{tabular}{@{}c@{}}BCL--13 \\ \cite{Jianshu2018}\end{tabular} 
& \begin{tabular}{@{}c@{}}Pneumatic \\ Actuators\end{tabular} 
& 13 
& \begin{tabular}{@{}c@{}@{}c}8.5\,N \\ (single \\ finger) \end{tabular}
& \begin{tabular}{@{}c@{}@{}c}1.27\,kg \\ (incl. \\ actuators) \end{tabular}
& -- & Y & 8/10 & -- & N & Y \\ \hline
\begin{tabular}{@{}c@{}@{}c}Pneumatic \\ Prosthetic \\ Hand \cite{deviProsthetic2018}\end{tabular} 
& \begin{tabular}{@{}c@{}}Pneumatic \\ Actuators \end{tabular} 
& 10 
& \begin{tabular}{@{}c@{}}2.16\,N \\ (220\,g) \end{tabular}
& \begin{tabular}{@{}c@{}@{}c}950\,g \\ (incl. \\ actuators) \end{tabular}
& \begin{tabular}{@{}c@{}@{}c}\$800 \\ (incl. \\ actuators) \end{tabular}
& N & -- & -- & N & N\\ \hline
\begin{tabular}{@{}c@{}}RBO Hand \\ 2 \cite{Deimal2016}\end{tabular} 
& \begin{tabular}{@{}c@{}}Pneumatic \\ Actuators\end{tabular} 
& 7 & 8\,N & -- 
& \begin{tabular}{@{}c@{}@{}c}\$100 \\ (excl. \\ actuators) \end{tabular}
& Y & 7/8 & 31 & N & N\\ \hline
\begin{tabular}{@{}c@{}c@{}c@{}}SMA \\ Tendon- \\ Driven \\Hand \cite{kimSMA2016}\end{tabular} 
& \begin{tabular}{@{}c@{}}SMA \\ Tendons \end{tabular} 
& 5 
& \begin{tabular}{@{}c@{}}0.59\,N \\ (60\,g) \end{tabular}
& -- & -- & N & -- & -- & N & N\\ \hline

\end{tabular}
\end{table*}

{Comparing dexterity was also difficult, as performance metrics varied in their reporting and application in their respective papers. For this work, we chose to use the Kapandji Test \cite{KAPANDJI198667} and the comprehensive Feix GRASP Taxonomy \cite{Feix2016} to quantify the dexterity and grasping capabilities of our gripper. Application of the Kapandji Test varied between papers, with a minimum of 8 locations on the hand tested \cite{Deimal2016} and a maximum of 11 locations tested \cite{Jianshu2019, shorthose2022}. We tested 10 points, of which 6 were achieved. This was the lowest score on the Kapandji Test of these grippers. However, our 6 DOFs were also the lowest among these grippers, which helped to explain this evident lack of thumb dexterity. Although our gripper only had 6 DOFs, we were able to achieve 31 of the 33 grasps from the Feix grasps, which was comparable to other highly dexterous soft grippers. We also demonstrated basic in-hand manipulation abilities with our gripper. In-hand manipulation performance was not consistently quantified in existing literature, so this was also difficult to compare between grippers. Although our manipulation abilities were likely limited compared to other works \cite{RBO3, Jianshu2019} due to our fewer DOFs, the basic demonstrations shown in this paper prove that the precise control and dexterity required for in-hand manipulation can be achieved using TSAs.}

{The combination of high force output and low cost, along with reasonable thumb dexterity, high-performance grasping, and simple in-hand manipulation achieved by our gripper showed that TSAs can be highly-effective actuators for dexterous soft robotic grippers.}

\section{Limitations}

\subsection{Usage of Soft Material}
{As described in Section I, the use of soft materials to fabricate dexterous anthropomorphic grippers has multiple advantages. However, the inherent limitations of using soft materials, namely, (1) slightly unpredictable deformation of the material, (2) increased nonlinearity of the components, and (3) low force outputs in comparison to rigid grippers, naturally appear in the proposed gripper.} 

Firstly, although the fingers had a maximum theoretical bending angle of 270$\degree$, the experimental characterization revealed that the maximum bending angle was approximately 230$\degree$ (Fig. \ref{fig:max_bending}). This discrepancy could be a result of the compliance of the soft fingers. The theoretical maximum bending angle was calculated under the assumption that the finger would only deform due to bending while actuated. In reality, it appeared there was some axial deformation that reduced the internal angles of the triangular actuation cuts, which reduced the true maximum bending angle.

Secondly, the motor angle--fingertip angle correlations of the fingers and the thumb were nonlinear (Figs. \ref{fig:characterization} and \ref{fig:thumb}). The nonlinearities consisted of hysteresis with lonely stroke behavior and some creep. These nonlinearities could be due to the inherent material properties of the silicone \cite{DB_RK_2022} and the frictional force between the strings and the PTFE tubes \cite{palli_sliding_2016,suthar_conduit_2018}. {Readers are encouraged to read \cite{palli_sliding_2016}, where the force outputs of a TSA are modeled while sliding on a surface with friction.} Although all the fingers exhibited acceptable consistency in their respective motor angle--fingertip angle correlations under different orientations, the presence of the aforementioned nonlinearities could complicate the physics-based modeling process. In future work, data-driven approaches similar to the ones presented in \cite{ral_bombara21,Konda2022} could be used. The design of the fingers could also be modified to reduce the nonlinearities' effects.

Thirdly, the force outputs of the fingers could be increased by using higher-torque motors. However, this could result in decreased actuation speed of the fingers if the torque was increased with a gearbox. Since the actuation speed of the TSAs depends on the radius of the strings used, optimization techniques could be employed to select a motor-string combination that permits the required force output yet maximizes the actuation speed. Lastly, the adjustable stiffness capabilities of the proposed gripper can be further explored by using softer material to fabricate the gripper. {However, as discussed in Section III, using a softer material would also make the gripper less structured, potentially decreasing the lifespan of the silicone and the weight that can be supported by the gripper, as well as exacerbating the axial deformation issue noticed with the actuation cuts.} The silicone in this work caused mild variation of the fingers' stiffness.
%
%
\subsection{Grasping Capabilities and Thumb Dexterity}
The proposed gripper managed to achieve almost all the grasps presented in \cite{Feix2016}. However, it failed to realize the parallel extension grip (grasp \#22) and the adduction grip (grasp \#23). Furthermore, the thumb design presented in this work only achieved a score of 6 out of 10 on the Kapandji test. Although these results were deemed to be acceptable, further improvements could be incorporated in the next iteration of the gripper to address these grasping-related limitations. Firstly, by removing the pseudo joints from the finger design and adopting a continuous design similar to the design presented in \cite{Deimal2016,Jianshu2018,Jianshu2019}, the grasping performance would improve. This may also help the gripper achieve the extension grasp. With a continuous design, the compliance of the fingers increases, thereby allowing them to achieve the grasps in a more efficient manner. Secondly, including an additional controllable DOF in each finger could enable the gripper to achieve better grasping performance, as demonstrated in previous studies \cite{RBO3,Wood2020,Teeple2020}. Lastly, including a controllable DOF in the palm of the gripper could lead to a higher score on the Kapandji test and also achieve the adduction grip. Previous studies have shown that a controllable DOF in the palm aids the thumb to touch the locations on the little finger in the Kapandji test \cite{Deimal2016}, which could help our gripper to achieve the missed little finger and distal palmar crease contacts.

{However, increasing the degrees of freedom of the hand would also increase structural and manufacturing complexity, weight, and size. Accordingly, any additional degrees of freedom must significantly improve the grasping performance, dexterity, or manipulation capabilities of the hand beyond the benchmark of the current iteration to justify their inclusion.}
\subsection{Exclusion of Sensors}
{Despite limitations due to the usage of soft material and limited controllable DOFs, the proposed gripper performed satisfactorily in achieving different Feix GRASP taxonomy grasps, resistance to external force, and preliminary in-hand manipulation. However, all the aforementioned tasks were performed in an open-loop manner. Developing closed-loop and high-level control to improve the grasping capabilities of the gripper can be completed in future work. This is not possible without sensors that record the state of the gripper in terms of its posture and force outputs. Whereas the position of the fingers could be estimated using the motor rotations, this strategy would still be considered open-loop. This is because the data from the encoders are not directly used to control the posture of a particular finger \cite{DB_RK_2021}. Furthermore, sensors to measure the force outputs from the fingers and palm will also be required to perform effective grasping \cite{shorthose2022}. The inclusion of sensors in the gripper design will be explored as a part of future work. To maintain the compactness of the gripper, self-sensing TSAs, which use conductive supercoiled polymer (SCP) strings, can be used to measure the pose of a finger \cite{DavidSoro20,ral_bombara21}. In addition, tactile sensors \cite{shorthose2022} could be embedded into the design of the gripper for force sensing.}    

\section{Conclusion and Future Work}
In this paper, the use of TSAs to drive a soft robotic gripper was explored. Firstly, the design and fabrication of the gripper were discussed. The design included a soft thumb with two controllable DOFs. It additionally included an adjustable stiffness mechanism. Secondly, the fingers were experimentally characterized in terms of their bending angles, force outputs, actuation speeds, and adjustable stiffness capabilities, and these experimental data were statistically analyzed. Finally, the grasping capabilities of the gripper were demonstrated. The experimental results and comparison confirmed the high performance of the TSA-driven soft robotic gripper.

Despite the satisfactory capabilities of the proposed robotic gripper, further improvements can be made. Firstly, the design of the fingers and thumb could be modified to minimize the nonlinear effects in their actuation profiles and to improve the grasping capabilities and the dexterity of the gripper. Additional controllable DOFs could also be included to improve the grasping performance and manipulation abilities of the gripper. Secondly, mathematical models could be developed to accurately predict the behavior of the gripper. Thirdly, the in-hand manipulation capabilities of the proposed robotic gripper could be further studied. For this purpose, advanced motion planning and grasping algorithms which utilize machine learning techniques \cite{ML1,ML_2} could be developed.

\section{Acknowledgements}
The authors would like to thank Mr. Evan Laske from the NASA Johnson Space Center for intellectual discussions.

\bibliography{0_main}

\begin{thebibliography}{10}
\providecommand{\url}[1]{#1}
\csname url@samestyle\endcsname
\providecommand{\newblock}{\relax}
\providecommand{\bibinfo}[2]{#2}
\providecommand{\BIBentrySTDinterwordspacing}{\spaceskip=0pt\relax}
\providecommand{\BIBentryALTinterwordstretchfactor}{4}
\providecommand{\BIBentryALTinterwordspacing}{\spaceskip=\fontdimen2\font plus
\BIBentryALTinterwordstretchfactor\fontdimen3\font minus
  \fontdimen4\font\relax}
\providecommand{\BIBforeignlanguage}[2]{{%
\expandafter\ifx\csname l@#1\endcsname\relax
\typeout{** WARNING: IEEEtran.bst: No hyphenation pattern has been}%
\typeout{** loaded for the language `#1'. Using the pattern for}%
\typeout{** the default language instead.}%
\else
\language=\csname l@#1\endcsname
\fi
#2}}
\providecommand{\BIBdecl}{\relax}
\BIBdecl

\bibitem{Shih2017}
B.~{Shih}, D.~{Drotman}, C.~{Christianson}, Z.~{Huo}, R.~{White}, H.~I.
  {Christensen}, and M.~T. {Tolley}, ``Custom soft robotic gripper sensor skins
  for haptic object visualization,'' in \emph{2017 IEEE/RSJ International
  Conference on Intelligent Robots and Systems (IROS)}, 2017, pp. 494--501.

\bibitem{Yafeng2021}
Y.~Cui, X.-J. Liu, X.~Dong, J.~Zhou, and H.~Zhao, ``Enhancing the universality
  of a pneumatic gripper via continuously adjustable initial grasp postures,''
  \emph{IEEE Transactions on Robotics}, vol.~37, no.~5, pp. 1604--1618, 2021.

\bibitem{MATTAR2013517}
E.~Mattar, ``A survey of bio-inspired robotics hands implementation: New
  directions in dexterous manipulation,'' \emph{Robotics and Autonomous
  Systems}, vol.~61, no.~5, pp. 517--544, 2013.

\bibitem{Deimal2016}
R.~Deimel and O.~Brock, ``A novel type of compliant and underactuated robotic
  hand for dexterous grasping,'' \emph{The International Journal of Robotics
  Research}, vol.~35, no. 1-3, pp. 161--185, 2016.

\bibitem{SoftGripperReview}
J.~Shintake, V.~Cacucciolo, D.~Floreano, and H.~Shea, ``Soft robotic
  grippers,'' \emph{Advanced Materials}, vol.~30, no.~29, p. 1707035, 2018.

\bibitem{SMA_Hand2015}
Y.~She, C.~Li, J.~Cleary, and H.-J. Su, ``Design and fabrication of a soft
  robotic hand with embedded actuators and sensors,'' \emph{Journal of
  Mechanisms and Robotics}, vol.~7, no.~2, 2015.

\bibitem{RBO3}
S.~Puhlmann, J.~Harris, and O.~Brock, ``{R}{B}{O} {H}and 3 -- a platform for
  soft dexterous manipulation,'' \emph{arXiv:2201.10883}, 2022.

\bibitem{tro_zhang}
J.~Zhang, J.~Sheng, C.~O'Neill, C.~J. Walsh, R.~J. Wood, J.~H. Ryu, J.~P.
  Desai, and M.~C. Yip, ``Robotic artificial muscles: Current progress and
  future perspectives for biomimetic actuators,'' \emph{IEEE Transactions on
  Robotics}, vol.~35, no.~3, pp. 761--781, 2019.

\bibitem{Sun2021}
J.~Sun, B.~Tighe, Y.~Liu, and J.~Zhao, ``Twisted-and-coiled actuators with free
  strokes enable soft robots with programmable motions,'' \emph{Soft Robotics},
  vol.~8, no.~2, pp. 213--225, 2021.

\bibitem{Hasel_2021}
P.~Rothemund, N.~Kellaris, S.~K. Mitchell, E.~Acome, and C.~Keplinger,
  ``{HASEL} artificial muscles for a new generation of lifelike robots—recent
  progress and future opportunities,'' \emph{Advanced Materials}, vol.~33,
  no.~19, 2021.

\bibitem{dea_control_17}
Z.~Ye, Z.~Chen, R.~Asmatulu, and H.~Chan, ``Robust control of dielectric
  elastomer diaphragm actuator for human pulse signal tracking,'' \emph{Smart
  Mater. Struct.}, vol.~26, no.~8, p. 085043, 2017.

\bibitem{SMA_Octopus2012}
C.~Laschi, M.~Cianchetti, B.~Mazzolai, L.~Margheri, M.~Follador, and P.~Dario,
  ``Soft robot arm inspired by the octopus,'' \emph{Advanced Robotics},
  vol.~27, no.~7, pp. 709--727, 2012.

\bibitem{michael2017}
M.~C. Yip and G.~Niemeyer, ``On the control and properties of supercoiled
  polymer artificial muscles,'' \emph{IEEE Transactions on Robotics}, vol.~33,
  no.~3, pp. 689--699, 2017.

\bibitem{Tang_2019}
X.~Tang, K.~Li, Y.~Liu, D.~Zhou, and J.~Zhao, ``A general soft robot module
  driven by twisted and coiled actuators,'' \emph{Smart Materials and
  Structures}, vol.~28, no.~3, p. 035019, 2019.

\bibitem{lauDEA2017}
G.-K. Lau, K.-R. Heng, A.~S. Ahmed, and M.~Shrestha, ``Dielectric elastonomer
  fingers for versatile grasping and nimble pinching,'' \emph{Applied Physics
  Letters}, vol. 110, no.~18, 2017.

\bibitem{Model_Tmech14}
I.~Gaponov, D.~Popov, and J.~H. Ryu, ``Twisted string actuation systems: A
  study of the mathematical model and a comparison of twisted strings,''
  \emph{IEEE/ASME Transactions on Mechatronics}, vol.~19, no.~4, pp.
  1331--1342, 2014.

\bibitem{auxilio_17}
I.~Gaponov, D.~Popov, S.~J. Lee, and J.-H. Ryu, ``Auxilio: A portable
  cable-driven exosuit for upper extremity assistance,'' \emph{International
  Journal of Control, Automation and Systems}, vol.~15, no.~1, pp. 73--84,
  2017.

\bibitem{impedance_iccas14}
I.~W. Park and V.~SunSpiral, ``Impedance controlled twisted string actuators
  for tensegrity robots,'' in \emph{Proc. Int. Conf. Control, Automation and
  Systems}, 2014, pp. 1331--1338.

\bibitem{Shin_TRO_2012}
Y.~J. Shin, H.~J. Lee, K.-S. Kim, and S.~Kim, ``A robot finger design using a
  dual-mode twisting mechanism to achieve high-speed motion and large grasping
  force,'' \emph{IEEE Transactions on Robotics}, vol.~28, no.~6, pp.
  1398--1405, 2012.

\bibitem{thulanijournal2021}
T.~Tsabedze, E.~Hartman, and J.~Zhang, ``A compact, compliant, and biomimetic
  robotic assistive glove driven by twisted string actuation,''
  \emph{International Journal of Intelligent Robotics and Applications},
  vol.~5, no.~3, pp. 381--394, 2021.

\bibitem{OT4}
D.~{Lee}, D.~H. {Kim}, C.~H. {Che}, J.~B. {In}, and D.~{Shin}, ``Highly durable
  bidirectional joint with twisted string actuators and variable radius
  pulley,'' \emph{IEEE/ASME Transactions on Mechatronics}, vol.~25, no.~1, pp.
  360--370, 2020.

\bibitem{MULLER2015207}
R.~Müller, M.~Hessinger, H.~Schlaak, and P.~Pott, ``Modelling and
  characterisation of twisted string actuation for usage in active knee
  orthoses,'' \emph{IFAC-PapersOnLine}, vol.~48, no.~20, pp. 207--212, 2015,
  9th IFAC Symposium on Biological and Medical Systems BMS 2015.

\bibitem{Shisheie2015}
R.~Shisheie, L.~Jiang, L.~Banta, and M.~Cheng, ``{Modeling and control of a
  bidirectional twisted-string actuation for an upper arm robotic device},''
  \emph{Proceedings of the American Control Conference}, vol. 2015-July, pp.
  5794--5799, 2015.

\bibitem{Popov2014}
D.~Popov, I.~Gaponov, and J.~H. Ryu, ``{Towards variable stiffness control of
  antagonistic twisted string actuators},'' \emph{IEEE International Conference
  on Intelligent Robots and Systems}, no. Iros, pp. 2789--2794, 2014.

\bibitem{DB_RK_2022}
D.~Bombara, R.~Konda, E.~Chow, and J.~Zhang, ``Physics-based kinematic modeling
  of a twisted string actuator-driven soft robotic manipulator,'' in \emph{2022
  American Control Conference (ACC)}, 2022.

\bibitem{6697204}
D.~{Popov}, I.~{Gaponov}, and J.~{Ryu}, ``Bidirectional elbow exoskeleton based
  on twisted-string actuators,'' in \emph{2013 IEEE/RSJ International
  Conference on Intelligent Robots and Systems}, 2013, pp. 5853--5858.

\bibitem{TSA_Suit_2020}
M.~Hosseini, R.~Meattini, A.~San-Millan, G.~Palli, C.~Melchiorri, and J.~Paik,
  ``A {sEMG}-driven soft exosuit based on twisted string actuators for elbow
  assistive applications,'' \emph{IEEE Robotics and Automation Letters},
  vol.~5, no.~3, pp. 4094--4101, 2020.

\bibitem{DB_RK_2021}
D.~Bombara, R.~Coulter, R.~Konda, and J.~Zhang, ``A twisted string
  actuator-driven soft robotic manipulator,'' in \emph{2021 Modeling,
  Estimation, and Control Conference (MECC)}, 2021.

\bibitem{Zhou2019}
H.~Zhou, A.~Mohammadi, D.~Oetomo, and G.~Alici, ``A novel monolithic soft
  robotic thumb for an anthropomorphic prosthetic hand,'' \emph{IEEE Robotics
  and Automation Letters}, vol.~4, no.~2, pp. 602--609, 2019.

\bibitem{DEXMART}
G.~Palli, C.~Melchiorri, G.~Vassura, U.~Scarcia, L.~Moriello, G.~Berselli,
  A.~Cavallo, G.~D. Maria, C.~Natale, S.~Pirozzi, C.~May, F.~Ficuciello, and
  B.~Siciliano, ``The {D}{E}{X}{M}{A}{R}{T} hand: Mechatronic design and
  experimental evaluation of synergy-based control for human-like grasping,''
  \emph{The International Journal of Robotics Research}, vol.~33, no.~5, pp.
  799--824, 2014.

\bibitem{UCHand}
M.~Tavakoli, R.~Batista, and L.~Sgrigna, ``The {U}{C} softhand: Light weight
  adaptive bionic hand with a compact twisted string actuation system,''
  \emph{Actuators}, vol.~5, no.~1, 2016.

\bibitem{TSAHandRAM2013}
C.~Melchiorri, G.~Palli, G.~Berselli, and G.~Vassura, ``Development of the ub
  hand iv: Overview of design solutions and enabling technologies,'' \emph{IEEE
  Robotics and Automation Magazine}, vol.~20, no.~3, pp. 72--81, 2013.

\bibitem{TSAHand2010}
T.~Sonoda and I.~Godler, ``Multi-fingered robotic hand employing strings
  transmission named “twist drive”,'' in \emph{2010 IEEE/RSJ International
  Conference on Intelligent Robots and Systems}, 2010, pp. 2733--2738.

\bibitem{TSAHandRAL2017}
S.~H. Jeong, K.-S. Kim, and S.~Kim, ``Designing anthropomorphic robot hand with
  active dual-mode twisted string actuation mechanism and tiny tension
  sensors,'' \emph{IEEE Robotics and Automation Letters}, vol.~2, no.~3, pp.
  1571--1578, 2017.

\bibitem{TSAHand2013}
Y.~J. Shin, K.-H. Rew, K.-S. Kim, and S.~Kim, ``Development of anthropomorphic
  robot hand with dual-mode twisting actuation and electromagnetic joint
  locking mechanism,'' in \emph{2013 IEEE International Conference on Robotics
  and Automation}, 2013, pp. 2759--2764.

\bibitem{Shiva2016}
A.~Shiva, A.~Stilli, Y.~Noh, A.~Faragasso, I.~D. Falco, G.~Gerboni,
  M.~Cianchetti, A.~Menciassi, K.~Althoefer, and H.~A. Wurdemann,
  ``Tendon-based stiffening for a pneumatically actuated soft manipulator,''
  \emph{IEEE Robotics and Automation Letters}, vol.~1, no.~2, pp. 632--637,
  2016.

\bibitem{Stilli2014}
A.~Stilli, H.~A. Wurdemann, and K.~Althoefer, ``Shrinkable,
  stiffness-controllable soft manipulator based on a bio-inspired antagonistic
  actuation principle,'' in \emph{2014 IEEE/RSJ International Conference on
  Intelligent Robots and Systems}, 2014, pp. 2476--2481.

\bibitem{Maghooa2015}
F.~Maghooa, A.~Stilli, Y.~Noh, K.~Althoefer, and H.~A. Wurdemann, ``Tendon and
  pressure actuation for a bio-inspired manipulator based on an antagonistic
  principle,'' in \emph{2015 IEEE International Conference on Robotics and
  Automation (ICRA)}, 2015, pp. 2556--2561.

\bibitem{spie15_variable}
S.~Kianzad, J.~D. Pandit, Milind~Lewis, A.~R. Berlingeri, K.~J. Haebler, and
  J.~D. Madden, ``Variable stiffness and recruitment using nylon actuators
  arranged in a pennate configuration,'' in \emph{Proceedings of SPIE on
  Electroactive Polymer Actuators and Devices (EAPAD)}, vol. 9430, 2015.

\bibitem{icra19_scp}
Y.~{Yang}, Z.~{Kan}, Y.~{Zhang}, Y.~A. {Tse}, and M.~Y. {Wang}, ``A novel
  variable stiffness actuator based on pneumatic actuation and supercoiled
  polymer artificial muscles,'' in \emph{Proceedings of IEEE International
  Conference on Robotics and Automation}, 2019, pp. 3983--3989.

\bibitem{Bhatt2021}
A.~Bhatt, A.~Sieler, S.~Puhlmann, and O.~Brock, ``Surprisingly robust in-hand
  manipulation: An empirical study,'' in \emph{2021 Robotics: Science and
  Systems (RSS)}, 2021.

\bibitem{Jianshu2018}
J.~Zhou, J.~Yi, X.~Chen, Z.~Liu, and Z.~Wang, ``{BCL}-13: A 13-{DOF} soft
  robotic hand for dexterous grasping and in-hand manipulation,'' \emph{IEEE
  Robotics and Automation Letters}, vol.~3, no.~4, pp. 3379--3386, 2018.

\bibitem{Jianshu2019}
J.~Zhou, X.~Chen, U.~Chang, J.-T. Lu, C.~C.~Y. Leung, Y.~Chen, Y.~Hu, and
  Z.~Wang, ``A soft-robotic approach to anthropomorphic robotic hand
  dexterity,'' \emph{IEEE Access}, vol.~7, pp. 101\,483--101\,495, 2019.

\bibitem{Shunya2019}
S.~Ohkura, D.~Shinohira, T.~Yoshida, T.~Kanno, T.~Miyazaki, T.~Kawase, and
  K.~Kawashima, ``Development of three-fingered end-effector using pneumatic
  soft actuators,'' in \emph{2019 IEEE/SICE International Symposium on System
  Integration (SII)}, 2019, pp. 701--706.

\bibitem{Amir2017}
A.~Firouzeh and J.~Paik, ``Grasp mode and compliance control of an
  underactuated origami gripper using adjustable stiffness joints,''
  \emph{IEEE/ASME Transactions on Mechatronics}, vol.~22, no.~5, pp.
  2165--2173, 2017.

\bibitem{Charbel2019}
C.~Tawk, A.~Gillett, M.~in~het Panhuis, G.~M. Spinks, and G.~Alici, ``A
  {3D}-printed omni-purpose soft gripper,'' \emph{IEEE Transactions on
  Robotics}, vol.~35, no.~5, pp. 1268--1275, 2019.

\bibitem{Feix2016}
T.~Feix, J.~Romero, H.-B. Schmiedmayer, A.~M. Dollar, and D.~Kragic, ``The
  {G}{R}{A}{S}{P} taxonomy of human grasp types,'' \emph{IEEE Transactions on
  Human-Machine Systems}, vol.~46, no.~1, pp. 66--77, 2016.

\bibitem{ProstheticHand2020}
A.~Mohammadi, J.~Lavranos, H.~Zhou, R.~Mutlu, G.~Alici, Y.~Tan, P.~Choong, and
  D.~Oetomo, ``A practical 3{D}-printed soft robotic prosthetic hand with
  multi-articulating capabilities,'' \emph{PLOS ONE}, vol.~15, no.~5, pp.
  1--23, 05 2020.

\bibitem{Mariangela2015}
M.~Manti, T.~Hassan, G.~Passetti, N.~D'Elia, C.~Laschi, and M.~Cianchetti, ``A
  bioinspired soft robotic gripper for adaptable and effective grasping,''
  \emph{Soft Robotics}, vol.~2, no.~3, pp. 107--116, 2015.

\bibitem{act5010001}
M.~Tavakoli, R.~Batista, and L.~Sgrigna, ``The {UC} softhand: Light weight
  adaptive bionic hand with a compact twisted string actuation system,''
  \emph{Actuators}, vol.~5, no.~1, 2016.

\bibitem{polym10080846}
V.~Slesarenko, S.~Engelkemier, P.~I. Galich, D.~Vladimirsky, G.~Klein, and
  S.~Rudykh, ``Strategies to control performance of 3{D}-printed, cable-driven
  soft polymer actuators: From simple architectures to gripper prototype,''
  \emph{Polymers}, vol.~10, no.~8, 2018.

\bibitem{Tavakoli2016}
M.~Tavakoli, R.~Batista, and P.~Neto, ``{A compact two-phase twisted string
  actuation system: Modeling and validation},'' \emph{Mechanism and Machine
  Theory}, vol. 101, pp. 23--35, 2016.

\bibitem{suthar_conduit_2018}
B.~{Suthar}, M.~{Usman}, H.~{Seong}, I.~{Gaponov}, and J.~{Ryu}, ``Preliminary
  study of twisted string actuation through a conduit toward soft and wearable
  actuation,'' in \emph{2018 IEEE International Conference on Robotics and
  Automation (ICRA)}, 2018, pp. 2260--2265.

\bibitem{palli_sliding_2016}
G.~{Palli}, M.~{Hosseini}, and C.~{Melchiorri}, ``Twisted string actuation with
  sliding surfaces,'' in \emph{2016 IEEE/RSJ International Conference on
  Intelligent Robots and Systems (IROS)}, 2016, pp. 260--265.

\bibitem{ModelControl_Tmech13}
G.~Palli, C.~Natale, C.~May, C.~Melchiorri, and T.~Wurtz, ``Modeling and
  control of the twisted string actuation system,'' \emph{IEEE/ASME
  Transactions on Mechatronics}, vol.~18, no.~2, pp. 664--673, 2013.

\bibitem{quaglini2011friction}
V.~Quaglini and P.~Dubini, ``Friction of polymers sliding on smooth surfaces,''
  \emph{Advances in Tribology}, vol. 2011, 2011.

\bibitem{lee2019effect}
D.~Lee, I.~Gaponov, and J.-H. Ryu, ``Effect of vibration on twisted string
  actuation through conduit at high bending angles,'' in \emph{2019 IEEE/RSJ
  International Conference on Intelligent Robots and Systems (IROS)}.\hskip 1em
  plus 0.5em minus 0.4em\relax IEEE, 2019, pp. 5965--5970.

\bibitem{Konda2022}
R.~Konda and J.~Zhang, ``Hysteresis with lonely stroke in artificial muscles:
  Characterization, modeling, and inverse compensation,'' \emph{Mechanical
  Systems and Signal Processing}, vol. 164, p. 108240, 2022.

\bibitem{encoder1}
J.~Santoso, E.~H. Skorina, M.~Luo, R.~Yan, and C.~D. Onal, ``Design and
  analysis of an origami continuum manipulation module with torsional
  strength,'' in \emph{2017 IEEE/RSJ International Conference on Intelligent
  Robots and Systems (IROS)}, 2017, pp. 2098--2104.

\bibitem{encoder2}
H.-S. Juang and K.-Y. Lum, ``Design and control of a two-wheel self-balancing
  robot using the arduino microcontroller board,'' in \emph{2013 10th IEEE
  International Conference on Control and Automation (ICCA)}, 2013, pp.
  634--639.

\bibitem{Heinemann2015}
F.~Heinemann, S.~Puhlmann, C.~Eppner, J.~Élvarez Ruiz, M.~Maertens, and
  O.~Brock, ``A taxonomy of human grasping behavior suitable for transfer to
  robotic hands,'' in \emph{2015 IEEE International Conference on Robotics and
  Automation (ICRA)}, 2015, pp. 4286--4291.

\bibitem{KAPANDJI198667}
A.~Kapandji, ``Clinical test of apposition and counter-apposition of the
  thumb,'' \emph{Annales de Chirurgie de la Main}, vol.~5, no.~1, pp. 67--73,
  1986.

\bibitem{shorthose2022}
O.~Shorthose, A.~Albini, L.~He, and P.~Maiolino, ``Design of a 3{D}-printed
  soft robotic hand with integrated distributed tactile-sensing,'' \emph{IEEE
  Robotics and Automation Letters}, vol.~7, no.~2, pp. 3945--3952, 2022.

\bibitem{wangGripper2021}
H.~Wang, F.~J. Abu-Dakka, T.~Nguyen~Le, V.~Kyrki, and H.~Xu, ``A novel soft
  robotic hand design with human-inspired soft palm: achieving a great
  diversity of grasps,'' \emph{IEEE Robotics and Automation Magazine}, vol.~28,
  no.~2, pp. 37--49, 2021.

\bibitem{kimSMA2016}
H.-I. Kim, M.-W. Han, S.-H. Song, and S.-H. Ahn, ``Soft morphing hand driven by
  {SMA} tendon wire,'' \emph{Composites. Part B, Engineering}, vol. 105, pp.
  138--148, 2016.

\bibitem{liDualActuator2021}
Y.~Li, Y.~Chen, T.~Ren, Y.~Hu, H.~Liu, L.~Senyuan, Y.~Yang, Y.~Li, and J.~Zhou,
  ``A dual-mode actuator for soft robotic hand,'' \emph{IEEE Robotics and
  Automation Letters}, vol.~6, no.~2, pp. 1144--1151, 2021.

\bibitem{farhan2021}
M.~Farhan, M.~Behl, K.~Kratz, and A.~Lendlein, ``Origami hand for soft robotics
  driven by thermally controlled polymeric fiber actuators,'' \emph{MRS
  Communications}, vol.~11, no.~4, pp. 476--482, 2021.

\bibitem{zhang3dprintedsoftfinger2020}
N.~Zhang, L.~Ge, H.~Xu, X.~Zhu, and G.~Gu, ``3d printed, modularized
  rigid-flexible integrated soft finger actuators for anthropomorphic hands,''
  \emph{Sensors and actuators. A. Physical}, vol. 312, p. 112090, 2020.

\bibitem{liBiotensegrity2020}
W.-Y. Li, H.~Nabae, G.~Endo, and K.~Suzumori, ``New soft robot hand
  configuration with combined biotensegrity and thin artificial muscle,''
  \emph{IEEE Robotics and Automation Letters}, vol.~5, no.~3, pp. 4345--4351,
  2020.

\bibitem{deviProsthetic2018}
M.~A. Devi, G.~Udupa, and P.~Sreedharan, ``A novel underactuated multi-fingered
  soft robotic hand for prosthetic application,'' \emph{Robotics and Autonomous
  Systems}, vol. 100, pp. 267--277, 2018.

\bibitem{ral_bombara21}
D.~{Bombara}, R.~{Konda}, and J.~{Zhang}, ``Experimental characterization and
  modeling of the self-sensing property in compliant twisted string
  actuators,'' \emph{IEEE Robotics and Automation Letters}, vol.~6, no.~2, pp.
  974--981, 2021.

\bibitem{Wood2020}
S.~Abondance, C.~B. Teeple, and R.~J. Wood, ``A dexterous soft robotic hand for
  delicate in-hand manipulation,'' \emph{IEEE Robotics and Automation Letters},
  vol.~5, no.~4, pp. 5502--5509, 2020.

\bibitem{Teeple2020}
C.~B. Teeple, T.~N. Koutros, M.~A. Graule, and R.~J. Wood, ``Multi-segment soft
  robotic fingers enable robust precision grasping,'' \emph{The International
  Journal of Robotics Research}, vol.~39, no.~14, pp. 1647--1667, 2020.

\bibitem{DavidSoro20}
D.~Bombara, S.~Fowzer, and J.~Zhang, ``Compliant, large-strain, and
  self-sensing twisted string actuators,'' \emph{Soft Robotics}, vol.~9, no.~1,
  pp. 72--88, 2022.

\bibitem{ML1}
K.~Chin, T.~Hellebrekers, and C.~Majidi, ``Machine learning for soft robotic
  sensing and control,'' \emph{Advanced Intelligent Systems}, vol.~2, no.~6, p.
  1900171, 2020.

\bibitem{ML_2}
D.~Kim, S.-H. Kim, T.~Kim, B.~B. Kang, M.~Lee, W.~Park, S.~Ku, D.~Kim, J.~Kwon,
  H.~Lee, J.~Bae, Y.-L. Park, K.-J. Cho, and S.~Jo, ``Review of machine
  learning methods in soft robotics,'' \emph{PLOS ONE}, vol.~16, no.~2, pp.
  1--24, 02 2021.

\end{thebibliography}
\bibliographystyle{IEEEtran}

\begin{IEEEbiography}[{\includegraphics[width=1in,height=1.25in,clip,keepaspectratio]{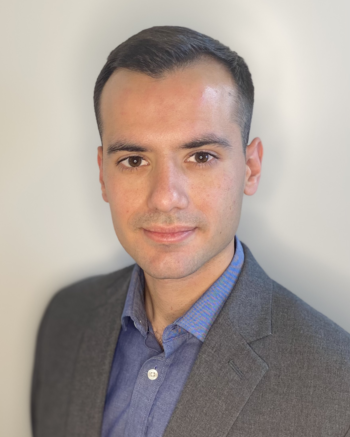}}]%
{David Bombara}
received his {Bachelor of Science (B.S.) and Master of Science (M.S.) degrees in mechanical engineering from the University of Nevada, Reno, NV, USA, in 2020 and 2022, respectively.} {During 2021--2022, he was a fellow for the NASA Space Technology Graduate Research Opportunities (NSTGRO) program. As of August 2022, he is a  Ph.D. student in Materials Science \& Mechanical Engineering at the Harvard John A. Paulson School of Engineering and Applied Sciences. In 2022, he was selected for the National Science Foundation Graduate Research Fellowship Program. His research interests include actuators, soft robots, and robotic manipulation.}
\end{IEEEbiography}
\vskip -1\baselineskip plus -1fil
\begin{IEEEbiography}[{\includegraphics[width=1in,height=1.25in,clip,keepaspectratio]{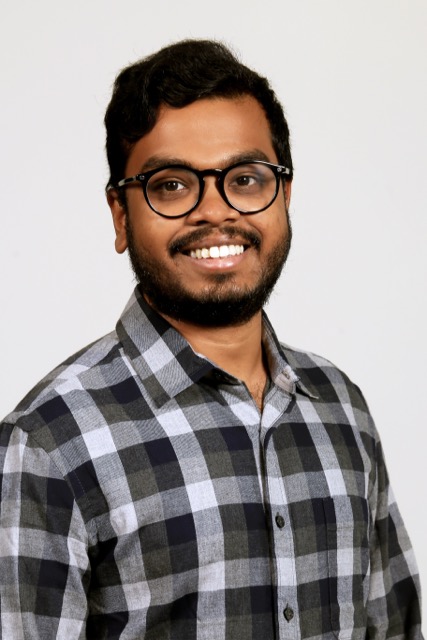}}]%
{Revanth Konda}
received the B.S. degree in mechatronics from Jawaharlal Nehru Technological University, Hyderabad, India, in 2016, and the MSc. degree in mechanical engineering from the Northern Illinois University in 2018. He is currently pursuing his doctoral degree in mechanical engineering at the University of Nevada, Reno. He is interested in artificial muscles and their employment in real-world applications.
\end{IEEEbiography}
\vskip -1\baselineskip plus -1fil
\begin{IEEEbiography}[{\includegraphics[width=1in,height=1.25in,clip,keepaspectratio]{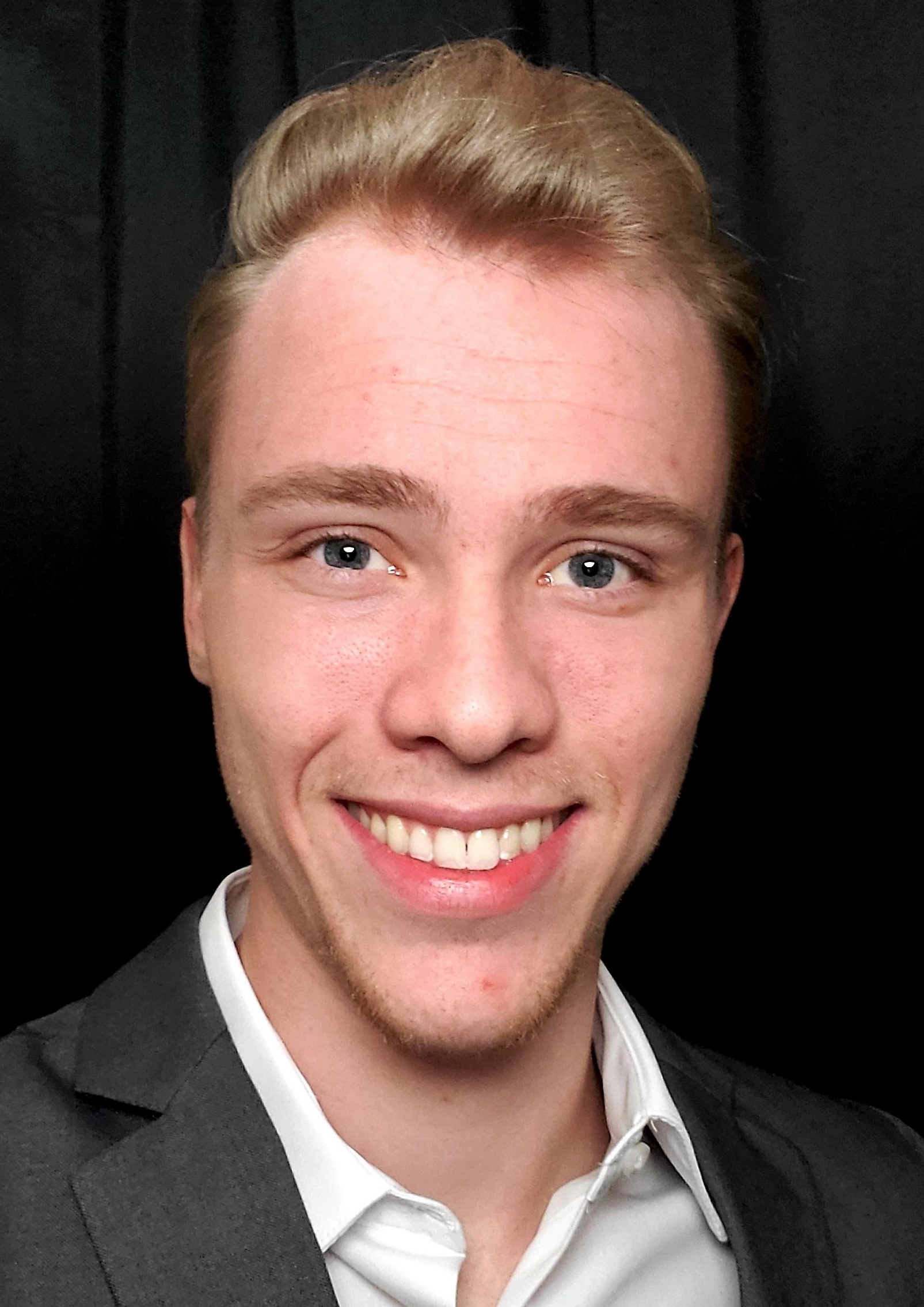}}]%
{Steven Swanbeck}
received his B.S. degree in mechanical engineering from the University of Nevada, Reno, NV, USA, in 2021. As of August 2022, he is pursuing his M.S. degree in mechanical engineering at The University of Texas at Austin, TX, USA. His research interests include grasping and manipulation, human-robot collaboration, and robot deployment in hazardous environments.
\end{IEEEbiography}
\vskip -1\baselineskip plus -1fil
\begin{IEEEbiography}[{\includegraphics[width=1in,height=1.25in,clip,keepaspectratio]{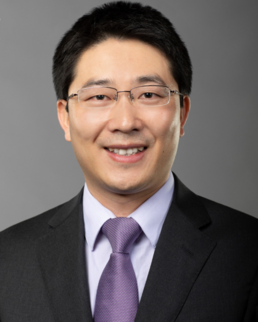}}]%
{Jun Zhang}
(S'12–M'16) received a B.S. degree in automation from the University of Science and Technology of China, Hefei, China, in 2011, and a Ph.D. degree in electrical and computer engineering from the Michigan State University, East Lansing, MI, USA, in 2015.

From 2016 to 2018, he was a Postdoctoral Scholar in electrical and computer engineering with the University of California, San Diego, La Jolla, CA, USA. He is an Assistant Professor with the Department of Mechanical Engineering, University of Nevada, Reno, NV, USA. His research interests include smart materials and artificial muscles for biomimetic, soft, and wearable robots.
\end{IEEEbiography}

\end{document}